\documentclass[11pt]{article}

\usepackage[preprint]{acl}

\usepackage{times}
\usepackage{latexsym}

\usepackage[T1]{fontenc}
\usepackage[utf8]{inputenc}

\usepackage{microtype}

\usepackage{inconsolata}

\usepackage{graphicx}
\usepackage{booktabs}

\usepackage{amssymb}
\usepackage{amsmath}
\usepackage[table]{xcolor}
\usepackage{booktabs}
\usepackage{multirow}
\usepackage{caption}
\usepackage{graphicx}
\usepackage{array}
\usepackage{tabularx}

\usepackage{pifont}

\usepackage{textcomp,colortbl}
\usepackage{enumitem}
\usepackage{cuted}
\usepackage{capt-of}

\usepackage[most]{tcolorbox}

\tcbset{
    promptstage/.style={
        enhanced,
        colback=white,
        boxrule=0.5pt,
        arc=1.5mm,
        left=2mm,
        right=2mm,
        top=1.5mm,
        bottom=1.5mm,
        fonttitle=\small\bfseries,
        fontupper=\small,
        before skip=5pt,
        after skip=5pt
    }
}

\definecolor{preservecolor}{HTML}{b55448}
\definecolor{changecolor}{HTML}{4780a8}

\newtcolorbox{priorityinitial}{
    colback=white,
    colframe=preservecolor,
    boxrule=0.6pt,
    arc=2pt,
    left=6pt,right=6pt,top=5pt,bottom=5pt,
    before skip=0pt,after skip=4pt,
    fontupper=\small,
    coltext=black
}

\newtcolorbox{priorityretraction}{
    colback=white,
    colframe=changecolor,
    boxrule=0.6pt,
    arc=2pt,
    left=6pt,right=6pt,top=5pt,bottom=5pt,
    before skip=4pt,after skip=0pt,
    fontupper=\small,
    coltext=black
}

\usepackage{amsthm}
\theoremstyle{definition}

\theoremstyle{plain}

\tcbuselibrary{breakable}
\title{Do Language Models Rethink After Retraction? \\
Characterizing Belief Revision Under Support Withdrawal}

\definecolor{mutedrose}{HTML}{526D87}
\definecolor{ineeji}{HTML}{A65A62}

\newcommand{\affmark}{%
  \textsuperscript{\textcolor{mutedrose}{$\clubsuit$}}%
}
\newcommand{\ineejimark}{%
  \textsuperscript{\textcolor{ineeji}{$\spadesuit$}}%
}

\author{
  \textbf{Bumjin Park\affmark},
  \textbf{Jiyu Lim},
  \textbf{Jaesik Choi\affmark\ineejimark}
\\[4pt]
  \affmark\ KAIST AI \qquad
  \ineejimark\ INEEJI
}
\begin{document}
\maketitle
\begin{abstract}
Large language models (LLMs) must revise their judgments
when evidence changes, yet withdrawing support does not
always require withdrawing the conclusion.
Alternative evidence may still justify it, requiring LLMs
to reconsider what remains.
We evaluate this ability across a controlled selection task
and three reasoning benchmarks, with cases requiring answer
preservation or change.
Across ten open-weight LLMs, support withdrawal frequently
disrupts correct judgments.
Even on the simple selection task, accuracy falls from
100\% to 46.7\% when the answer should remain unchanged,
compared with 93.1\% when it should change.
Explicit reassessment instructions offer limited improvement,
while targeted fine-tuning substantially improves
post-retraction accuracy within each domain, with modest
and variable gains across domains.
Our findings expose a gap between reacting to support
withdrawal and preserving justified conclusions, showing
that targeted training can substantially reduce this gap.
\end{abstract}

% ============================
% ============================
% ============================
% ============================
% ============================

\begin{figure*}[ht!]
    \centering
    \includegraphics[width=1.0\linewidth]{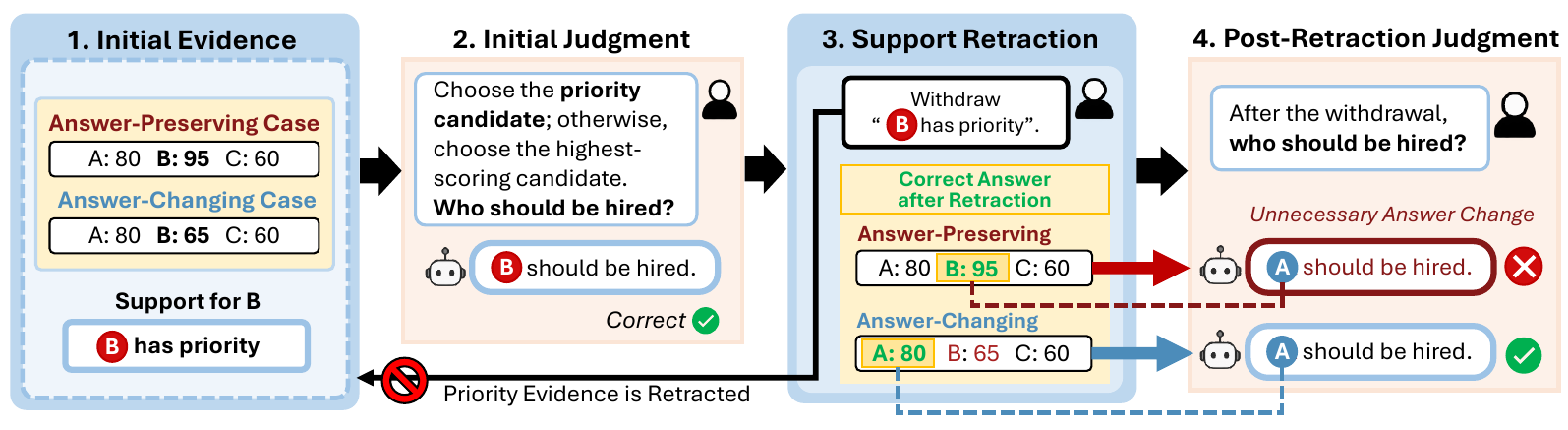}
\caption{Illustration of support retraction with answer preservation and change.
After \texttt{B}'s priority is withdrawn, the highest score determines the answer: \texttt{B} (95) in the \textcolor{preservecolor}{answer-preserving case} and \texttt{A} (80) in the \textcolor{changecolor}{answer-changing case}. Language models often retract the correct answer along with its support, despite alternative evidence.}
    \label{fig:fig_main_intro}
\end{figure*}

% ======================================

\section{Introduction}

Large language models (LLMs) must maintain beliefs consistent with information provided in context \cite{NEURIPS2020_1457c0d6, wilie2024belief, zhou2023context}.
As conversations progress, this requires retaining, revising, or retracting earlier conclusions \cite{xu2024knowledge, wu2024reasoning}.
Such belief management is essential for conversational consistency, appropriate responses to user corrections, and reliable reasoning \cite{jiayang2024econ, marjanovic2024dynamicqa}.
Prior studies have shown that LLMs exhibit instability in belief revision when previously accepted information is withdrawn \cite{deshpande2025multichallenge, yang2026when, helwe2025reviseqa}.

To better understand this instability, we ask:
\emph{How do LLMs revise their beliefs when supporting evidence is retracted?}
Withdrawing support does not always require changing the conclusion, since alternative evidence may still support it \cite{doyle1979truth}.
We therefore distinguish an \textcolor{preservecolor}{answer-preserving case}, where the original answer remains justified, from an \textcolor{changecolor}{answer-changing case}, where the remaining evidence requires a different answer.

For example, a candidate may initially be selected by priority.
After that priority is withdrawn, the candidate remains the correct choice if their score is highest; otherwise, another candidate should be selected.
Appropriate belief revision therefore requires reconsidering the remaining evidence, rather than automatically retracting the earlier conclusion, as illustrated in Figure~\ref{fig:fig_main_intro}.

We evaluate this ability through a two-stage interaction: LLMs first answer a question, then receive a support retraction and answer again.
Across ten open-weight LLMs, Direct accuracy on the synthetic Priority task averages 46.7\% in \textcolor{preservecolor}{answer-preserving cases}, compared with 93.1\% in \textcolor{changecolor}{answer-changing cases}.
This instability also appears in \textcolor{preservecolor}{answer-preserving cases} across other reasoning domains (FOLIO and ProofWriter).
These results reveal a limited ability to reconsider the remaining evidence and reliably maintain conclusions after their original support is withdrawn.
Chain-of-thought \cite{NEURIPS2022_9d560961} improves performance but leaves substantial errors, while explicit reassessment instructions yield modest and inconsistent gains.

We first examine whether explicit instructions help LLMs
reconsider the remaining evidence.
The instructions ask them to assess surviving support,
verify the previous answer, or judge independently without
assuming that the answer should change.
Compared with general accuracy instructions, these targeted
prompts yield modest and inconsistent gains, indicating that
requests to reconsider do not reliably resolve retraction errors.

We then examine reported information needs and response
sensitivity through evidence-revisit questions and
attention-blocking interventions.
Both vary across models and datasets, without establishing
whether the remaining evidence is reassessed.
Targeted fine-tuning yields substantially larger in-domain
gains, but transfer from SimplePriority is limited and
model-dependent.
This contrast highlights the difficulty of improving
post-retraction judgments beyond the training domain
and identifying the processes underlying correct answers.

\section{Related Work}

\paragraph{Belief Revision in AI.}
Classical AI studies how to accommodate changes in knowledge while preserving justified beliefs.
AGM theory distinguishes revision from contraction, which removes a belief without asserting its negation \citep{alchourron1985logic}.
Truth maintenance systems track belief justifications as assumptions change \citep{doyle1979truth}.
Circumscription, default logic, and defeasible logic formalize reasoning with incomplete information and exceptions \citep{mccarthy1980circumscription,reiter1980logic,nute2001defeasible}, while argumentation frameworks address argument acceptability under conflict \citep{dung1995acceptability}.
Continual learning, model editing, and unlearning address related changes in parameterized knowledge.
We examine in-context belief maintenance when supporting evidence is withdrawn.

\paragraph{Belief Revision and Representations in LLMs.}
Research on belief revision examines how LLM judgments respond to changing evidence.
Defeasible NLI studies how additional evidence weakens or strengthens an inference \citep{rudinger2020thinking}, while Belief-R evaluates conclusion revision and maintenance under new premises \citep{wilie2024belief}.
DEFREASING extends this question to property inheritance \citep{allaway-mckeown-2025-evaluating}, and STALE examines whether agents recognize and act on memories invalidated by later observations \citep{chao2026stalellmagentsknow}.
Beyond generated answers, studies of internal belief representations show that statement truth is decodable from hidden states \citep{azaria2023the,marks2024geometrytruthemergentlinear,burns2022discovering} and that internal correctness signals can differ from model responses \citep{orgad2025llms, li2023inference}.
Related work connects internal representations to self-retraction and answer changes under user pressure \citep{yang2026when,genadi2026sycophancy}.
We examine support withdrawal across
\textcolor{preservecolor}{answer-preserving} and
\textcolor{changecolor}{answer-changing} cases through
behavioral evaluation, evidence-revisit self-reports,
and attention-blocking interventions.
Appendix~\ref{app:comparison_prior_work} provides
a detailed comparison.
\section{Task Definition}
\label{sec:task}

\subsection{Revision under Support Retraction}
\label{sec:revision_task}

Let $E_0$ be the initial evidence, consisting of facts
and task rules, and let $R\subseteq E_0$ be the items
withdrawn from that evidence.
The remaining evidence is $E_1=E_0\setminus R$.
For a query $q$, we distinguish the gold answers
$y_0,y_1$ from the LLM's generated responses
$a_{\mathrm{prior}},a_{\mathrm{post}}$ before and
after retraction, respectively.

Let $\operatorname{Cl}_D(E)$ denote the closure of
evidence $E$ under the inference semantics of task $D$.
An answer function $f_D$ determines the gold answers:
\begin{equation}
\begin{aligned}
    y_0 &= f_D(\operatorname{Cl}_D(E_0),q),\\
    y_1 &= f_D(\operatorname{Cl}_D(E_1),q).
\end{aligned}
\label{eq:logical_retraction}
\end{equation}
An \textcolor{preservecolor}{answer-preserving case}
has $y_1=y_0$, whereas an
\textcolor{changecolor}{answer-changing case}
has $y_1\neq y_0$.
These conditions concern the gold answers,
regardless of what the LLM initially generates.

Retraction excludes information from the active premises.
Withdrawing $Q$ removes its assertion without adding
$\neg Q$, whereas adding $\neg Q$ while retaining $Q$
creates a contradiction (both are asserted).
The answer is then inferred from $E_1$,
so a conclusion remains justified only if
the remaining evidence supports it.

We evaluate LLMs in two stages:
\begin{equation}
    E_0,\;q_1
    \;\longrightarrow\; a_{\mathrm{prior}}
    \;;\;
    \rho(R),\;q_2
    \;\longrightarrow\; a_{\mathrm{post}},
    \label{eq:interaction}
\end{equation}
where $\rho(R)$ is a textual instruction withdrawing
the items in $R$.
The LLM first generates $a_{\mathrm{prior}}$ given
$E_0$ and $q_1$.
We then append $\rho(R)$ and $q_2$, retaining
the full preceding context, including the actual
initial response.
For subsequent answer generation, $q_2=q_1=q$.
We score $a_{\mathrm{prior}}$ against $y_0$
and $a_{\mathrm{post}}$ against $y_1$.
We also assess post-retraction judgments by asking
whether a candidate statement is valid under $E_1$.
\subsection{State Reuse and Evidence Access}
\label{sec:state_access}

Logical retraction removes $R$ from $E_0$ and recomputes
the closure of $E_1$ (Equation~\ref{eq:logical_retraction}).
An LLM instead answers from a retained conversation
containing both the original evidence and $a_{\mathrm{prior}}$.
Table~\ref{tab:retraction_requirements} summarizes
the procedural differences.

Three considerations guide our analysis:
\begin{itemize}
    \item \textbf{Prior judgment.}
    The previous query--answer pair may influence
    subsequent judgments through context-induced
    answer biases \citep{zhao2021calibrate},
    despite support withdrawal.

    \item \textbf{Selective evidence exclusion.}
    The LLM must identify withdrawn items from $\rho(R)$
    and exclude them as support while retaining surviving
    facts and rules in the unchanged conversation context.

    \item \textbf{Retraction and recomputation.}
    An inference loses its justification when its support
    is withdrawn, but alternative support in $E_1$
    may justify the same conclusion.
    Neither rejecting nor repeating the prior answer
    alone demonstrates correct reassessment.
\end{itemize}

Decoder-only transformers retain contextualized states
at preceding token positions
\citep{oren-etal-2024-transformers}.
To examine the information sources available to the LLM,
we assume that evidence-derived information may be
represented both at the original evidence positions 
and in states formed during the initial interaction.
Let $\mathcal{I}_{E_0}$ and
$\mathcal{I}_{\mathrm{prior}}$ denote the token positions
of $E_0$ and $(q_1,a_{\mathrm{prior}})$, respectively.
Their corresponding cache states are
\begin{equation}
    H_X =
    \left\{
        \bigl(\mathbf{k}^{(\ell)}_t,
              \mathbf{v}^{(\ell)}_t\bigr)
    \right\}_{\substack{
        t\in\mathcal{I}_X\\
        \ell=1,\ldots,L
    }},
    \label{eq:attention_states}
\end{equation}
for $X\in\{E_0,\mathrm{prior}\}$, where
$\mathbf{k}^{(\ell)}_t$ and $\mathbf{v}^{(\ell)}_t$
are the attention key and value at position $t$
in decoder layer $\ell$, and $L$ is the number of layers \cite{vaswani2023attentionneed}.

We refer to $H_{E_0}$ and $H_{\mathrm{prior}}$ as the
\emph{initial evidence state} and \emph{prior interaction state}.
Because the initial interaction follows $E_0$,
$H_{\mathrm{prior}}$ may encode evidence-derived information;
identical initial answers need not imply identical states.

We distinguish three access configurations:
\begin{equation}
\begin{aligned}
    (H_{\mathrm{prior}},\rho(R),q)
        &\longrightarrow a_{\mathrm{post}}
        && \text{\textsc{Prior}},\\
    (H_{E_0},\rho(R),q)
        &\longrightarrow a_{\mathrm{post}}
        && \text{\textsc{Evidence}},\\
    (H_{E_0},H_{\mathrm{prior}},\rho(R),q)
        &\longrightarrow a_{\mathrm{post}}
        && \text{\textsc{Both}}.
\end{aligned}
\label{eq:revision_paths}
\end{equation}
Shared instruction and formatting tokens are omitted.
These configurations motivate comparisons of evidence
and prior-interaction access.
Our interventions block attention from selected input
positions to the designated segments, rather than
removing access throughout generation.

\begin{table}[t]
\centering
\small
\setlength{\tabcolsep}{3pt}
\renewcommand{\arraystretch}{1.2}
\begin{tabularx}{\columnwidth}{
    @{}lc
    >{\footnotesize\raggedright\arraybackslash}X
    >{\footnotesize\raggedright\arraybackslash}X@{}
}
\toprule
Aspect & & Symbolic procedure & LLM processing \\
\midrule
Prior
& $a_{\mathrm{prior}}$
& Prior answer not required
& Prior response may influence judgment \\
\addlinespace[2pt]

Target
& $R$
& Premises explicitly specified
& Instruction matched to evidence \\
\addlinespace[2pt]

Removal
& $\rho(R)$
& Withdrawn premises excluded
& Evidence use suppressed despite retained tokens \\
\addlinespace[2pt]

Inference
& $(E_1,q_2)$
& Closure explicitly recomputed
& Reassessment not explicitly guaranteed \\
\bottomrule
\end{tabularx}
\caption{Procedural differences in support retraction.}
\label{tab:retraction_requirements}
\end{table}

% ================================================
% ================================================
% ================================================
% ================================================
% ================================================

\begin{figure*}[t!]
    \centering
    \includegraphics[width=\linewidth]{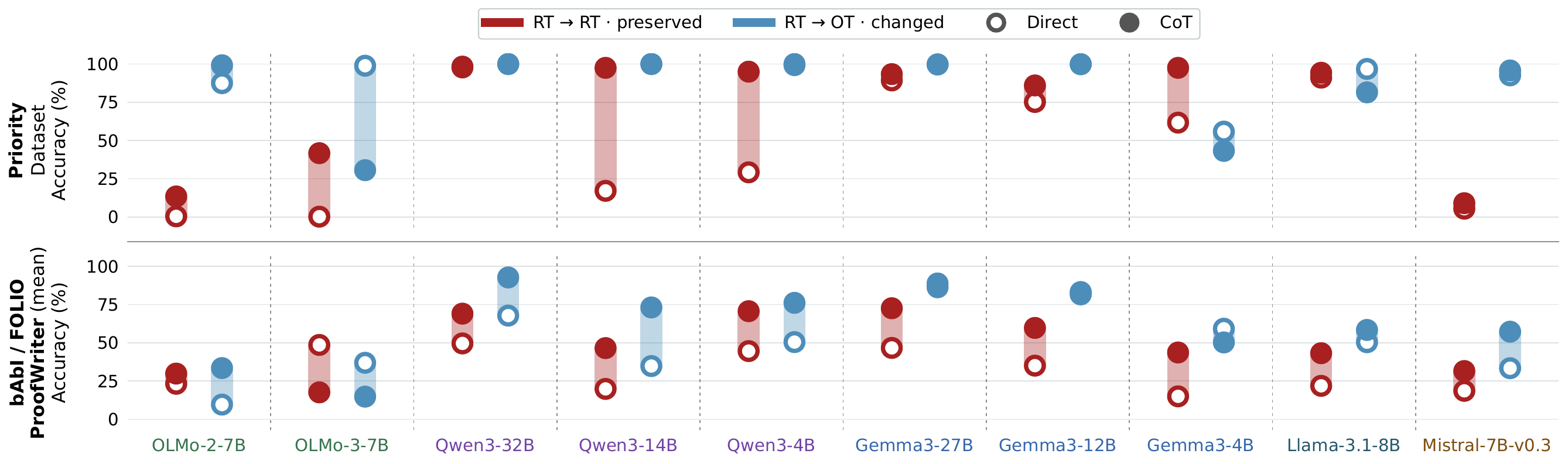}
\caption{Post-retraction accuracy by LLM on
SimplePriority (top) and averaged equally across
bAbI, FOLIO, and ProofWriter (bottom).
Red and blue denote \textcolor{preservecolor}{answer-preserving
(\textsc{rt}$\to$\textsc{rt})} and \textcolor{changecolor}{answer-changing
(\textsc{rt}$\to$\textsc{ot})} cases.
Hollow and filled markers indicate Direct and CoT,
respectively.}
    \label{fig:fig_main_model_wise_retraction}
\end{figure*}

\begin{figure}[b!]
\centering
\begin{priorityretraction}
\textbf{Stage 1: Initial Question}\par
\textbf{User:}
Choose the priority candidate; otherwise, choose
the highest-scoring candidate.\par
\textbf{Scores:} A: 7; B: \textbf{9}; C: 5.\par
\textbf{Priority:}
\textcolor{preservecolor}{B has priority.}\par
Who should be chosen? Answer only A, B, or C.\par
\smallskip
\textbf{Assistant:} \textit{[Initial response]}
\end{priorityretraction}

\vspace{4pt}

\begin{priorityinitial}
\textbf{Stage 2: Support Retraction}\par
\textbf{User:}
\textcolor{preservecolor}{Withdraw ``B has priority''.}\par
Who should be chosen? Answer only A, B, or C.\par
\smallskip
\textbf{Assistant:} \textit{[Post-retraction response]}
\end{priorityinitial}

\caption{An \textcolor{preservecolor}{answer-preserving} Priority example.
B remains correct after its priority is withdrawn
because it has the highest score. Red marks withdrawn support.}
\label{fig:priority-example}
\end{figure}

\section{Experimental Setup}
\label{sec:experimental_setup}

\paragraph{Datasets.}
We introduce SimplePriority, a synthetic candidate-selection
dataset where withdrawing priority requires choosing by score
(Figure~\ref{fig:priority-example}).
We also adapt bAbI for temporal state tracking \cite{weston2015towards_babi},
and FOLIO \cite{han2022folio} and ProofWriter \cite{tafjord2021proofwriter} for logical inference.
Each dataset includes
\textcolor{preservecolor}{answer-preserving} and
\textcolor{changecolor}{answer-changing} retractions.
We use 250/100/500 base cases per dataset for
training/validation/testing.
Each base case yields two retraction conditions, giving
500/200/1,000 examples for training/validation/test.
SimplePriority additionally includes two inventory variants,
doubling these counts.
Unless otherwise stated, all reported results are on the test split.
Construction and selection details are provided
in Appendix~\ref{app:dataset}.

\paragraph{Models and Generation.}
We evaluate ten instruction-tuned LLMs:
Qwen3-4B/14B/32B \cite{yang2025qwen3technicalreport}, Gemma-3-4B/12B/27B \cite{gemmateam2025gemma3technicalreport},
Llama-3.1-8B \cite{grattafiori2024llama3herdmodels}, Mistral-7B-v0.3 \cite{jiang2023mistral7b},
OLMo-2-1124-7B, and OLMo-3-7B \cite{groeneveld-etal-2024-olmo}.
Instruction and fine-tuning experiments use five LLMs:
Qwen3-4B/14B , Gemma-3-4B/12B ,
and Llama-3.1-8B.

For post-retraction responses, we compare Direct
generation with short chain-of-thought (CoT) \cite{NEURIPS2022_9d560961},
which produces a brief reason and a final answer in JSON.
Both modes share the actual initial response
generated under Direct, and only the final answer is scored.
The appendix provides task prompts and instructions;
Appendix~\ref{app:experimental-settings} details
the experimental settings.
% ==============================================================================
\subsection{Baseline Retraction Performance}
\label{sec:support-retraction}

\paragraph{Performance after Support Retraction.}
Figure~\ref{fig:fig_main_model_wise_retraction}
shows post-retraction accuracy for individual LLMs
under Direct and short CoT generation.
\textsc{rt} (\emph{retraction target}) denotes the initial
gold answer, and \textsc{ot} (\emph{other target}) denotes
a different post-retraction gold answer.
The transitions describe gold answers, not generated responses.

\textbf{LLMs often struggle to preserve answers
after support retraction.}
On SimplePriority, several LLMs achieve high
\textcolor{changecolor}{answer-changing} accuracy
but lower
\textcolor{preservecolor}{answer-preserving} accuracy
under Direct generation.
CoT narrows this gap for many LLMs, although its
effects vary across models and conditions.
The remaining datasets also show substantial variation,
indicating that preservation and change are not
uniformly handled across tasks.

Table~\ref{tab:run1-datasets-1} summarizes accuracy
before and after retraction, averaged equally over ten LLMs.
On SimplePriority, Direct accuracy is 46.7\%
in \textcolor{preservecolor}{preserving cases} versus 93.1\% in \textcolor{changecolor}{changing cases}.
FOLIO and ProofWriter show similar preservation deficits,
whereas bAbI has lower changing accuracy under Direct.
Across datasets, \textcolor{preservecolor}{preserving accuracy} falls by
30.3 percentage points under Direct and 10.7 under CoT
relative to the initial judgment.

\textbf{CoT improves overall post-retraction accuracy
on every dataset, but does not ensure reliable preservation.}
Compared with Direct, preserving accuracy rises
from 46.7\% to 74.5\% on SimplePriority ($+27.8$ pp),
64.5\% to 79.8\% on bAbI ($+15.3$ pp),
30.9\% to 53.5\% on FOLIO ($+22.6$ pp), and
28.6\% to 41.2\% on ProofWriter ($+12.6$ pp).
Yet LLMs may conflate support retraction with conclusion
retraction, discarding the answer despite surviving support.
Generated reasoning can still overlook that support
or misinterpret its withdrawal.
Appendix~\ref{app:cot_generation_examples} provides
CoT examples and analyses for each dataset.

% ==============================================================================

% ==============================================================================

\begin{table}[t]
\centering
\small
\setlength{\tabcolsep}{0pt}\setlength{\arrayrulewidth}{0.3pt}
\renewcommand{\arraystretch}{1.20}
\setlength{\aboverulesep}{0pt}\setlength{\belowrulesep}{0pt}
\resizebox{\columnwidth}{!}{%
\begin{tabular}{@{}l@{\hspace{2pt}}|@{\hspace{2pt}}c@{\hspace{2pt}}|@{\hspace{2pt}}l@{\hspace{2pt}}|@{\hspace{2pt}}c@{\hspace{2pt}}|@{\hspace{2pt}}c@{\hspace{2pt}}|@{\hspace{2pt}}c@{}}
\toprule
\multirow{2}{*}{Dataset} & \multirow{2}{*}{\shortstack{Initial\\Answer\\(Label: \textsc{rt})}} & \multicolumn{4}{c}{\shortstack{\rule{0pt}{2.8ex}Follow-up Evaluation\\(After Evidence Retraction)\rule[-1.0ex]{0pt}{1ex}}} \\
\cline{3-6}
 & & Task & \shortstack{\rule{0pt}{2.8ex}Overall\\(all cases)} & \shortstack{\rule{0pt}{2.8ex}\textsc{rt} $\to$ \textsc{rt}\\(label preserved)} & \shortstack{\rule{0pt}{2.8ex}\textsc{rt} $\to$ \textsc{ot}\\(label changed)} \\
\midrule
\multirow{2}{*}{Priority} & \multirow{2}{*}{100.0} & Direct & \makebox[2.4em][r]{69.9}\hspace{0.2em}\makebox[5.5em][l]{\colorbox{red!18.5}{\makebox[4.8em][c]{\strut\textcolor{black}{($\Delta -30.0$)}}}} & \makebox[2.4em][r]{\textbf{46.7}}\hspace{0.2em}\makebox[5.5em][l]{\colorbox{red!25.0}{\makebox[4.8em][c]{\strut\textcolor{black}{($\Delta -53.3$)}}}} & \makebox[2.4em][r]{93.1}\hspace{0.2em}\makebox[5.5em][l]{\colorbox{red!11.9}{\makebox[4.8em][c]{\strut\textcolor{black}{($\Delta -6.8$)}}}} \\
 &  & CoT & \makebox[2.4em][r]{83.2}\hspace{0.2em}\makebox[5.5em][l]{\colorbox{red!14.7}{\makebox[4.8em][c]{\strut\textcolor{black}{($\Delta -16.8$)}}}} & \makebox[2.4em][r]{\textbf{74.5}}\hspace{0.2em}\makebox[5.5em][l]{\colorbox{red!17.2}{\makebox[4.8em][c]{\strut\textcolor{black}{($\Delta -25.5$)}}}} & \makebox[2.4em][r]{91.8}\hspace{0.2em}\makebox[5.5em][l]{\colorbox{red!12.3}{\makebox[4.8em][c]{\strut\textcolor{black}{($\Delta -8.1$)}}}} \\
\specialrule{0.3pt}{0pt}{0pt}
\multirow{2}{*}{bAbI} & \multirow{2}{*}{76.1} & Direct & \makebox[2.4em][r]{60.4}\hspace{0.2em}\makebox[5.5em][l]{\colorbox{red!14.4}{\makebox[4.8em][c]{\strut\textcolor{black}{($\Delta -15.6$)}}}} & \makebox[2.4em][r]{64.5}\hspace{0.2em}\makebox[5.5em][l]{\colorbox{red!13.3}{\makebox[4.8em][c]{\strut\textcolor{black}{($\Delta -11.6$)}}}} & \makebox[2.4em][r]{\textbf{56.4}}\hspace{0.2em}\makebox[5.5em][l]{\colorbox{red!15.5}{\makebox[4.8em][c]{\strut\textcolor{black}{($\Delta -19.7$)}}}} \\
 &  & CoT & \makebox[2.4em][r]{80.0}\hspace{0.2em}\makebox[5.5em][l]{\colorbox{green!13.2}{\makebox[4.8em][c]{\strut\textcolor{black}{($\Delta +3.9$)}}}} & \makebox[2.4em][r]{\textbf{79.8}}\hspace{0.2em}\makebox[5.5em][l]{\colorbox{green!13.0}{\makebox[4.8em][c]{\strut\textcolor{black}{($\Delta +3.7$)}}}} & \makebox[2.4em][r]{80.1}\hspace{0.2em}\makebox[5.5em][l]{\colorbox{green!13.4}{\makebox[4.8em][c]{\strut\textcolor{black}{($\Delta +4.1$)}}}} \\
\specialrule{0.3pt}{0pt}{0pt}
\multirow{2}{*}{FOLIO} & \multirow{2}{*}{63.1} & Direct & \makebox[2.4em][r]{56.1}\hspace{0.2em}\makebox[5.5em][l]{\colorbox{red!12.0}{\makebox[4.8em][c]{\strut\textcolor{black}{($\Delta -7.0$)}}}} & \makebox[2.4em][r]{\textbf{30.9}}\hspace{0.2em}\makebox[5.5em][l]{\colorbox{red!19.1}{\makebox[4.8em][c]{\strut\textcolor{black}{($\Delta -32.3$)}}}} & \makebox[2.4em][r]{81.3}\hspace{0.2em}\makebox[5.5em][l]{\colorbox{green!25.0}{\makebox[4.8em][c]{\strut\textcolor{black}{($\Delta +18.2$)}}}} \\
 &  & CoT & \makebox[2.4em][r]{67.2}\hspace{0.2em}\makebox[5.5em][l]{\colorbox{green!13.4}{\makebox[4.8em][c]{\strut\textcolor{black}{($\Delta +4.1$)}}}} & \makebox[2.4em][r]{\textbf{53.5}}\hspace{0.2em}\makebox[5.5em][l]{\colorbox{red!12.7}{\makebox[4.8em][c]{\strut\textcolor{black}{($\Delta -9.6$)}}}} & \makebox[2.4em][r]{80.9}\hspace{0.2em}\makebox[5.5em][l]{\colorbox{green!24.6}{\makebox[4.8em][c]{\strut\textcolor{black}{($\Delta +17.7$)}}}} \\
\specialrule{0.3pt}{0pt}{0pt}
\multirow{2}{*}{ProofWriter} & \multirow{2}{*}{52.7} & Direct & \makebox[2.4em][r]{35.4}\hspace{0.2em}\makebox[5.5em][l]{\colorbox{red!14.9}{\makebox[4.8em][c]{\strut\textcolor{black}{($\Delta -17.3$)}}}} & \makebox[2.4em][r]{\textbf{28.6}}\hspace{0.2em}\makebox[5.5em][l]{\colorbox{red!16.8}{\makebox[4.8em][c]{\strut\textcolor{black}{($\Delta -24.2$)}}}} & \makebox[2.4em][r]{42.3}\hspace{0.2em}\makebox[5.5em][l]{\colorbox{red!12.9}{\makebox[4.8em][c]{\strut\textcolor{black}{($\Delta -10.4$)}}}} \\
 &  & CoT & \makebox[2.4em][r]{49.8}\hspace{0.2em}\makebox[5.5em][l]{\colorbox{red!10.8}{\makebox[4.8em][c]{\strut\textcolor{black}{($\Delta -2.9$)}}}} & \makebox[2.4em][r]{\textbf{41.2}}\hspace{0.2em}\makebox[5.5em][l]{\colorbox{red!13.2}{\makebox[4.8em][c]{\strut\textcolor{black}{($\Delta -11.5$)}}}} & \makebox[2.4em][r]{58.4}\hspace{0.2em}\makebox[5.5em][l]{\colorbox{green!14.7}{\makebox[4.8em][c]{\strut\textcolor{black}{($\Delta +5.7$)}}}} \\
\specialrule{0.9pt}{0pt}{0pt}
\multirow{2}{*}{\textbf{Dataset mean}} &  & Direct &  & \makebox[2.4em][r]{}\hspace{0.2em}\makebox[5.5em][l]{\colorbox{red!18.5}{\makebox[4.8em][c]{\strut\textcolor{black}{($\Delta -30.3$)}}}} & \makebox[2.4em][r]{}\hspace{0.2em}\makebox[5.5em][l]{\colorbox{red!11.3}{\makebox[4.8em][c]{\strut\textcolor{black}{($\Delta -4.7$)}}}} \\
 &  & CoT &  & \makebox[2.4em][r]{}\hspace{0.2em}\makebox[5.5em][l]{\colorbox{red!13.0}{\makebox[4.8em][c]{\strut\textcolor{black}{($\Delta -10.7$)}}}} & \makebox[2.4em][r]{}\hspace{0.2em}\makebox[5.5em][l]{\colorbox{green!14.0}{\makebox[4.8em][c]{\strut\textcolor{black}{($\Delta +4.9$)}}}} \\
\bottomrule
\end{tabular}}
\captionsetup{justification=raggedright,singlelinecheck=false}
\caption{Initial and post-retraction accuracy (\%),
averaged over ten LLMs sharing the same initial Direct response.
$\Delta$: change from initial accuracy (pp);
bold: lower condition accuracy;
red/green: decrease/increase.
Bottom rows average changes across datasets.}
\label{tab:run1-datasets-1}
\end{table}

% =======================================================================

\paragraph{Dependence on Evidence and Prior Responses.}
Table~\ref{tab:dataset-deltas} reports Direct accuracy
under attention blocking, averaged over LLMs.
$E$ and $A_1$ denote evidence and prior-answer token
segments, distinct from $H_{E_0}$ and $H_{\mathrm{prior}}$.
Attention from the final input token to each segment
is blocked at every layer; generated queries remain unblocked.
Details appear in Appendix~\ref{app:experimental-settings}.

\textbf{The effects vary across segments and datasets.}
On SimplePriority, blocking $A_1$ while retaining $E$
raises answer-preserving accuracy by 0.9 pp,
whereas blocking $E$ while retaining $A_1$
is associated with a 12.3 pp decline.
The other datasets show no consistent contrast between
the two segments.
Per-model results appear in
Appendix~\ref{tab:appendix-run3-datasets-by-model}.

\paragraph{Self-Reported Need to Revisit Evidence.}
\label{sec:evidence-verbalization}

\begin{figure}[b!]
    \centering
    \includegraphics[width=\linewidth]{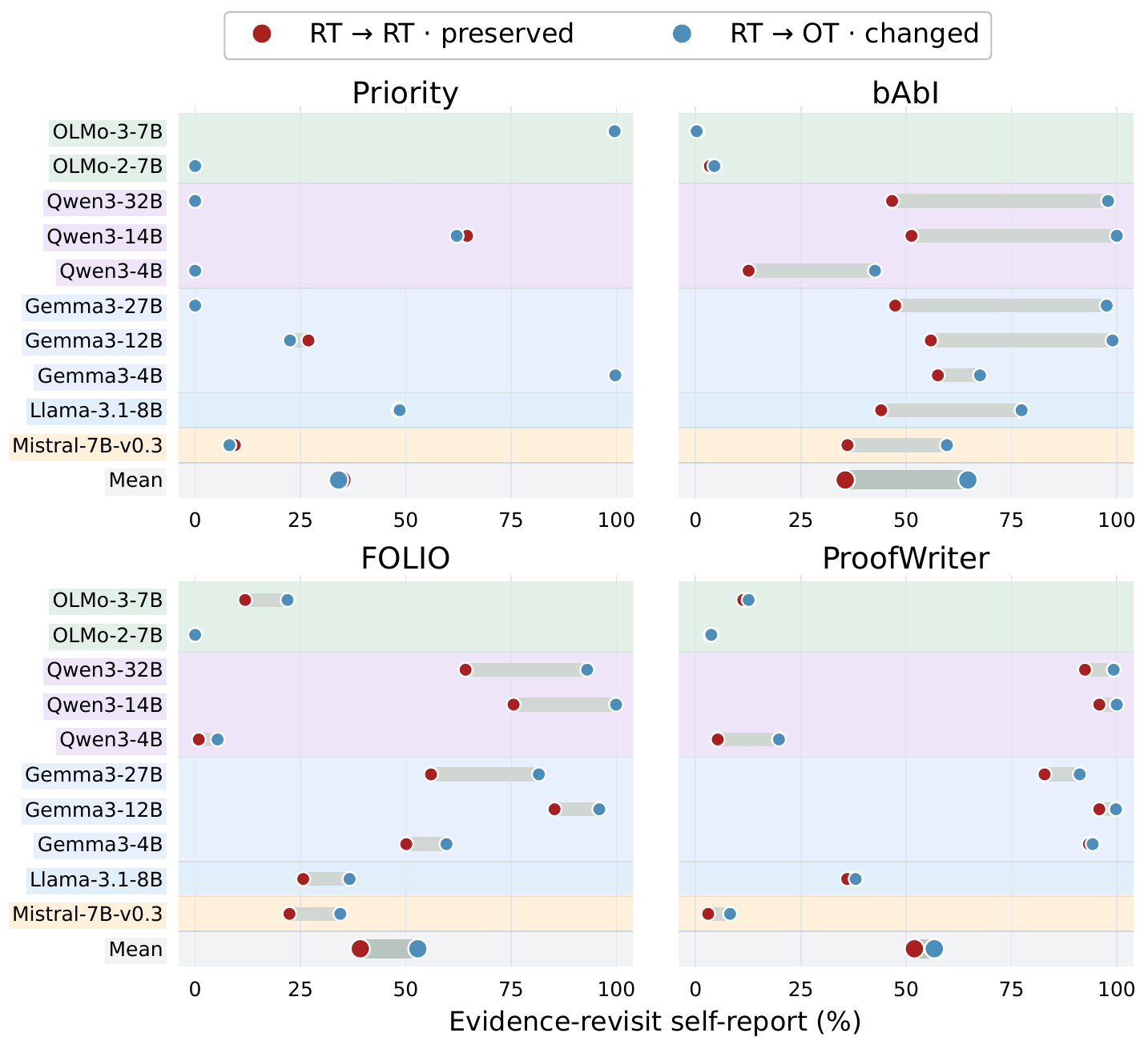}
\caption{Self-reported evidence-revisit rates (\%)
by LLM and gold-answer condition.
Mean averages all evaluated LLMs equally across
all datasets and conditions.
Higher values indicate a greater reported need
to revisit the original evidence.}
    \label{fig:fig_main_verbalization}
\end{figure}

Following prior work eliciting whether LLMs require
additional information \citep{wang2023self,ni2024llms},
we assess their reported need to revisit the evidence
after receiving a support-withdrawal instruction.
Figure~\ref{fig:fig_main_verbalization} reports revisiting rates
under three question formulations.
Although rates vary, answer-preserving cases receive
consistently lower rates than \textcolor{changecolor}{answer-changing cases} on bAbI,
FOLIO, and ProofWriter; the two conditions are similar on
SimplePriority (Appendix~\ref{app:verbalization}).

\textbf{LLMs report less need to revisit evidence in
\textcolor{preservecolor}{answer-preserving cases} than in
\textcolor{changecolor}{answer-changing cases}.}
Although verbalization does not fully reveal the model's
belief state, this pattern suggests that retained information
is more often perceived as sufficient in preserving cases.
Related CoT errors illustrate how an incorrect interpretation
of withdrawal can shape subsequent reasoning:
after \texttt{S}'s priority is withdrawn, some responses appear to
exclude \texttt{S} and select \texttt{X} over \texttt{U},
rather than compare all three candidates
(Table~\ref{tab:apdx-run1-example-priority} in
Appendix~\ref{app:cot_generation_examples}).
These examples do not establish a link to self-reports,
but show how reconsideration can proceed without correctly
assessing the surviving evidence.
% ==============================================================================

% =======================================

\begin{table}[t]
\centering\small
\setlength{\tabcolsep}{4pt}
\renewcommand{\arraystretch}{1.18}
\resizebox{\columnwidth}{!}{%
\begin{tabular}{lccrrrr}
\toprule
Transition & $E$ & $A_1$ & Priority & FOLIO & ProofWriter & bAbI \\
\midrule
\rowcolor{gray!8}
\multirow{4}{*}{RT $\to$ RT} & \textcolor{green!40!black}{$\checkmark$} & \textcolor{green!40!black}{$\checkmark$} & \textbf{51.3} & \textbf{30.8} & \textbf{28.8} & \textbf{69.4} \\
 & \textcolor{green!40!black}{$\checkmark$} & \textcolor{red!70!black}{$\times$} & \cellcolor{green!6.8}+0.9 & 0.0 & \cellcolor{red!13.5}-4.3 & \cellcolor{red!23.0}-9.0 \\
 & \textcolor{red!70!black}{$\times$} & \textcolor{green!40!black}{$\checkmark$} & \cellcolor{red!29.6}-12.3 & \cellcolor{green!5.4}+0.2 & \cellcolor{red!8.7}-1.9 & \cellcolor{red!19.8}-7.4 \\
 & \textcolor{red!70!black}{$\times$} & \textcolor{red!70!black}{$\times$} & \cellcolor{red!21.4}-8.2 & \cellcolor{red!5.4}-0.2 & \cellcolor{red!10.1}-2.5 & \cellcolor{red!45.0}-39.3 \\
\midrule
\rowcolor{gray!8}
\multirow{4}{*}{RT $\to$ OT} & \textcolor{green!40!black}{$\checkmark$} & \textcolor{green!40!black}{$\checkmark$} & \textbf{93.2} & \textbf{82.6} & \textbf{42.6} & \textbf{60.4} \\
 & \textcolor{green!40!black}{$\checkmark$} & \textcolor{red!70!black}{$\times$} & \cellcolor{red!7.3}-1.1 & \cellcolor{red!7.8}-1.4 & \cellcolor{red!7.2}-1.1 & \cellcolor{red!37.7}-16.4 \\
 & \textcolor{red!70!black}{$\times$} & \textcolor{green!40!black}{$\checkmark$} & \cellcolor{red!5.7}-0.3 & 0.0 & \cellcolor{green!8.6}+1.8 & \cellcolor{red!45.0}-25.1 \\
 & \textcolor{red!70!black}{$\times$} & \textcolor{red!70!black}{$\times$} & \cellcolor{red!10.0}-2.5 & \cellcolor{red!7.2}-1.1 & 0.0 & \cellcolor{red!45.0}-37.4 \\
\bottomrule
\end{tabular}}
\caption{Direct accuracy under access ablation,
averaged over LLMs.
Bold: full-access accuracy (\%); other rows: changes (pp).
$\checkmark$/$\times$: access allowed/blocked to
initial evidence ($E_0$) and response ($a_{\mathrm{prior}}$).
Independent histories and backends limit causal interpretation.}
% \caption{Change in Direct follow-up accuracy ($\Delta$, pp) from the corresponding Run1 baseline with both E and $A_1$ access allowed. The first row in each transition group shows baseline accuracy (\%) in bold on gray; subsequent rows show changes (pp). Each dataset is shown separately, averaging equally over 6 common completed models. Negative values indicate a decrease. $\checkmark$: access allowed; $\times$: access blocked. Color intensity uses the same 0--20 pp scale across datasets, saturated at 20 pp. For Priority, identical uppercase-letter repetitions are collapsed before scoring in both runs. Run1 and Run3 use independent initial histories and different backends; differences are not isolated causal effects of access.}
\label{tab:dataset-deltas}
\end{table}

\begin{figure*}[t!]
    \centering
    \includegraphics[width=\linewidth]{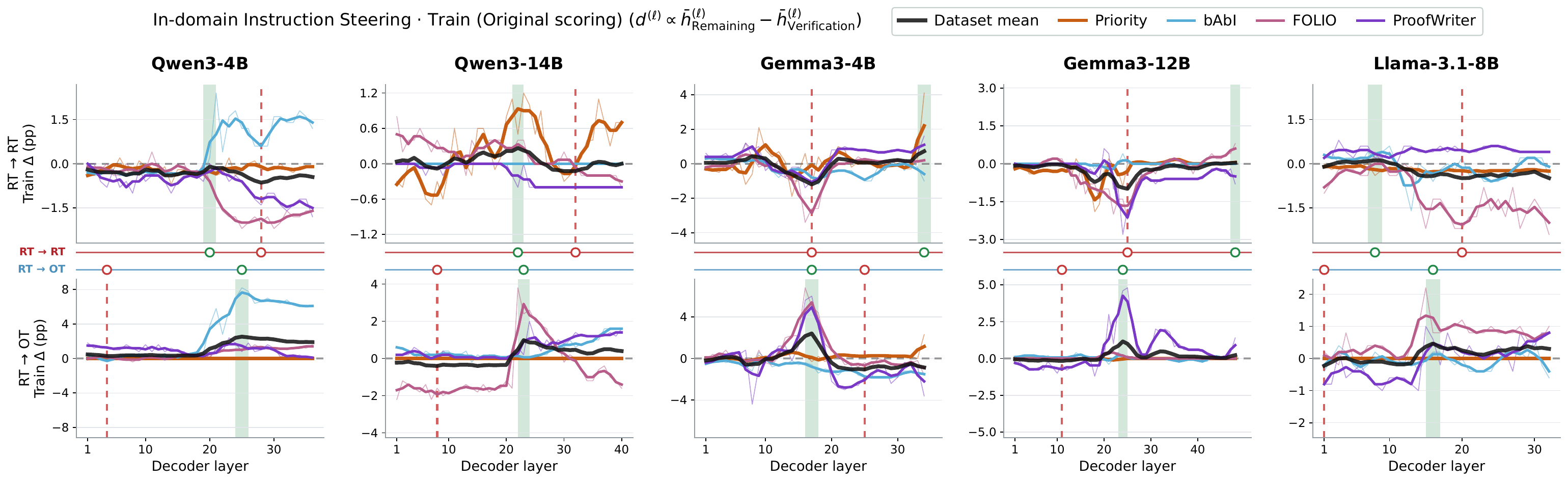}
\caption{Layer-wise changes in training accuracy (pp)
under instruction-derived steering.
Top/bottom: \textcolor{preservecolor}{answer-preserving}/
\textcolor{changecolor}{answer-changing} cases.
Green shading and red dashed lines mark the
dataset-mean maximum and minimum.
Maximizing layers coincide across conditions for
Qwen3-14B but differ for Gemma-3-12B.}
    \label{fig:fig_main_activation_steering_train}
\end{figure*}

% =======================================================
\subsection{Instructional Control under Support Retraction}
\label{sec:instruction}

We test whether instructions improve
post-retraction judgments, following prior work using
prompts to guide belief revision \citep{wilie2024belief}.
In Stage 2, instructions are inserted between $\rho(R)$
and $q_2$, with initial responses and output requirements
fixed across instruction variants within each generation mode.
We use two formulations for each of three targeted types:
remaining-evidence review, answer verification, and
independent judgment, plus accuracy instructions
and one reconsideration prompt
(Appendix~\ref{app:retraction_support_instruction}).
Results are averaged within each type and LLM, then across
five LLMs (Gemma-3-4B/12B, Qwen3-4B/14B, Llama-3.1-8B).

% 벡터 기반 행동 통제 steering
% \cite{kim-etal-2026-incomplete}
% \cite{hendel-etal-2023-context}

\begin{figure}[t!]
    \centering
    \includegraphics[width=\linewidth]{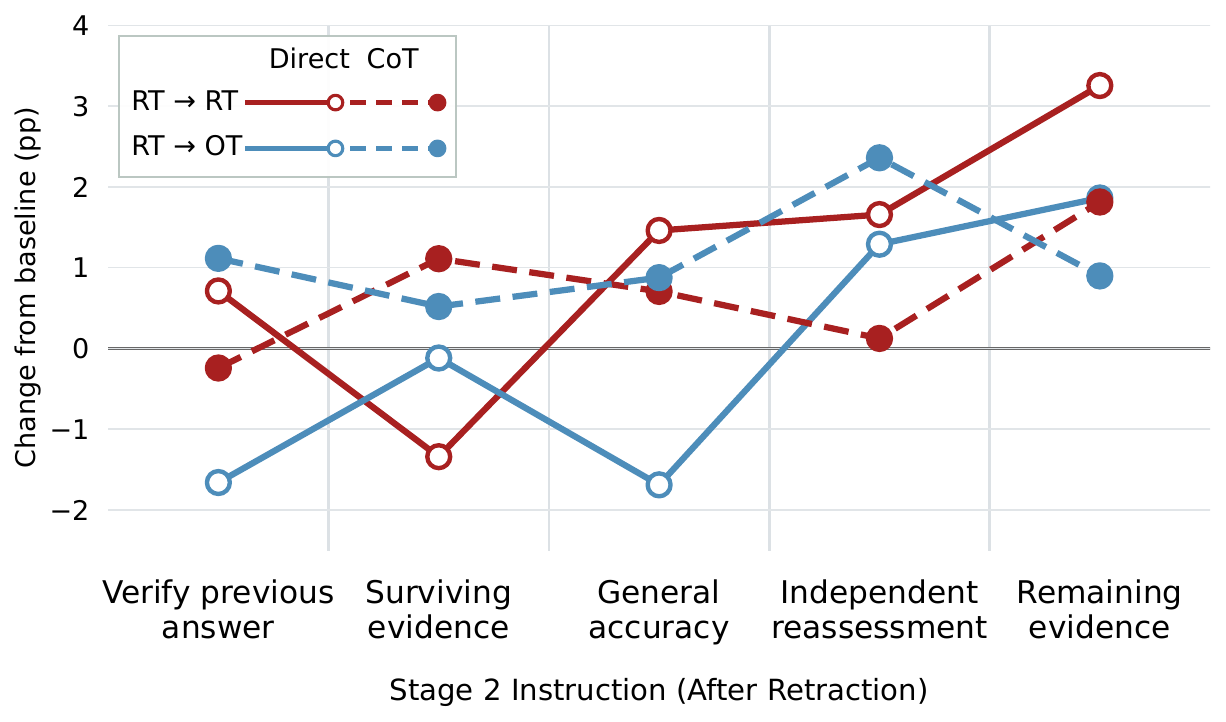}
\caption{Accuracy changes from the no-instruction
baseline (pp), averaged equally over formulations
within each type, five LLMs, and four datasets.
Hollow/solid-line markers: Direct;
filled/dashed-line markers: CoT.}
    \label{fig:fig_main_instruction_performance}
\end{figure}

\textbf{Explicit instructions yield modest and
task-dependent improvements.}
Figure~\ref{fig:fig_main_instruction_performance}
summarizes Direct and CoT effects across four datasets.
Under Direct, remaining-evidence instructions improve
preservation by 3.2 pp on average, compared with
1.4 pp for general accuracy instructions.
The gain varies substantially across datasets,
from +8.6 pp on SimplePriority to -0.8 pp on FOLIO
(Table~\ref{tab:instruction-category-gains}).
Effects on changing cases are also mixed:
remaining-evidence instructions yield +1.9 pp,
whereas verification yields -1.7 pp.
CoT likewise provides small, instruction-dependent gains.

\paragraph{Activation Steering.}
Prior work motivates manipulating task-relevant activations
\citep{hendel-etal-2023-context,kim-etal-2026-incomplete}.
We derive directions from reassessment instructions and
inject them at individual layers during generation
(Appendix~\ref{app:experimental-settings}).
Figure~\ref{fig:fig_main_activation_steering_train} shows
exploratory training-set results.
Some layers improve both
\textcolor{preservecolor}{answer-preserving} and
\textcolor{changecolor}{answer-changing} accuracy,
while others produce opposing effects.
Maximizing layers coincide for Qwen3-14B but differ
for Gemma-3-12B, suggesting model-dependent intervention
points for these directions.
Thus, a layer favorable to one condition need not benefit
the other, complicating joint control of preservation
and change.

\begin{table}[t!]
\centering\small
\setlength{\tabcolsep}{4pt}
\renewcommand{\arraystretch}{1.12}
\resizebox{\columnwidth}{!}{%
\begin{tabular}{llrrrrr}
\toprule
Transition & Instruction & Priority & FOLIO & ProofWriter & bAbI & Mean $\Delta$ \\
\midrule
\rowcolor{gray!8}
RT $\to$ RT & Baseline & \textbf{54.9} & \textbf{20.6} & \textbf{24.2} & \textbf{68.1} & -- \\
 & Remaining & \cellcolor{green!22.1}+8.6 & \cellcolor{red!6.6}-0.8 & \cellcolor{green!9.4}+2.2 & \cellcolor{green!11.0}+3.0 & \cellcolor{green!11.5}+3.2 \\
 & Verification & \cellcolor{green!6.3}+0.7 & \cellcolor{red!7.3}-1.2 & \cellcolor{green!10.0}+2.5 & \cellcolor{green!7.6}+1.3 & \cellcolor{green!6.6}+0.8 \\
 & Independent & \cellcolor{green!23.8}+9.4 & \cellcolor{red!9.0}-2.0 & \cellcolor{green!8.0}+1.5 & \cellcolor{red!8.5}-1.8 & \cellcolor{green!8.6}+1.8 \\
 & General accuracy & \cellcolor{green!19.0}+7.0 & \cellcolor{red!8.5}-1.7 & \cellcolor{red!6.2}-0.6 & \cellcolor{green!6.7}+0.9 & \cellcolor{green!7.8}+1.4 \\
 & Additional & \cellcolor{red!12.7}-3.9 & \cellcolor{red!6.8}-0.9 & \cellcolor{green!6.4}+0.7 & \cellcolor{red!7.8}-1.4 & \cellcolor{red!7.8}-1.4 \\
\midrule
\rowcolor{gray!8}
RT $\to$ OT & Baseline & \textbf{90.5} & \textbf{93.5} & \textbf{48.6} & \textbf{55.0} & -- \\
 & Remaining & \cellcolor{red!7.1}-1.0 & \cellcolor{green!8.5}+1.8 & \cellcolor{red!8.2}-1.6 & \cellcolor{green!21.8}+8.4 & \cellcolor{green!8.8}+1.9 \\
 & Verification & 0.0 & \cellcolor{red!6.3}-0.7 & \cellcolor{red!15.4}-5.2 & \cellcolor{red!6.6}-0.8 & \cellcolor{red!8.3}-1.7 \\
 & Independent & \cellcolor{green!7.6}+1.3 & \cellcolor{green!8.6}+1.8 & \cellcolor{green!7.0}+1.0 & \cellcolor{green!7.0}+1.0 & \cellcolor{green!7.5}+1.3 \\
 & General accuracy & \cellcolor{green!6.4}+0.7 & \cellcolor{red!5.8}-0.4 & \cellcolor{red!9.9}-2.5 & \cellcolor{red!14.4}-4.7 & \cellcolor{red!8.4}-1.7 \\
 & Additional & \cellcolor{green!9.8}+2.4 & \cellcolor{red!5.2}-0.1 & \cellcolor{red!17.2}-6.1 & \cellcolor{green!11.7}+3.4 & \cellcolor{red!5.2}-0.1 \\
\bottomrule
\end{tabular}}
\caption{Direct instruction effects, averaged equally
over formulations within each type and five LLMs.
Bold: baseline accuracy (\%); other rows: changes (pp).
Mean $\Delta$ averages four datasets equally;
green/red indicate gains/losses.}
% \caption{
% Direct test accuracy (\%) without additional instructions (bold) and changes for five Run4 instruction categories (pp). Equal-weight means over 5 models per dataset, using the same complete model cohort for every category. Templates are averaged equally within each model and category (2 Remaining, 2 Verification, 2 Independent, 6 General accuracy, 1 Additional), then models are averaged equally. The last column equally averages the changes across the four datasets. Answer normalization v2; green/red indicate gains/losses on a shared 0--20 pp scale.
% }
\label{tab:instruction-category-gains}
\end{table}

\begin{table*}[t]
\centering\footnotesize
\setlength{\tabcolsep}{3.5pt}
\renewcommand{\arraystretch}{1.05}
\resizebox{\textwidth}{!}{%
\begin{tabular}{lrrrrrrrrrrr}
\toprule
 & \multicolumn{2}{c}{Priority} & \multicolumn{3}{c}{FOLIO} & \multicolumn{3}{c}{ProofWriter} & \multicolumn{3}{c}{bAbI} \\
\cmidrule(lr){2-3}\cmidrule(lr){4-6}\cmidrule(lr){7-9}\cmidrule(lr){10-12}
Model & \multicolumn{1}{c}{Base (\%)} & \multicolumn{1}{c}{Tuned} & \multicolumn{1}{c}{Base (\%)} & \multicolumn{1}{c}{In-domain} & \multicolumn{1}{c}{Transfer} & \multicolumn{1}{c}{Base (\%)} & \multicolumn{1}{c}{In-domain} & \multicolumn{1}{c}{Transfer} & \multicolumn{1}{c}{Base (\%)} & \multicolumn{1}{c}{In-domain} & \multicolumn{1}{c}{Transfer} \\
\midrule
\rowcolor{gray!8}
\multicolumn{12}{l}{\textbf{RT $\to$ RT}} \\
Qwen3-4B & 29.2 & \cellcolor{green!20.0}99.4 (+70.2) & 35.6 & \cellcolor{green!20.0}72.0 (+36.4) & \cellcolor{green!10.0}40.2 (+4.6) & 49.0 & \cellcolor{green!20.0}90.4 (+41.4) & \cellcolor{green!7.4}51.6 (+2.6) & 73.8 & \cellcolor{green!20.0}100.0 (+26.2) & \cellcolor{green!8.7}77.4 (+3.6) \\
Qwen3-14B & 17.1 & \cellcolor{green!20.0}99.8 (+82.7) & 31.6 & \cellcolor{green!20.0}61.8 (+30.2) & \cellcolor{green!10.2}36.4 (+4.8) & 27.0 & \cellcolor{green!20.0}65.8 (+38.8) & \cellcolor{green!4.3}27.2 (+0.2) & 88.8 & \cellcolor{green!18.6}100.0 (+11.2) & \cellcolor{green!5.3}89.8 (+1.0) \\
Gemma3-4B & 61.7 & \cellcolor{green!20.0}99.0 (+37.3) & 7.2 & \cellcolor{green!20.0}29.0 (+21.8) & \cellcolor{green!20.0}30.6 (+23.4) & 6.0 & \cellcolor{green!20.0}76.6 (+70.6) & \cellcolor{green!14.1}13.8 (+7.8) & 54.8 & \cellcolor{green!20.0}100.0 (+45.2) & \cellcolor{green!25.8}71.6 (+16.8) \\
Gemma3-12B & 75.2 & \cellcolor{green!20.0}99.9 (+24.7) & 12.2 & \cellcolor{green!20.0}63.8 (+51.6) & \cellcolor{green!14.4}20.2 (+8.0) & 17.0 & \cellcolor{green!20.0}84.2 (+67.2) & \cellcolor{green!5.8}18.4 (+1.4) & 93.8 & \cellcolor{green!11.5}99.6 (+5.8) & \cellcolor{green!6.9}96.0 (+2.2) \\
Llama-3.1-8B & 91.1 & \cellcolor{green!14.0}98.8 (+7.7) & 16.4 & \cellcolor{green!20.0}57.0 (+40.6) & \cellcolor{red!18.0}5.6 (-10.8) & 22.0 & \cellcolor{green!20.0}70.0 (+48.0) & \cellcolor{red!6.3}20.2 (-1.8) & 29.4 & \cellcolor{green!20.0}98.6 (+69.2) & \cellcolor{green!11.3}35.0 (+5.6) \\
\cmidrule(lr){1-12}
\textbf{Model mean} & -- & \cellcolor{green!20.0}+44.5 & -- & \cellcolor{green!20.0}+36.1 & \cellcolor{green!11.8}+6.0 & -- & \cellcolor{green!20.0}+53.2 & \cellcolor{green!6.7}+2.0 & -- & \cellcolor{green!20.0}+31.5 & \cellcolor{green!11.6}+5.8 \\
\midrule
\rowcolor{gray!8}
\multicolumn{12}{l}{\textbf{RT $\to$ OT}} \\
Qwen3-4B & 100.0 & 100.0 (0.0) & 91.0 & \cellcolor{green!5.3}92.0 (+1.0) & \cellcolor{red!11.0}85.6 (-5.4) & 15.4 & \cellcolor{green!20.0}78.2 (+62.8) & \cellcolor{green!7.9}18.4 (+3.0) & 51.4 & \cellcolor{green!20.0}99.8 (+48.4) & \cellcolor{red!12.8}44.6 (-6.8) \\
Qwen3-14B & 100.0 & 100.0 (0.0) & 99.2 & \cellcolor{red!7.1}96.8 (-2.4) & \cellcolor{red!5.3}98.2 (-1.0) & 68.8 & \cellcolor{green!20.0}94.4 (+25.6) & \cellcolor{green!7.6}71.6 (+2.8) & 56.2 & \cellcolor{green!20.0}99.8 (+43.6) & \cellcolor{green!19.6}68.2 (+12.0) \\
Gemma3-4B & 55.8 & \cellcolor{green!20.0}99.4 (+43.6) & 89.8 & \cellcolor{green!15.4}98.6 (+8.8) & \cellcolor{red!20.1}77.4 (-12.4) & 69.2 & \cellcolor{red!9.2}65.2 (-4.0) & \cellcolor{red!6.3}67.4 (-1.8) & 27.6 & \cellcolor{green!20.0}99.6 (+72.0) & \cellcolor{green!20.0}51.8 (+24.2) \\
Gemma3-12B & 99.9 & \cellcolor{red!4.1}99.8 (-0.1) & 97.0 & \cellcolor{red!4.3}96.8 (-0.2) & \cellcolor{green!7.6}99.8 (+2.8) & 69.4 & \cellcolor{green!21.7}83.0 (+13.6) & \cellcolor{green!7.4}72.0 (+2.6) & 97.0 & \cellcolor{green!7.9}100.0 (+3.0) & \cellcolor{red!5.8}95.6 (-1.4) \\
Llama-3.1-8B & 96.8 & \cellcolor{green!7.2}99.3 (+2.5) & 90.6 & \cellcolor{green!9.5}94.8 (+4.2) & \cellcolor{red!6.6}88.6 (-2.0) & 20.4 & \cellcolor{green!20.0}90.8 (+70.4) & \cellcolor{green!12.1}26.6 (+6.2) & 42.8 & \cellcolor{green!20.0}99.2 (+56.4) & \cellcolor{red!7.4}40.2 (-2.6) \\
\cmidrule(lr){1-12}
\textbf{Model mean} & -- & \cellcolor{green!16.0}+9.2 & -- & \cellcolor{green!7.0}+2.3 & \cellcolor{red!8.7}-3.6 & -- & \cellcolor{green!20.0}+33.7 & \cellcolor{green!7.3}+2.6 & -- & \cellcolor{green!20.0}+44.7 & \cellcolor{green!10.6}+5.1 \\
\bottomrule
\end{tabular}}
\caption{Direct test accuracy (\%) after LoRA fine-tuning.
In-domain uses a dataset-specific adapter; Transfer uses
the SimplePriority adapter. The final row reports mean
changes from Base (pp), averaged equally across five LLMs.}
\label{tab:target-vs-priority-tuning}
\end{table*}

\begin{figure*}[t!]
    \centering
    \includegraphics[width=\linewidth, trim=0 8 0 8, clip]{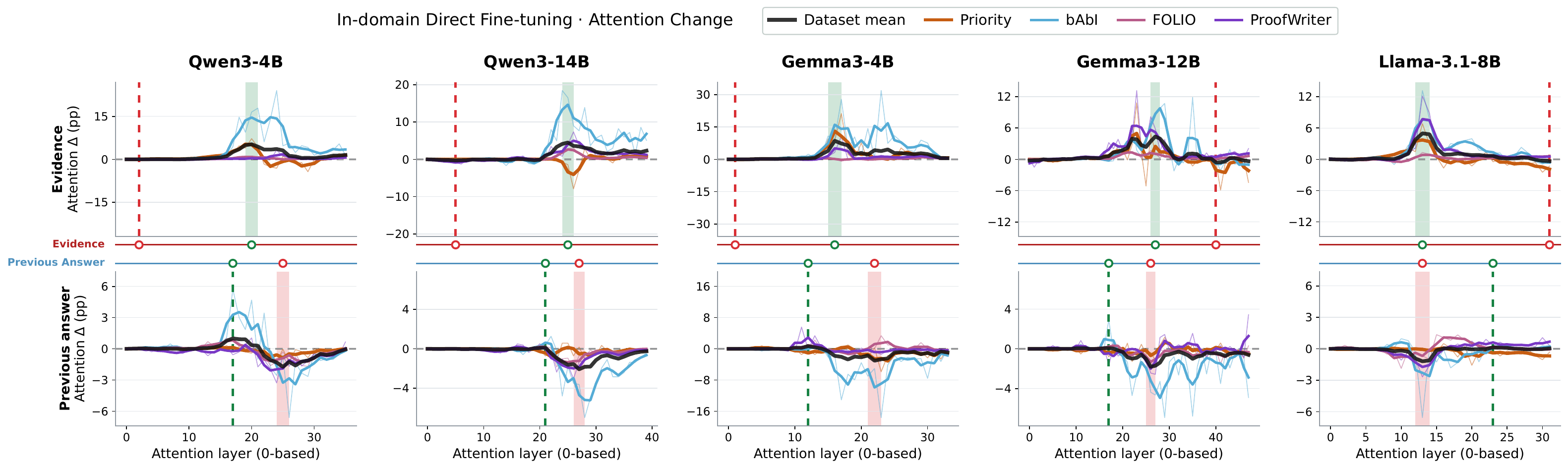}
\caption{Layer-wise attention changes after in-domain tuning (pp).
Top/bottom: evidence/previous answer.
Green shading/red dashed lines mark mean maxima/minima;
circles compare layer positions across sources.}
    \label{fig:fig_main_attention_change}
\end{figure*}

% ================================================================================

% ================================================================================
\subsection{Learning Belief Revision and Cross-Domain Transfer}
\label{sec:learning}

We test whether targeted fine-tuning improves
post-retraction judgments and transfers across domains.
We train five LLMs with LoRA \citep{hu2022lora} on Direct-format
answers, using one adapter per dataset.
Transfer uses the SimplePriority adapter without further tuning.
Evaluation uses Direct generation without reassessment
instructions, regardless of initial-response correctness.

\textbf{In-domain fine-tuning substantially improves
post-retraction accuracy.}
Table~\ref{tab:target-vs-priority-tuning} shows preservation
gains for every LLM and dataset, with dataset means
ranging from 31.5 to 53.2 pp.
Changing accuracy also improves on average, with gains
on ProofWriter (+33.7 pp) and bAbI (+44.7 pp),
despite individual regressions.
These gains indicate improved accuracy when the original
gold answer remains justified, without establishing
how surviving evidence is reassessed.
These gains exceed those from explicit instructions.
Appendix~\ref{app:cross-domain-transfer} reports all
dataset-pair transfers, showing
\textcolor{preservecolor}{answer-preserving} gains
beyond the training domain.

\paragraph{Attention Changes after Fine-Tuning.}
Figure~\ref{fig:fig_main_attention_change} shows changes
in attention to evidence and prior answers after tuning.
\textbf{Fine-tuning alters attention allocation across layers
and sources.}
Several LLMs attend more to evidence at intermediate layers,
while prior-answer attention shifts in either direction.
These patterns do not localize belief revision to a specific
module, and attention alone does not establish causal
recomputation from remaining evidence
\citep{jain-wallace-2019-attention}.

\begin{figure}[ht!]
\centering
\includegraphics[width=\linewidth]
{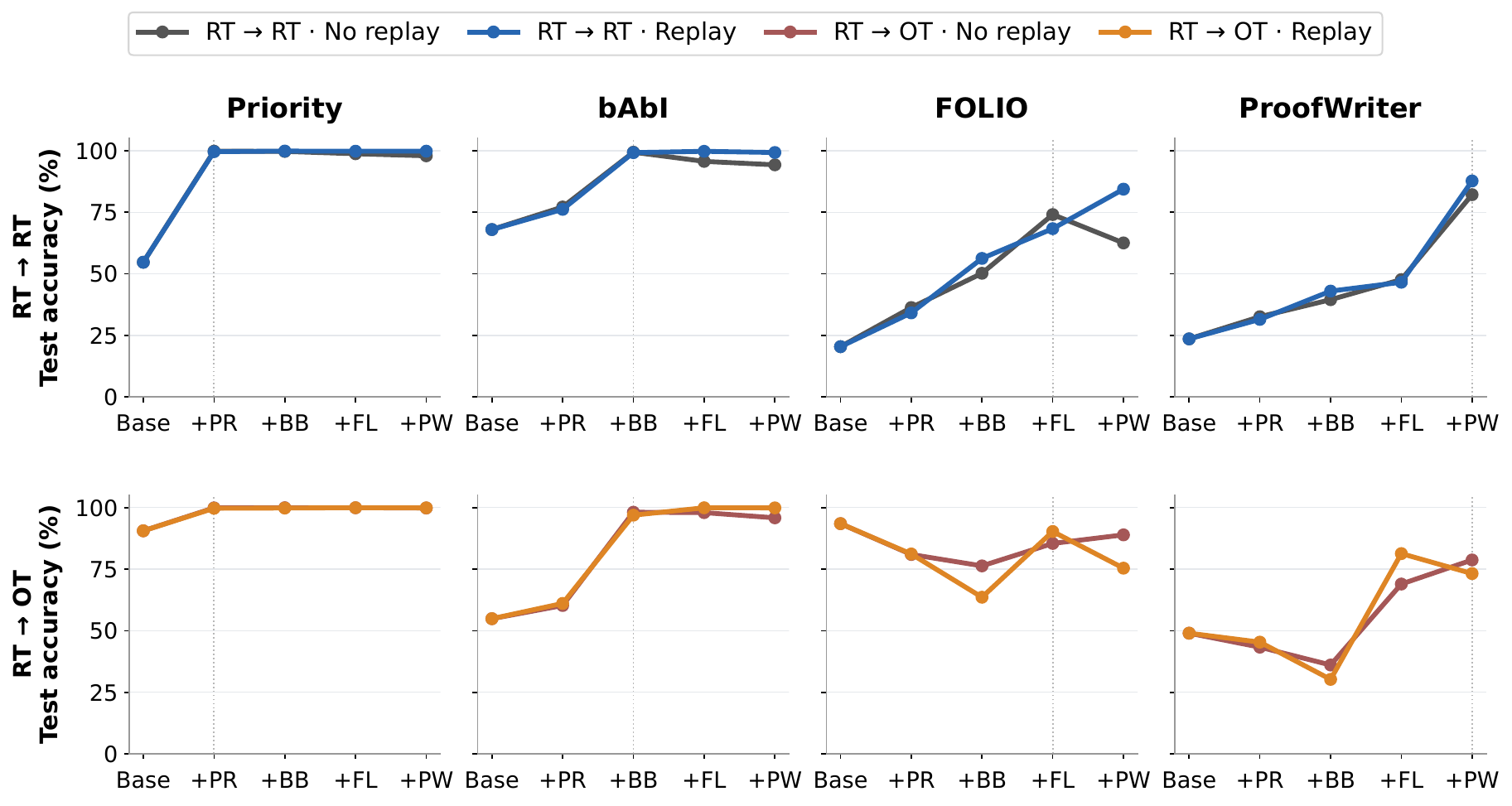}
\caption{
Sequential training in the order PR, BB, FL, and PW,
with and without replay.
}
\label{fig:main_learning_curve_model_mean}
\end{figure}

\textbf{Answer preservation shows modest average transfer
and gains under sequential training.}
SimplePriority training raises mean
\textcolor{preservecolor}{answer-preserving} accuracy by
6.0 pp on FOLIO, 2.0 pp on ProofWriter, and 5.8 pp on bAbI,
with less consistent gains in
\textcolor{changecolor}{answer-changing} cases.
Following continual fine-tuning
\citep{scialom-etal-2022-fine}, we train sequentially on
SimplePriority (PR), bAbI (BB), FOLIO (FL), and
ProofWriter (PW).
Figure~\ref{fig:main_learning_curve_model_mean} shows
preservation gains with and without replay in this fixed order;
model-wise results appear in
Appendix~\ref{app:sequential-finetuning}.
Base comparisons are descriptive because
initial responses may differ across runs.

\paragraph{Access Sensitivity after Fine-Tuning.}
Table~\ref{tab:run7-nine-settings-vllm} compares attention-blocking
effects before and after LoRA tuning across nine settings:
the last 1, 2, or 4 positions within the withdrawal
instruction, repeated question, or full input
(Appendix~\ref{app:experimental-settings}).
\textbf{Fine-tuned LLMs generally show smaller changes
under attention blocking.}
Reduced sensitivity may reflect access through unblocked
paths, distributed information, or learned task regularities.
The interventions neither distinguish these explanations
nor establish reassessment of the remaining evidence.
\section{Discussion}

\paragraph{Support Withdrawal Requires Reassessment.}
Belief contraction removes support without necessarily
negating the conclusion
\citep{alchourron1985logic,doyle1979truth}.
After withdrawal, a conclusion loses its justification
only if no alternative support remains.
Our results show that LLMs often miss this distinction,
changing answers that remain justified.
Reliable revision therefore requires reassessing the
surviving evidence, rather than rejecting conclusions
associated with withdrawn support.

\paragraph{Reasoning and Belief Maintenance.}
Correct reasoning from a fixed context does not ensure
correct judgment when evidence changes.
After support withdrawal, a model must identify which
inferences lose their justification and which remain
supported by alternative evidence.
Answer-preserving cases expose this gap because simply
repeating or rejecting the prior answer cannot handle
both revision conditions.
External memory and symbolic reasoning may help by making
updated facts and their dependencies explicit
\citep{zhong-etal-2023-mquake,
kassner-etal-2021-beliefbank,
lyu-etal-2023-faithful}.
Such systems should therefore be evaluated on whether
they preserve surviving justifications while revising
conclusions that depend on withdrawn evidence.

\paragraph{Learning to Revise after Support Withdrawal.}
The modest instruction gains and substantially larger
fine-tuning gains suggest that reliable revision is more
readily learned than elicited through reconsideration prompts,
consistent with findings on prompted and trained
self-correction
\citep{huang2024large,kumar2025training}.
Transfer from SimplePriority indicates that some learned
behavior generalizes, although uneven gains across LLMs
and conditions do not establish a domain-general procedure.
The small access-blocking effects after tuning may reflect
distributed information or learned task regularities.
Future evaluations should vary both regularities and
information paths to distinguish evidence reassessment
from task-specific response strategies.

\begin{table}[t]
\centering\small
\setlength{\tabcolsep}{2.5pt}
\renewcommand{\arraystretch}{1.12}
\resizebox{\columnwidth}{!}{%
\begin{tabular}{@{}cc|rr|rr|rr|rr@{}}
\hline
$E$ & $A_1$ & \multicolumn{2}{c}{Priority} & \multicolumn{2}{c}{FOLIO} & \multicolumn{2}{c}{ProofWriter} & \multicolumn{2}{c}{bAbI} \\
\cline{3-10}
 & & \multicolumn{1}{c}{Base} & \multicolumn{1}{c|}{LoRA} & \multicolumn{1}{c}{Base} & \multicolumn{1}{c|}{LoRA} & \multicolumn{1}{c}{Base} & \multicolumn{1}{c|}{LoRA} & \multicolumn{1}{c}{Base} & \multicolumn{1}{c}{LoRA} \\
\hline
\rowcolor{gray!15}
\multicolumn{2}{c|}{\rule{0pt}{2.5ex}} & \multicolumn{8}{c}{\bfseries RT $\to$ RT} \\
\rowcolor{gray!8}
\textcolor{green!40!black}{$\checkmark$} & \textcolor{green!40!black}{$\checkmark$} & {53.4} & {99.5} & {24.1} & {63.4} & {31.7} & {75.1} & {71.3} & {99.6} \\
\textcolor{green!40!black}{$\checkmark$} & \textcolor{red!70!black}{$\times$} & \cellcolor{green!6.0}$+2.8$ & $\mathbf{0.0}$ & \cellcolor{red!4.6}$\mathbf{-0.9}$ & \cellcolor{red!4.7}$-1.0$ & \cellcolor{green!5.1}$+1.5$ & \cellcolor{red!4.4}$\mathbf{-0.5}$ & \cellcolor{red!6.5}$-3.6$ & $\mathbf{0.0}$ \\
\textcolor{red!70!black}{$\times$} & \textcolor{green!40!black}{$\checkmark$} & \cellcolor{red!6.8}$-4.0$ & \cellcolor{red!4.1}$\mathbf{-0.1}$ & \cellcolor{green!4.1}$\mathbf{+0.2}$ & \cellcolor{red!4.1}$\mathbf{-0.2}$ & $\mathbf{0.0}$ & \cellcolor{red!4.3}$-0.5$ & \cellcolor{red!4.2}$\mathbf{-0.2}$ & \cellcolor{red!5.2}$-1.7$ \\
\textcolor{red!70!black}{$\times$} & \textcolor{red!70!black}{$\times$} & \cellcolor{red!4.7}$-1.0$ & \cellcolor{red!4.1}$\mathbf{-0.1}$ & \cellcolor{red!4.2}$\mathbf{-0.3}$ & \cellcolor{red!4.8}$-1.2$ & \cellcolor{green!4.8}$\mathbf{+1.1}$ & \cellcolor{red!4.7}$\mathbf{-1.1}$ & \cellcolor{red!13.6}$-13.8$ & \cellcolor{red!7.4}$\mathbf{-4.8}$ \\
\hline
\rowcolor{gray!15}
\multicolumn{2}{c|}{\rule{0pt}{2.5ex}} & \multicolumn{8}{c}{\bfseries RT $\to$ OT} \\
\rowcolor{gray!8}
\textcolor{green!40!black}{$\checkmark$} & \textcolor{green!40!black}{$\checkmark$} & {99.2} & {99.8} & {94.2} & {95.0} & {35.7} & {87.5} & {61.6} & {99.6} \\
\textcolor{green!40!black}{$\checkmark$} & \textcolor{red!70!black}{$\times$} & \cellcolor{red!4.1}$-0.1$ & $\mathbf{0.0}$ & \cellcolor{green!4.7}$+0.9$ & \cellcolor{green!4.3}$\mathbf{+0.4}$ & \cellcolor{red!5.3}$-1.9$ & \cellcolor{green!4.0}$\mathbf{+0.1}$ & \cellcolor{red!8.8}$-6.9$ & \cellcolor{red!4.1}$\mathbf{-0.1}$ \\
\textcolor{red!70!black}{$\times$} & \textcolor{green!40!black}{$\checkmark$} & \cellcolor{red!4.2}$-0.3$ & \cellcolor{red!4.0}$\mathbf{-0.1}$ & \cellcolor{red!4.5}$\mathbf{-0.7}$ & \cellcolor{red!4.5}$\mathbf{-0.7}$ & \cellcolor{red!4.2}$-0.3$ & \cellcolor{green!4.1}$\mathbf{+0.1}$ & \cellcolor{red!14.4}$-14.9$ & \cellcolor{red!8.6}$\mathbf{-6.6}$ \\
\textcolor{red!70!black}{$\times$} & \textcolor{red!70!black}{$\times$} & \cellcolor{red!4.2}$-0.3$ & \cellcolor{red!4.0}$\mathbf{-0.1}$ & $\mathbf{0.0}$ & \cellcolor{red!4.4}$-0.5$ & \cellcolor{red!5.3}$-1.9$ & \cellcolor{green!4.1}$\mathbf{+0.2}$ & \cellcolor{red!16.0}$-17.2$ & \cellcolor{red!10.8}$\mathbf{-9.8}$ \\
\hline
\end{tabular}}
\caption{Attention-blocking sensitivity before and after
LoRA fine-tuning.
Bold: full-access accuracy (\%); other entries:
mean accuracy changes (pp) across nine settings.}
% \caption{vLLM sensitivity to direct access blocking. Baseline rows show unmasked accuracy (\%); blocked rows show mean $\Delta\pm\mathrm{SD}$ (pp) over nine settings: R/Q2/G $\times$ $k=1,2,4$. Within each setting, models are averaged equally; SD is the population standard deviation across the nine model-averaged settings, not uncertainty or between-model variability. Each dataset uses the same common models for both branches, transitions and all blocked targets (Priority: 2, bAbI: 0, FOLIO: 0, ProofWriter: 1). Checks/crosses indicate direct access allowed/blocked only at selected input queries; generated queries remain unblocked. Missing datasets are --. Small signed means can conceal opposing effects; mean absolute changes and model-level statistics are saved separately.}
\label{tab:run7-nine-settings-vllm}
\end{table}

\section{Conclusion}

% 언어모델이 지속적으로 정보에 대한 믿음을 구성하면서, 정보의 상태는 지속적으로 변화한다. 
% 이러한 구조 속에서 LLM이 올바르게 결론을 제공하는지는 중요한 과제이다. 
LLMs increasingly reason over information whose status
changes over time.
Maintaining correct judgments under such changes is therefore
an important challenge.
Withdrawing support does not necessarily require withdrawing
the conclusion itself when alternative evidence remains.
Our experiments show that LLMs struggle with this distinction,
often abandoning answers still justified by the remaining evidence.
Explicit instructions offer limited improvement, whereas
targeted training produces substantial gains with uneven
transfer across domains.
These results leave open whether training develops a general
capacity for reassessment or task-specific response patterns.
Reliable revision requires more than responding to withdrawal:
LLMs must reevaluate the remaining evidence and preserve
the conclusion when it remains justified.

% ========================================================================

\section*{Limitations}

Our two-stage protocol makes the initial judgment and
subsequent support withdrawal explicit, enabling controlled
comparisons between answer preservation and change.
It focuses on judgments elicited within the interaction;
revision involving implicit beliefs or less explicit
updates remains to be studied.
Our tasks also use constrained answer spaces.
Extending the evaluation to open-ended generation would
require criteria for identifying support and assessing
whether a response remains justified.

Attention-access interventions and steering characterize
how selected information sources and representations
affect post-retraction responses.
They do not uniquely identify the reasoning procedure,
since contextualized states may retain information
from multiple sources.
The mechanistic interpretation of these effects therefore
remains open.

Fine-tuning yields substantial in-domain gains, while
cross-domain transfer varies by model and revision condition.
The extent to which these gains reflect general
evidence reassessment or learned task regularities
requires further evaluation with broader domains
and different withdrawal patterns.

\section*{Ethical Considerations}

We use publicly available benchmarks and a synthetic
candidate-selection dataset with fictional candidates
and rule-defined answers.
Our evaluation reveals biases in belief revision,
including a tendency to change answers after support
withdrawal even when they remain justified.
By distinguishing warranted revision from unnecessary
answer changes, this work contributes to understanding
biases in LLM judgments.
These behavioral biases are distinct from social biases;
our findings do not establish fairness in real-world use.
% ============================
% ============================
% ============================
% ============================
% ============================

% \subsection{References}


\bibliography{custom}

\clearpage
\appendix

% \section*{Appendix}
% \label{sec:appendix}
% ============================================================
% Preamble: 이미 선언된 패키지는 중복해서 추가하지 않습니다.
% ============================================================
% \usepackage{capt-of}
% \usepackage{tabularx}
% \usepackage{booktabs}
% \usepackage{array}

% ============================================================
% Appendix
% ============================================================

% \clearpage
% \appendix

% Preamble: 기존 xcolor 선언이 있다면 table 옵션만 추가

\twocolumn[{%
\begin{minipage}{\textwidth}
\begingroup
\centering
\small
\setlength{\tabcolsep}{5pt}
\renewcommand{\arraystretch}{1.18}

\begin{tabularx}{\textwidth}{
    @{}
    >{\raggedright\arraybackslash}p{2.5cm}
    >{\raggedright\arraybackslash}X
    >{\raggedright\arraybackslash}X
    >{\raggedright\arraybackslash}X
    @{}
}
\toprule
\textbf{Study}
& \textbf{Update setting}
& \textbf{Judgment target}
& \textbf{Preservation after withdrawal} \\
\midrule

Geometry of Truth
\cite{marks2024geometrytruthemergentlinear}
&
No evidence update.
&
Represented factual truth.
&
Not evaluated.
\\
\addlinespace[5pt]

Retraction
\cite{yang2026when}
&
Answer reconsideration without new evidence.
&
Acknowledgment of an incorrect answer.
&
Answer retraction, not support withdrawal.
\\
\addlinespace[5pt]

Sycophancy heads
\cite{genadi2026sycophancy}
&
User pushback without new evidence.
&
Correct-to-incorrect answer changes.
&
No support withdrawal.
\\

\midrule

Defeasible NLI
\cite{rudinger2020thinking}
&
Evidence added to strengthen or weaken an inference.
&
Updated hypothesis support.
&
Evidence addition, not withdrawal.
\\
\addlinespace[5pt]

Belief-R
\cite{wilie2024belief}
&
Additional premises.
&
Conclusion maintenance or revision.
&
Premise addition, not withdrawal.
\\

\midrule

ReviseQA
\cite{helwe2025reviseqa}
&
Premises added or removed across turns.
&
Conclusion entailment after each edit.
&
Removal only in answer-changing edits.
\\
\addlinespace[5pt]

DeltaLogic
\cite{dhanda2026deltalogic}
&
Minimal premise edits, including removal.
&
Conclusion-label stability or revision.
&
Support removal yields Unknown.
\\

\midrule
\rowcolor{gray!10}

\textbf{Ours}
&
Support withdrawn without asserting its negation.
&
Answer validity under remaining evidence.
&
\textbf{The original answer may remain justified.}
\\

\bottomrule
\end{tabularx}
\par

\captionof{table}{
Comparison of belief-related settings.
Row groups distinguish no evidence update, evidence addition,
and premise edits including removal.
}
\label{tab:belief_comparison}

\endgroup
\vspace{1em}
\end{minipage}%
}]

% ============================================================
% Two-column discussion
% ============================================================
\section{Comparison with Prior Work}
\label{app:comparison_prior_work}

We evaluate how language models maintain and revise their beliefs
after supporting evidence is withdrawn.
In particular, we examine cases in which the original answer
should remain unchanged because the remaining evidence still
supports it.
Table~\ref{tab:belief_comparison} compares what prior studies
treat as a belief, what is updated or withdrawn, and whether
they evaluate answer preservation after support withdrawal.

\paragraph{Belief and Retraction Targets.}
Prior studies examine different objects under the term belief.
Geometry of Truth
\cite{marks2024geometrytruthemergentlinear}
studies representations of factual truth without changing
the evidence.
Retraction
\cite{yang2026when}
examines whether a model acknowledges that its own previously
generated answer is incorrect, while Sycophancy heads
\cite{genadi2026sycophancy}
examines answer changes under user pushback without new evidence.
Defeasible NLI
\cite{rudinger2020thinking}
and Belief-R
\cite{wilie2024belief}
evaluate how additional evidence affects an inference
or conclusion.
ReviseQA
\cite{helwe2025reviseqa}
and DeltaLogic
\cite{dhanda2026deltalogic}
also consider premise removal.
Our setting distinguishes withdrawing an answer from withdrawing
its support: the support is withdrawn externally, and the model
must determine whether the previous answer remains justified.
\paragraph{Answer Preservation After Support Withdrawal.}
Withdrawing evidence neither asserts its negation nor makes
the previous answer incorrect.
The remaining evidence determines whether the answer should
be preserved or changed.
ReviseQA \cite{helwe2025reviseqa} removes premises only in
answer-changing edits, sometimes adding negated facts.
DeltaLogic \cite{dhanda2026deltalogic} assigns Unknown after
support removal and preserves labels under irrelevant-fact additions.
Neither construction evaluates support removal that leaves
the original answer justified. 
We evaluate this case alongside answer-changing cases. 
\textcolor{black}{Including both cases means that a model cannot succeed by always changing its answer after a withdrawal or by always repeating its previous answer.}
\paragraph{Attention Interventions and Information Access.}
We examine how post-retraction responses vary when attention
to the original evidence and previous answer is blocked.
Unlike text deletion, these interventions retain the supplied
tokens and their contextualized states.
Unlike activation steering in Retraction \cite{yang2026when}
and Sycophancy heads \cite{genadi2026sycophancy}, they target
attention paths to designated evidence and answer segments.
Blocking applies only to selected input-query positions;
subsequent generated queries remain unblocked.

% Preamble

%  ========================================================================
\newcommand{\preserving}{%
    \textcolor{preservecolor}{\textbf{Preserving:}}}
\newcommand{\changing}{%
    \textcolor{changecolor}{\textbf{Changing:}}}
\newcommand{\keepanswer}[1]{%
    \textcolor{preservecolor}{#1}}
\newcommand{\changeanswer}[1]{%
    \textcolor{changecolor}{#1}}

%  ========================================================================
%  ========================================================================

\begin{table*}[t!]
\centering
\small
\setlength{\tabcolsep}{6pt}
\renewcommand{\arraystretch}{1.15}
\begin{tabularx}{\textwidth}{
    @{}
    >{\raggedright\arraybackslash}p{1.5cm}
    >{\raggedright\arraybackslash}p{3.1cm}
    >{\raggedright\arraybackslash}X
    @{}
}
\toprule
Dataset & Task & Illustrative Example \\
\midrule

\textbf{Priority}
&
Select the priority candidate;
after priority is withdrawn, select by score.
&
\textbf{Initial:} A has priority. Answer: A. \newline
\preserving{} Scores A=9, B=7, C=5.
Withdraw A's priority $\rightarrow$ \keepanswer{A}. \newline
\changing{} Scores A=7, B=9, C=5.
Withdraw A's priority $\rightarrow$ \changeanswer{B}.
\\
\midrule

\textbf{bAbI}
&
Determine a person's location
from the remaining movement reports.
&
\textbf{Reports:} E1: Mary moved to the kitchen.
E2: Mary moved to the garden. \newline
\textbf{Question:} Where is Mary? Initial answer: garden. \newline
\preserving{} Withdraw E1
$\rightarrow$ \keepanswer{garden}. \newline
\changing{} Withdraw E2
$\rightarrow$ \changeanswer{kitchen}.
\\
\midrule

\textbf{FOLIO}
&
Classify a conclusion from
the remaining logical premises.
&
\textbf{Premises:} P1: Every cat is an animal.
P2: Milo is a cat. P3: Anna is a teacher. \newline
\textbf{Conclusion:} Milo is an animal.
Initial answer: True. \newline
\preserving{} Withdraw P3
$\rightarrow$ \keepanswer{True}. \newline
\changing{} Withdraw P2
$\rightarrow$ \changeanswer{Unknown}.
\\
\midrule

\textbf{ProofWriter}
&
Determine whether a statement
remains derivable after support is withdrawn.
&
\textbf{Facts:} F1: Anne is blue. F2: Anne is kind.
F3: Bob is tall. F4: Bob is young. \newline
\textbf{Rules:} R1: Blue people are quiet.
R2: Kind people are quiet. \newline
\textbf{Statement:} Anne is quiet.
Initial answer: True. \newline
\preserving{} Withdraw F1, F3, F4
$\rightarrow$ \keepanswer{True}. \newline
\changing{} Withdraw F1, F2, F3
$\rightarrow$ \changeanswer{Unknown}.
\\

\bottomrule
\end{tabularx}
\caption{
Illustrative retraction examples.
\textcolor{preservecolor}{Preserving} and
\textcolor{changecolor}{changing} conditions refer to
gold-answer transitions.
Priority uses different score assignments across conditions;
the other datasets share the same initial evidence.
ProofWriter withdraws three items in each condition.
}
\label{tab:dataset-examples}
\end{table*}

%  ========================================================================
%  ========================================================================
\section{Dataset Construction}
\label{app:dataset}

\subsection{SimplePriority Dataset}
\label{app:dataset-priority}

We introduce \textbf{SimplePriority}, a synthetic candidate-selection
dataset that separates priority-based selection from
score-based selection.
Each instance contains three candidates with distinct integer
scores sampled from 1 to 999 and one candidate assigned priority.
The task selects the priority candidate while its priority
is active; otherwise, it selects the highest-scoring candidate.
The initial evidence $E_0$ contains the selection rule,
candidate scores, and priority assignment.
The withdrawn set $R$ contains only the priority assignment.
Thus, $E_1=E_0\setminus R$ retains all candidates, scores,
and the selection rule.

For each base case, we construct two score configurations.
In the \textcolor{preservecolor}{answer-preserving configuration},
the priority candidate also has the highest score,
so $y_1=y_0$.
In the \textcolor{changecolor}{answer-changing configuration},
we swap its score with another candidate's score so that
a different candidate has the highest score, giving $y_1\neq y_0$.
The two configurations share the initial gold answer
and the same priority withdrawal, but differ in their score
assignments and post-retraction gold answers.

Each configuration is presented with and without an irrelevant
numeric inventory containing three numbers.
This yields four examples per base case:
two answer conditions $\times$ two inventory variants.
The inventory does not affect either gold answer.
Candidate names and their order in the question are varied.
Table~\ref{tab:dataset-examples} illustrates the selection
and retraction procedure.

\subsection{Common Retraction Framework}
\label{app:dataset-framework}

We construct retraction tasks from Priority, bAbI \cite{weston2015towards_babi}, FOLIO \cite{han2022folio}, and ProofWriter \cite{tafjord2021proofwriter}, covering priority-based selection, temporal state tracking, and logical inference.
Priority is generated synthetically, while the other
three datasets are adapted from existing benchmarks.

Following Section~\ref{sec:revision_task}, let $E_0$ denote
the initial evidence, $q$ the query, and $R\subseteq E_0$
the items to be withdrawn.
The remaining evidence is $E_1=E_0\setminus R$.
For each task $D$, its inference semantics
$\operatorname{Cl}_D$ and answer function $f_D$ determine
the gold answers:
\begin{equation}
\begin{aligned}
    y_0 &= f_D(\operatorname{Cl}_D(E_0),q),\\
    y_1 &= f_D(\operatorname{Cl}_D(E_1),q).
\end{aligned}
\label{eq:dataset-retraction}
\end{equation}
Retraction excludes the items in $R$ from the active evidence
without asserting their negation.
The order of surviving evidence is preserved when relevant,
as in temporal state tracking.

Each example is evaluated through the two-stage interaction
in Equation~\ref{eq:interaction}.
The LLM first answers $q_1=q$ using $E_0$, producing
$a_{\mathrm{prior}}$.
We then append the withdrawal instruction $\rho(R)$
and repeat the question, $q_2=q$, to obtain
$a_{\mathrm{post}}$.
The full preceding conversation remains available,
including $E_0$ and the model's actual initial response;
$a_{\mathrm{prior}}$ is not replaced with the gold answer.

We label an example
\textcolor{preservecolor}{answer-preserving} when $y_1=y_0$
and \textcolor{changecolor}{answer-changing} when $y_1\neq y_0$.
These labels describe gold-answer transitions independently
of $a_{\mathrm{prior}}$.
For bAbI, FOLIO, and ProofWriter, paired conditions share
$E_0$ and $q$ but withdraw different sets $R$.
Priority pairs instead share $q$, $y_0$, and the priority
withdrawal $R$, while differing in their score assignments
within $E_0$.
Table~\ref{tab:dataset-examples} provides examples
of both conditions.

\subsection{Data Settings and Subset Selection}
\label{app:dataset-selection}

We select base cases admitting both answer-preserving
($y_1=y_0$) and answer-changing ($y_1\neq y_0$) conditions.
Tables~\ref{tab:retraction-labels} and~\ref{tab:dataset-sizes}
report post-retraction gold answers and split sizes.
For FOLIO and ProofWriter, \texttt{True}/\texttt{False}
denote entailment of the statement/its negation,
and \texttt{Unknown} denotes neither.
Selected answer-changing cases transition from
\texttt{True}/\texttt{False} to \texttt{Unknown}.

\paragraph{Priority.}
\begin{itemize}[leftmargin=*, topsep=2pt, itemsep=1pt, parsep=0pt]
    \item \textbf{Data setting.}
    Each base case yields four examples:
    answer-preserving/changing score configurations
    $\times$ inventory absent/present.
    Paired configurations share $y_0$ and $R$.

    \item \textbf{Filtering.}
    We verify distinct scores, rule-based gold answers,
    and complete variants.
    Candidate order is matched within pairs;
    sorted score triples are disjoint across splits.
\end{itemize}

\paragraph{bAbI.}
\begin{itemize}[leftmargin=*, topsep=2pt, itemsep=1pt, parsep=0pt]
    \item \textbf{Data setting.}
    We use Task 1, withdrawing one movement report:
    the queried person's latest report for answer-changing
    cases, or an earlier report or another person's report
    for answer-preserving cases.

    \item \textbf{Filtering.}
    We require source-answer agreement and both conditions,
    restricting changes to different known locations.
    Gold answers are recomputed from surviving reports
    in order.
    Preserving withdrawal types are recorded,
    and stories remain within one split.
\end{itemize}

\paragraph{FOLIO.}
\begin{itemize}[leftmargin=*, topsep=2pt, itemsep=1pt, parsep=0pt]
    \item \textbf{Data setting.}
    Initially \texttt{True}/\texttt{False} cases with FOL
    annotations yield two single-premise withdrawals:
    one preserving $y_0$ and one yielding \texttt{Unknown}.
    We expand 82/18/24 source cases into 250/100/500
    base cases (train/validation/test) through entity renaming,
    premise-order shuffling, and conclusion negation.

    \item \textbf{Filtering.}
    We require matching natural-language/FOL premise counts,
    parseable and consistent formulas, and solver--source
    label agreement.
    Both gold labels are recomputed for every variant;
    timeouts and inconclusive results are excluded.
    Selection favors initial-label balance where possible.
    Linked stories, identical premise sets, and source-case
    variants remain within one split.
\end{itemize}

\paragraph{ProofWriter.}
\begin{itemize}[leftmargin=*, topsep=2pt, itemsep=1pt, parsep=0pt]
    \item \textbf{Data setting.}
    We select initially \texttt{True}/\texttt{False} questions
    with one to three minimal support sets.
    Each condition withdraws three facts and/or rules:
    preserving leaves a support set intact, while changing
    intersects every support set and yields \texttt{Unknown}.

    \item \textbf{Filtering.}
    Gold labels are verified by full forward inference.
    Multiple support sets share no fact or rule items;
    paired withdrawals match fact/rule counts.
    Selection is stratified by support count and initial label.
    We record question negation, proof depth, and direct-fact
    support, select one question per theory, and retain source
    splits after duplicate checks.
\end{itemize}

Dataset revisions and sampling seeds are fixed before evaluation.
All paired examples and variants remain within the same split.

\begin{table}[t]
\centering
\footnotesize
\setlength{\tabcolsep}{4pt}
\renewcommand{\arraystretch}{1.12}

\begin{tabularx}{\columnwidth}{
    @{}l
    >{\raggedright\arraybackslash}X
    >{\raggedright\arraybackslash}X@{}
}
\toprule
Dataset
& \textcolor{preservecolor}{\textbf{Answer-preserving}}
& \textcolor{changecolor}{\textbf{Answer-changing}} \\
\midrule
Priority
& Same candidate
& Different candidate \\
bAbI
& Same location
& Different known location \\
FOLIO
& Same True/False label
& Unknown \\
ProofWriter
& Same True/False label
& Unknown \\
\bottomrule
\end{tabularx}

\caption{
Post-retraction gold answers $y_1$ relative to $y_0$
in open-answer evaluation.
}
\label{tab:retraction-labels}
\end{table}

\begin{table}[t]
\centering
\footnotesize
\setlength{\tabcolsep}{4pt}
\renewcommand{\arraystretch}{1.12}

\begin{tabular*}{\columnwidth}{
    @{\extracolsep{\fill}}lrrr@{}
}
\toprule
Dataset & Train & Validation & Test \\
\midrule
Priority
& 1,000 (250) & 400 (100) & 2,000 (500) \\
bAbI
& 500 (250) & 200 (100) & 1,000 (500) \\
FOLIO
& 500 (250) & 200 (100) & 1,000 (500) \\
ProofWriter
& 500 (250) & 200 (100) & 1,000 (500) \\
\bottomrule
\end{tabular*}

\caption{
Numbers of retraction examples, with base cases in parentheses.
Each base case yields two examples through answer-preserving
and answer-changing conditions.
Priority yields four through additional inventory variants.
FOLIO base cases are expanded from 82/18/24 source cases
for train/validation/test, respectively.
}
\label{tab:dataset-sizes}
\end{table}

% ========================================================================
% Preamble
% \usepackage[most]{tcolorbox}
% \usepackage{xcolor}

\newcommand{\withdrawn}[1]{\textcolor{changecolor}{#1}}

\newtcolorbox{promptcard}[1]{
    enhanced,
    title={#1},
    colback=white,
    colframe=black!30,
    colbacktitle=black!5,
    coltitle=black,
    fonttitle=\small\bfseries,
    fontupper=\footnotesize,
    boxrule=0.4pt,
    arc=1mm,
    left=1.5mm,
    right=1.5mm,
    top=1mm,
    bottom=1mm,
    before skip=0pt,
    after skip=0pt
}

\tcbset{
    promptstage/.style={
        enhanced,
        fonttitle=\footnotesize\bfseries,
        fontupper=\footnotesize,
        coltitle=white,
        boxrule=0.4pt,
        arc=1mm,
        left=1mm,
        right=1mm,
        top=1mm,
        bottom=1mm,
        before skip=3pt,
        after skip=3pt
    }
}

\newtcolorbox{initialstage}{
    promptstage,
    title={Stage 1: Initial Question},
    colframe=preservecolor,
    colback=preservecolor!4,
    colbacktitle=preservecolor
}

\newtcolorbox{retractionstage}{
    promptstage,
    title={Stage 2: Retraction},
    colframe=changecolor,
    colback=changecolor!4,
    colbacktitle=changecolor
}

\newcommand{\initialresponse}{%
    \textbf{Assistant:} \textit{[Actual initial response]}\par}
\newcommand{\retractionresponse}{%
    \textbf{Assistant:} \textit{[Generated post-retraction response]}\par}

\begin{figure*}[t!]
\centering

% ==================== Priority ====================
\begin{minipage}[t]{0.49\textwidth}
\vspace{0pt}
\begin{promptcard}{(a) Priority (Ours)}

\begin{initialstage}
\textbf{User}\par
\textbf{Scores:} A: 7; B: 9; C: 5.\par
\textbf{Priority:} \withdrawn{A has priority.}\par
Which candidate should be chosen: A, B, or C?\par
Answer with exactly one uppercase candidate letter
and no other text.
\end{initialstage}

\initialresponse

\begin{retractionstage}
\textbf{User}\par
\textbf{Update:}
\withdrawn{Withdraw the statement ``A has priority''.}\par
Which candidate should be chosen: A, B, or C?\par
Answer with exactly one uppercase candidate letter
and no other text.
\end{retractionstage}

\retractionresponse
\end{promptcard}
\end{minipage}
\hfill
% ==================== bAbI ========================
\begin{minipage}[t]{0.49\textwidth}
\vspace{0pt}
\begin{promptcard}{(b) bAbI}

\begin{initialstage}
\textbf{User}\par
\textbf{Reports:}
E1: Mary moved to the kitchen.
\withdrawn{E2: Mary moved to the garden.}\par
\textbf{Question:} Where is Mary?\par
Answer with exactly one location name or Unknown,
and no other text.
\end{initialstage}

\initialresponse

\begin{retractionstage}
\textbf{User}\par
\textbf{Update:}
\withdrawn{Withdraw report E2: ``Mary moved to the garden.''}\par
\textbf{Question:} Where is Mary?\par
Answer with exactly one location name or Unknown,
and no other text.
\end{retractionstage}

\retractionresponse
\end{promptcard}
\end{minipage}

\par\medskip

% ==================== FOLIO =======================
\begin{minipage}[t]{0.49\textwidth}
\vspace{0pt}
\begin{promptcard}{(c) FOLIO}

\begin{initialstage}
\textbf{User}\par
\textbf{Premises:}\par
P1: Every cat is an animal.\par
\withdrawn{P2: Milo is a cat.}\par
P3: Anna is a teacher.\par
\textbf{Conclusion:} Milo is an animal.\par
Is the conclusion True, False, or Unknown
given the active premises?
Answer with exactly one word: True, False, or Unknown.
\end{initialstage}

\initialresponse

\begin{retractionstage}
\textbf{User}\par
\textbf{Update:}
\withdrawn{Withdraw premise P2: ``Milo is a cat.''}\par
\textbf{Conclusion:} Milo is an animal.\par
Is the conclusion True, False, or Unknown
given the active premises?
Answer with exactly one word: True, False, or Unknown.
\end{retractionstage}

\retractionresponse
\end{promptcard}
\end{minipage}
\hfill
% ==================== ProofWriter =================
\begin{minipage}[t]{0.49\textwidth}
\vspace{0pt}
\begin{promptcard}{(d) ProofWriter}

\begin{initialstage}
\textbf{User}\par
\textbf{Facts:}
\withdrawn{F1: Anne is blue. F2: Anne is kind.
F3: Bob is tall.}
F4: Bob is young.\par
\textbf{Rules:}
R1: Blue people are quiet.
R2: Kind people are quiet.\par
\textbf{Statement:} Anne is quiet.\par
Does the statement follow from the active facts and rules?
Answer with exactly one word: True, False, or Unknown.
\end{initialstage}

\initialresponse

\begin{retractionstage}
\textbf{User}\par
\textbf{Update:} Withdraw the following facts or rules:\par
\withdrawn{-- F1: ``Anne is blue.'';
-- F2: ``Anne is kind.'';
-- F3: ``Bob is tall.''}\par
\textbf{Statement:} Anne is quiet.\par
Does the statement follow from the active facts and rules?
Answer with exactly one word: True, False, or Unknown.
\end{retractionstage}

\retractionresponse
\end{promptcard}
\end{minipage}

\caption{
Illustrative answer-changing prompts;
evidence withdrawn in Stage 2 is highlighted in pink.
}
\label{fig:task-prompts}
\end{figure*}

% =============================================================================
% =============================================================================
\subsection{Task Prompts and Response Formats}
\label{app:task-prompts}

Figure~\ref{fig:task-prompts} shows the two-stage prompts
for all four datasets.
In Stage 1, the model receives the initial evidence $E_0$
and question $q$, then generates $a_{\mathrm{prior}}$.
In Stage 2, we append the withdrawal instruction $\rho(R)$
and repeat the question to obtain $a_{\mathrm{post}}$.
The original evidence and the model's actual initial response
remain in the conversation history.
Pink text marks the items withdrawn in Stage 2.
System instructions are omitted, bracketed responses are
placeholders, and formatting is condensed for presentation.

\paragraph{Response Formats.}
Direct generation requests only the final answer.
Short CoT generation requests a JSON object containing
\texttt{reason} and \texttt{answer}, allowing a brief explanation
before the final answer.
Only \texttt{answer} is scored.
Both formats use the same evidence and withdrawal setting.

% =============================================================================
% =============================================================================

% Preamble% Preamble

\begin{table*}[t]
\centering
\footnotesize
\setlength{\tabcolsep}{5pt}
\renewcommand{\arraystretch}{1.05}
\begin{tabularx}{\textwidth}{
    @{}c
    >{\raggedright\arraybackslash}p{2.0cm}
    >{\raggedright\arraybackslash}X
    @{}
}
\toprule
ID & Condition & Instruction \\
\midrule

-- & Baseline
& No additional instruction. \\
\midrule

\multicolumn{3}{@{}l}{
\textbf{Remaining Evidence: Judge from the Active Evidence}} \\
\addlinespace[2pt]

1 & Remaining 1
& Exclude the withdrawn evidence and determine the answer
from the remaining evidence.
Keep the previous answer if it is still supported. \\

2 & Remaining 2
& Evaluate the remaining evidence after withdrawal.
Change the previous answer only when the remaining
evidence warrants it. \\
\midrule

\multicolumn{3}{@{}l}{
\textbf{Answer Verification: Reassess the Surviving Justification}} \\
\addlinespace[2pt]

3 & Verify 1
& Check whether the previous answer remains correct after
this withdrawal, then answer the question using
the remaining evidence. \\

4 & Verify 2
& Verify the previous response against the evidence
that remains. It may remain correct or need to change. \\
\midrule

\multicolumn{3}{@{}l}{
\textbf{Independent Judgment: Avoid Assuming Preservation or Change}} \\
\addlinespace[2pt]

5 & Independent 1
& Determine the answer independently from the remaining
evidence rather than assuming the previous response
is correct or incorrect. \\

6 & Independent 2
& Reassess the question using only the evidence still in force.
Do not change an answer merely because a withdrawal occurred. \\
\midrule

\multicolumn{3}{@{}l}{
\textbf{General Accuracy Controls: Encourage Careful Answering}} \\
\addlinespace[2pt]

7 & Careful 1
& Read the question carefully and determine the most
accurate answer from the information provided.
Check your response carefully before giving the final answer. \\

8 & Careful 2
& Consider the question attentively and evaluate the
information provided.
Give an accurate response that follows the task requirements
and requested answer format. \\

9 & Precise 1
& Check that your response answers the question precisely
and correctly.
Consider the available information carefully before
choosing and giving your final answer. \\

10 & Precise 2
& Pay close attention to the task and select an accurate answer.
Verify that the response is consistent with the
information provided. \\

11 & Accurate 1
& Focus on answering the question correctly using the
provided information.
Examine the details carefully without making unnecessary
assumptions about the appropriate response. \\

12 & Accurate 2
& Be careful and precise when answering this question.
Review the supplied information and choose the response
that best satisfies the stated task requirements. \\
\midrule

\multicolumn{3}{@{}l}{
\textbf{Additional Reconsideration Formulation}} \\
\addlinespace[2pt]

13 & Additional
& Check the surviving evidence carefully and give whichever
answer it supports, whether or not that agrees with
your earlier response. \\

\bottomrule
\end{tabularx}
\caption{
Thirteen post-retraction instructions and a baseline,
evaluated under Direct and short CoT.
}
\label{tab:reconsideration-prompts}
\end{table*}

% =============================================================================
% =============================================================================
\subsection{Instructions for Evidence Reassessment}
\label{app:retraction_support_instruction}

Our instructions distinguish support withdrawal from
answer invalidation.
Withdrawing evidence may remove one justification while
leaving alternative support intact.
The model should therefore reassess what the remaining
evidence warrants, rather than assume that the previous
answer must change.

We use targeted instructions with three emphases:
judging from the remaining evidence, verifying the previous
answer, and avoiding assumptions about its correctness,
with two formulations each.
All permit either answer preservation or change,
depending on the surviving support.
We compare them with general accuracy instructions,
one additional reassessment formulation, and a baseline
without additional instructions
(Table~\ref{tab:reconsideration-prompts}).
The general accuracy controls encourage careful answering
without explicitly directing evidence reassessment.

Each instruction is inserted between $\rho(R)$ and $q_2$
and evaluated under Direct and short CoT, holding $E_0$,
$R$, $q$, and $a_{\mathrm{prior}}$ fixed across instruction variants.
The additional formulation is excluded from
steering-direction extraction.

% =============================================================================
% Preamble

\newtcolorbox{reaccessprompt}[1]{
    breakable,
    colback=gray!4,
    colframe=gray!45,
    colbacktitle=gray!12,
    coltitle=black,
    fonttitle=\bfseries\footnotesize,
    fontupper=\footnotesize,
    title={#1},
    boxrule=0.4pt,
    arc=2pt,
    left=5pt,
    right=5pt,
    top=4pt,
    bottom=4pt,
    before skip=6pt,
    after skip=6pt
}

% =============================================================================

\subsection{\textcolor{black}{Instruction Effects by Model}}
\label{app:instruction_by_model}

We examine variation in instruction effects across models,
generation modes, and gold-answer conditions.
Table~\ref{tab:appendix-instruction-models} reports model-wise
accuracy averaged equally across four datasets.
Remaining-evidence instructions achieve the highest model-mean
accuracy in answer-preserving cases under Direct and CoT.
In answer-changing cases, remaining-evidence instructions
perform best under Direct, while independent judgment
performs best under CoT.
Differences are modest and model-dependent, with some
instructions associated with lower accuracy.
Initial responses are fixed across instruction variants,
but generated separately for the no-instruction baseline.
Comparisons with that baseline therefore do not isolate
instruction effects.

\textcolor{black}{At the model level, instruction effects are generally modest
and vary in direction. Still, two patterns hold for the across-model mean.
First, remaining-evidence instructions improve accuracy in all four settings,
by up to 3.2 pp. Second, instructions do not close the gap between
answer-preserving and answer-changing cases.
Under Direct generation, answer-preserving accuracy reaches at most 45.2\%
with any instruction, whereas answer-changing accuracy stays above 70\%.
Effects are also uneven across models, ranging from a 7.5 pp gain for Qwen3-14B in answer-changing cases under Direct generation to losses for Gemma3-4B
in answer-preserving cases under CoT.}

% ======================================================================

\begin{table}[t]
\centering\small
\setlength{\tabcolsep}{3pt}
\renewcommand{\arraystretch}{1.23}
\resizebox{\columnwidth}{!}{%
\begin{tabular}{@{}l|cc|cc|cc|cc|cc|cc@{}}
\toprule
Model & \multicolumn{2}{c|}{Baseline} & \multicolumn{2}{c|}{R} & \multicolumn{2}{c|}{V} & \multicolumn{2}{c|}{I} & \multicolumn{2}{c|}{G} & \multicolumn{2}{c}{A} \\
\cmidrule(lr){2-3}\cmidrule(lr){4-5}\cmidrule(lr){6-7}\cmidrule(lr){8-9}\cmidrule(lr){10-11}\cmidrule(l){12-13}
 & Direct & CoT & Direct & CoT & Direct & CoT & Direct & CoT & Direct & CoT & Direct & CoT \\
\midrule
\rowcolor{black!8}
\multicolumn{13}{c}{\rule{0pt}{2.6ex}\textbf{RT $\to$ RT}\rule[-0.7ex]{0pt}{0pt}} \\
Qwen3-4B & \cellcolor{black!4}46.9 & \cellcolor{black!4}76.7 & \textbf{52.8} & 79.9 & 50.2 & 78.7 & 51.2 & 79.3 & 50.7 & 79.0 & 46.0 & \textbf{81.0} \\
Qwen3-14B & \cellcolor{black!4}41.1 & \cellcolor{black!4}75.3 & 44.5 & \textbf{76.6} & 41.4 & 73.0 & \textbf{48.0} & 74.6 & 47.2 & 73.6 & 43.5 & 75.9 \\
Gemma3-4B & \cellcolor{black!4}32.4 & \cellcolor{black!4}64.2 & \textbf{35.1} & 63.1 & 33.5 & 63.3 & 30.8 & 62.6 & 31.5 & \textbf{65.2} & 30.6 & 57.5 \\
Gemma3-12B & \cellcolor{black!4}49.5 & \cellcolor{black!4}71.0 & 49.3 & 72.6 & 49.1 & 72.8 & \textbf{50.1} & 72.4 & 48.0 & 72.1 & 47.8 & \textbf{72.9} \\
Llama-3.1-8B & \cellcolor{black!4}39.7 & \cellcolor{black!4}55.9 & \textbf{44.2} & 59.6 & 39.6 & 53.7 & 38.7 & 54.5 & 39.2 & 56.7 & 35.0 & \textbf{61.1} \\
\midrule
\textbf{Model mean} & \cellcolor{black!4}41.9 & \cellcolor{black!4}68.6 & \textbf{45.2} & \textbf{70.4} & 42.8 & 68.3 & 43.7 & 68.7 & 43.3 & 69.3 & 40.6 & 69.7 \\
\midrule[0.8pt]
\rowcolor{black!8}
\multicolumn{13}{c}{\rule{0pt}{2.6ex}\textbf{RT $\to$ OT}\rule[-0.7ex]{0pt}{0pt}} \\
Qwen3-4B & \cellcolor{black!4}64.5 & \cellcolor{black!4}81.9 & \textbf{69.4} & \textbf{84.7} & 61.0 & 83.5 & 62.4 & 82.7 & 59.0 & 82.1 & 61.9 & 82.6 \\
Qwen3-14B & \cellcolor{black!4}81.0 & \cellcolor{black!4}89.8 & \textbf{88.5} & \textbf{92.5} & 83.8 & 90.5 & 85.2 & 90.9 & 83.2 & 88.6 & 84.8 & 90.7 \\
Gemma3-4B & \cellcolor{black!4}60.6 & \cellcolor{black!4}52.8 & 61.3 & 55.7 & 61.1 & 57.2 & 64.2 & \textbf{58.8} & 62.4 & 58.1 & \textbf{64.4} & 57.7 \\
Gemma3-12B & \cellcolor{black!4}90.8 & \cellcolor{black!4}91.1 & 89.7 & 90.6 & 89.1 & 89.6 & \textbf{91.5} & 90.4 & 89.0 & \textbf{91.1} & 87.7 & 89.6 \\
Llama-3.1-8B & \cellcolor{black!4}62.7 & \cellcolor{black!4}64.1 & 60.1 & 61.1 & 56.3 & 64.7 & \textbf{62.6} & \textbf{69.2} & 57.4 & 64.2 & 60.2 & 62.1 \\
\midrule
\textbf{Model mean} & \cellcolor{black!4}71.9 & \cellcolor{black!4}75.9 & \textbf{73.8} & 76.9 & 70.3 & 77.1 & 73.2 & \textbf{78.4} & 70.2 & 76.8 & 71.8 & 76.5 \\
\bottomrule
\end{tabular}}
% \captionsetup{justification=raggedright,singlelinecheck=false}
\caption{
Instruction-conditioned accuracy (\%), averaged over four
datasets and wording variants.
R/V/I: remaining evidence/verification/independent reassessment;
G: general accuracy; A: additional formulation.
Bold marks the best instruction type per response mode.
Baseline histories are generated independently.
}
\label{tab:appendix-instruction-models}
% \caption{Model-wise instruction-conditioned follow-up accuracy (\%), averaged equally over Priority, bAbI, FOLIO, and ProofWriter. Within each dataset and model, wording variants are averaged equally before averaging the four datasets. R: Remaining evidence (2 variants); V: Verification (2); I: Independent reassessment (2); G: General accuracy (6); A: Additional / surviving evidence (1). Direct and CoT are shown separately. RT $\to$ RT and RT $\to$ OT denote ground-truth label preservation and change. Bold marks the highest displayed accuracy across the five instruction types within each model and response mode; ties are bold. Model mean weights the same 5 common completed models equally. Baseline reports no-instruction Run1 accuracy; the other columns report absolute Run4 accuracy. Run1 and Run4 use the same test cases but independently generated initial histories. Scoring: answer\_normalization\_v2; original responses and strict format scores are retained.}
% \caption{eeEE}
% \label{tab:appendix-instruction-models}
\end{table}

\subsection{Self-Reported Need for Evidence Reaccess}
\label{app:verbalization}

We elicit the model's self-reported need to revisit the
original evidence before answering the question again.
This judgment concerns evidence reaccess, rather than
whether the previous answer should change.
We use three formulations to assess sensitivity to wording.
Each is evaluated with and without retraction, holding
$E_0$, $q$, and $a_{\mathrm{prior}}$ fixed.

\begin{reaccessprompt}{V1: Direct Choice}
To answer the question now, is the information retained
from your initial assessment sufficient, or do you need
to revisit the original evidence?
\end{reaccessprompt}

\begin{reaccessprompt}{V2: Sufficiency}
Is the information retained from your initial assessment
sufficient to determine the answer under the currently
active evidence, without revisiting the original evidence?
\end{reaccessprompt}

\begin{reaccessprompt}{V3: Necessity}
Do you need to revisit the original evidence to determine
the answer under the currently active evidence,
or is the information retained from your initial
assessment sufficient?
\end{reaccessprompt}

All formulations use the same response options:

\begin{reaccessprompt}{Shared Response Options}
A: The retained information is sufficient;
revisiting the original evidence is unnecessary.
\par\smallskip
B: The retained information is insufficient;
revisiting the original evidence is necessary.
\par\smallskip
Answer with exactly one letter: A or B.
\end{reaccessprompt}

V1 and V3 present the alternatives in opposite order,
whereas V2 asks directly about sufficiency.
The original evidence remains available throughout.
Responses are treated as self-reports, not as verified
measures of evidence access or computational necessity.

\begin{figure}[t]
\centering
\includegraphics[width=\columnwidth]{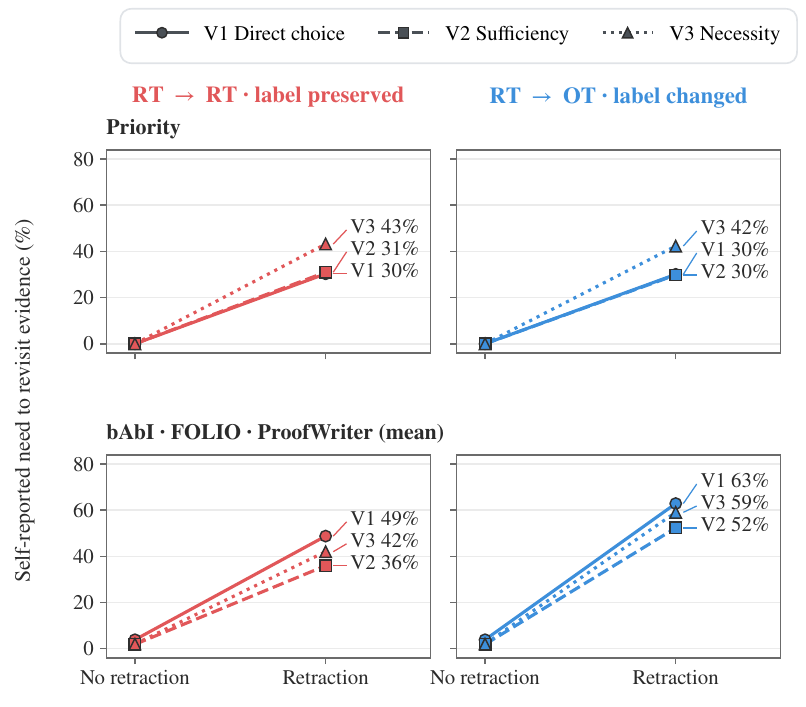}
\caption{
Evidence-reaccess self-reports (\%) across three formulations,
without and with retraction.
Rows show Priority and the mean over the other three datasets;
values are averaged equally over ten models.
Invalid responses count as no reaccess.
}
\label{fig:apdx-verbalization}
\end{figure}

\paragraph{Retraction and Wording Effects.}
Figure~\ref{fig:apdx-verbalization} compares the reported
need for evidence reaccess across three formulations.
Without retraction, the mean rate across models is at most 4\%.
Retraction increases this rate under every formulation,
although its magnitude varies with wording.
% The increase is therefore consistent across formulations, but the reported rates should not be interpreted as verified evidence-access behavior.
\textcolor{black}{The direction of the effect is thus consistent across formulations, but the reported rates should not be interpreted as verified evidence-access behavior.}

% =============================================================================
% =============================================================================
% =============================================================================
% =============================================================================

% =============================================================================
% =============================================================================

\clearpage
% ============================================================
\section{Additional Experimental Results}
\label{app:additional-experiments}

We report five sets of additional results:
\begin{enumerate}
    \item \textbf{Cross-Domain Transfer:}
    Evaluation of domain-specific adapters on all four
    domains without further target-domain training.

    \item \textbf{Sequential Domain Learning:}
    Performance across four training stages,
    comparing learning with and without replay.
    
    \item \textbf{Baseline Performance:}
    Accuracy by model and dataset, separated by answer
    condition and response format.

    \item \textbf{KV-Cache Access Interventions:}
    Performance under restricted attention access to
    the original evidence, the previous answer, or both.

    \item \textbf{CoT Failure Analysis:}
    Examination of generated reasoning when models fail
    to preserve answers justified by the remaining evidence.
\end{enumerate}

\subsection{Experimental Settings}
\label{app:experimental-settings}

\paragraph{Training and Transfer.}
We train separate domain-specific LoRA adapters on
post-retraction gold answers, retaining the base model's
actual initial response $a_{\mathrm{prior}}$.
Loss is applied only to the answer completion and EOS,
without CoT supervision.
We update query and value projections with rank 16,
LoRA scaling parameter 32, and dropout 0.05.
Training uses AdamW at a constant learning rate of
$2\times10^{-4}$ for three epochs, with an effective
batch size of 16 and gradient clipping at 1.0.
The checkpoint maximizing mean validation accuracy across
answer-preserving and answer-changing cases is selected.
For cross-domain evaluation, the adapter is frozen;
target-domain data are not used for training or selection.

\paragraph{Sequential Learning.}
We update one adapter in the fixed order
Priority, bAbI, FOLIO, and ProofWriter.
Each stage uses three epochs and evaluates the final checkpoint
on all four domains.
Adapter weights carry over, while the optimizer is reset.
No replay uses only the current domain; full replay includes
all previously observed training examples.
Initial responses are shared across stages and conditions.
Replay increases the training budget, so the comparison
is not compute-matched.

\paragraph{Activation Steering.}
We estimate directions from training activations at $G$,
contrasting remaining-evidence and answer-verification
instructions.
Directions are unit-normalized and scaled by the standard
deviation of baseline projections.
Without additional instructions, steering is applied from
the retraction message onward and at every decoding step.
We report layer-wise training accuracy changes averaged
over $\alpha\in\{0.5,1\}$.

\paragraph{Attention-Access Interventions.}
We restrict attention to the original evidence $E_0$,
the previous answer $a_{\mathrm{prior}}$, or both,
at every decoder layer and head.
The input follows the schematic structure:
\begin{equation}
\begin{aligned}
&\underbrace{E_0}_{\text{evidence}}
\;\;
\underbrace{q_1,\;a_{\mathrm{prior}}}_{\text{prior interaction}}
\\[-2pt]
&\qquad
\underbrace{\rho(R)}_{\text{withdrawal}}
\;\;
\underbrace{q_2}_{\text{follow-up}}
\;\;
\underbrace{\text{assistant prefix}}_{\text{ends at }t_{\mathrm{last}}}.
\end{aligned}
\label{eq:input-regions}
\end{equation}
Here, $t_{\mathrm{last}}$ is the final input-token position,
whose hidden state predicts the first response token.
System instructions and other chat-template tokens are omitted.

\begin{itemize}
    \item \textbf{Final-position intervention}
    (Table~\ref{tab:dataset-deltas}):
    We block attention only from $t_{\mathrm{last}}$
    to the selected evidence or answer positions.
    Accuracy changes are measured relative to the Direct run baseline, 
    which uses independently generated
    initial histories.

    \item \textbf{Multiple-position interventions}
    (Table~\ref{tab:run7-nine-settings-vllm}):
    We use matched Base/LoRA histories and block attention
    from the last $k$ positions within $\rho(R)$,
    within $q_2$, or ending at $t_{\mathrm{last}}$,
    for $k\in\{1,2,4,8\}$.
    The table averages nine settings:
    three anchors $\times$ $k\in\{1,2,4\}$.
    Changes are measured against each branch's own unmasked
    baseline on a common case set eligible for $k=8$.
\end{itemize}

Tokens and KV tensors are retained.
Each condition uses a fresh prefill, with masking applied
only to the selected input queries.
Subsequent generated queries remain unblocked.

\paragraph{Generation and Evaluation.}
We use greedy decoding and score post-retraction answers
against $y_1$, separately for preservation and change.
Direct and short CoT generation allow 32 and 256 new tokens,
respectively, in the fine-tuning evaluations.
% access-intervention experiments use Direct generation.
% For the extended interventions, 95\% confidence intervals
% use 2,000 paired cluster-bootstrap samples over source groups.

% ============================================================
% ============================================================
% ============================================================

% ============================================================
% Cross-domain transfer: full-width figures at page top
% ============================================================

% ================
\twocolumn[{%
\begin{minipage}{\textwidth}
\centering

\includegraphics[width=0.8\linewidth]
{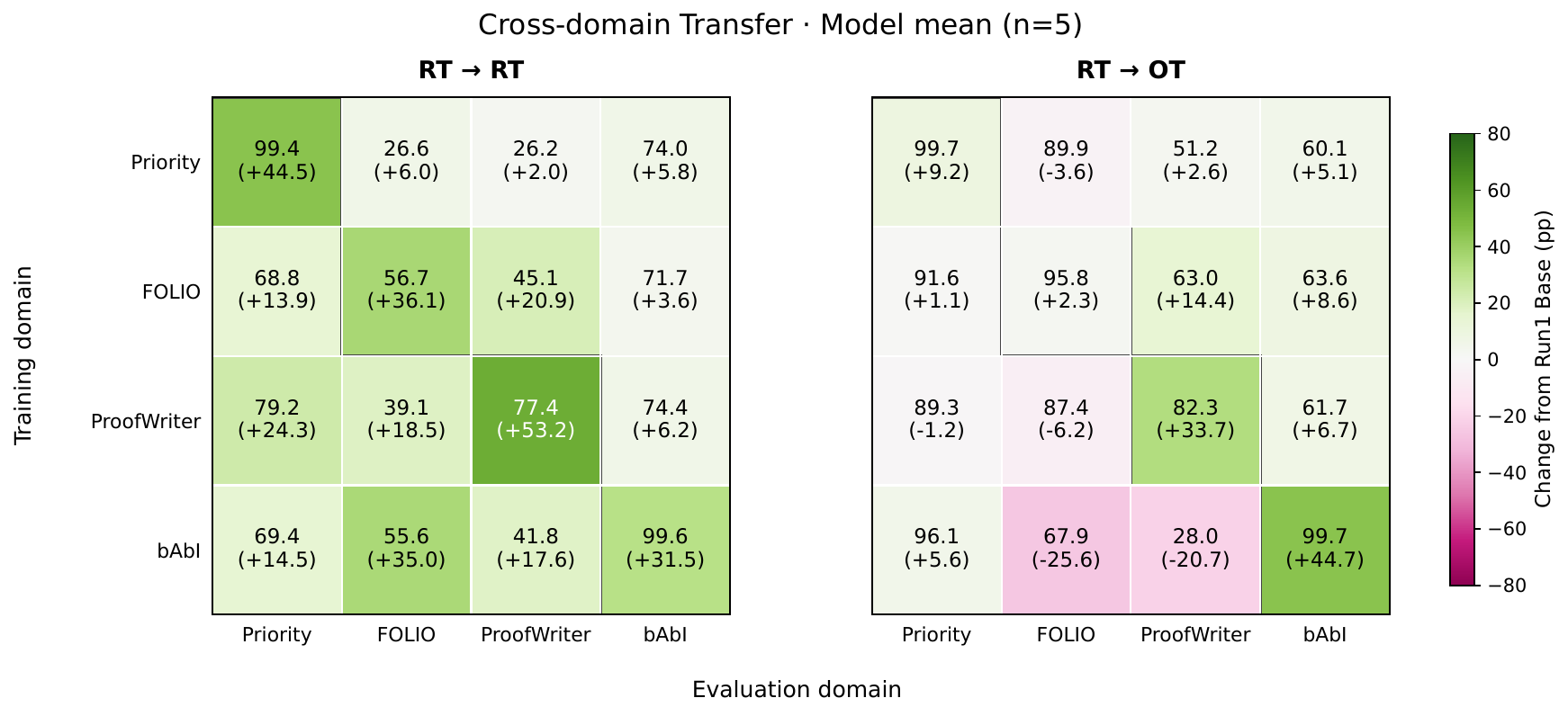}

\medskip

\includegraphics[width=\linewidth]
{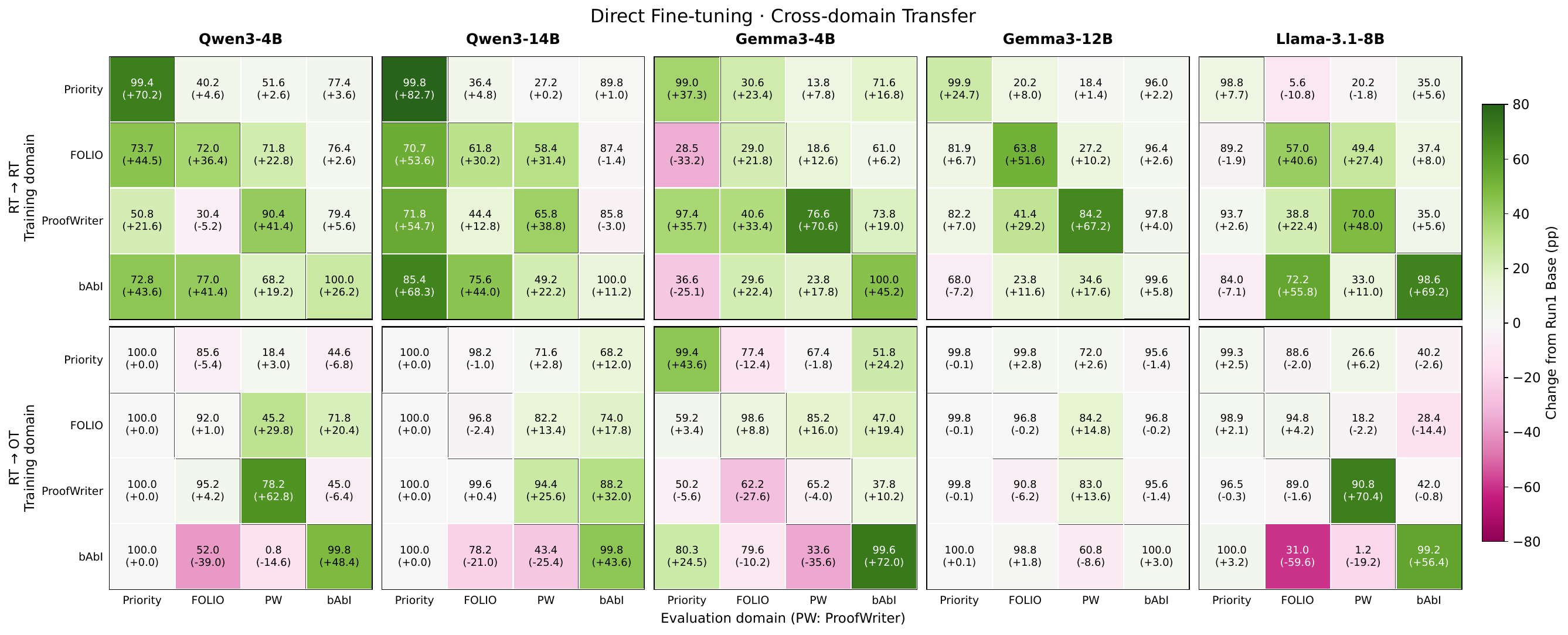}

\captionof{figure}{
Cross-domain Direct accuracy (\%) and changes (pp) from
the main Direct run, averaged across models (top) and by model (bottom).
\textcolor{black}{"Run1 Base" in the color bar denotes this run.}
Diagonal entries are in-domain; off-diagonal entries are transfer.
Adapters are fixed after source-domain selection.
}
\label{fig:fig_main_cross_domain_transfer}

\vspace{1em}
\end{minipage}%
}]
% ================

\subsection{Cross-Domain Transfer}
\label{app:cross-domain-transfer}

Figure~\ref{fig:fig_main_cross_domain_transfer} reports
all source--target combinations, averaged across models
and separately for each model.
Each adapter is trained and selected on its source domain,
then evaluated on all four domains without further training.
Diagonal entries show in-domain performance;
off-diagonal entries measure cross-domain transfer.

In-domain gains are larger and more consistent,
whereas transfer varies across source--target pairs,
models, and answer conditions.
Training on Priority benefits some other domains,
but these gains do not extend uniformly.
The results suggest that targeted training improves revision
within a domain without establishing a general procedure
that transfers reliably across domains.
Changes are measured relative to the corresponding main Direct run baseline, 
whose initial histories are generated independently.

\subsection{Sequential Fine-Tuning Across Domains}
\label{app:sequential-finetuning}

\begin{figure*}[p]
\centering
\includegraphics[width=0.9\linewidth]
{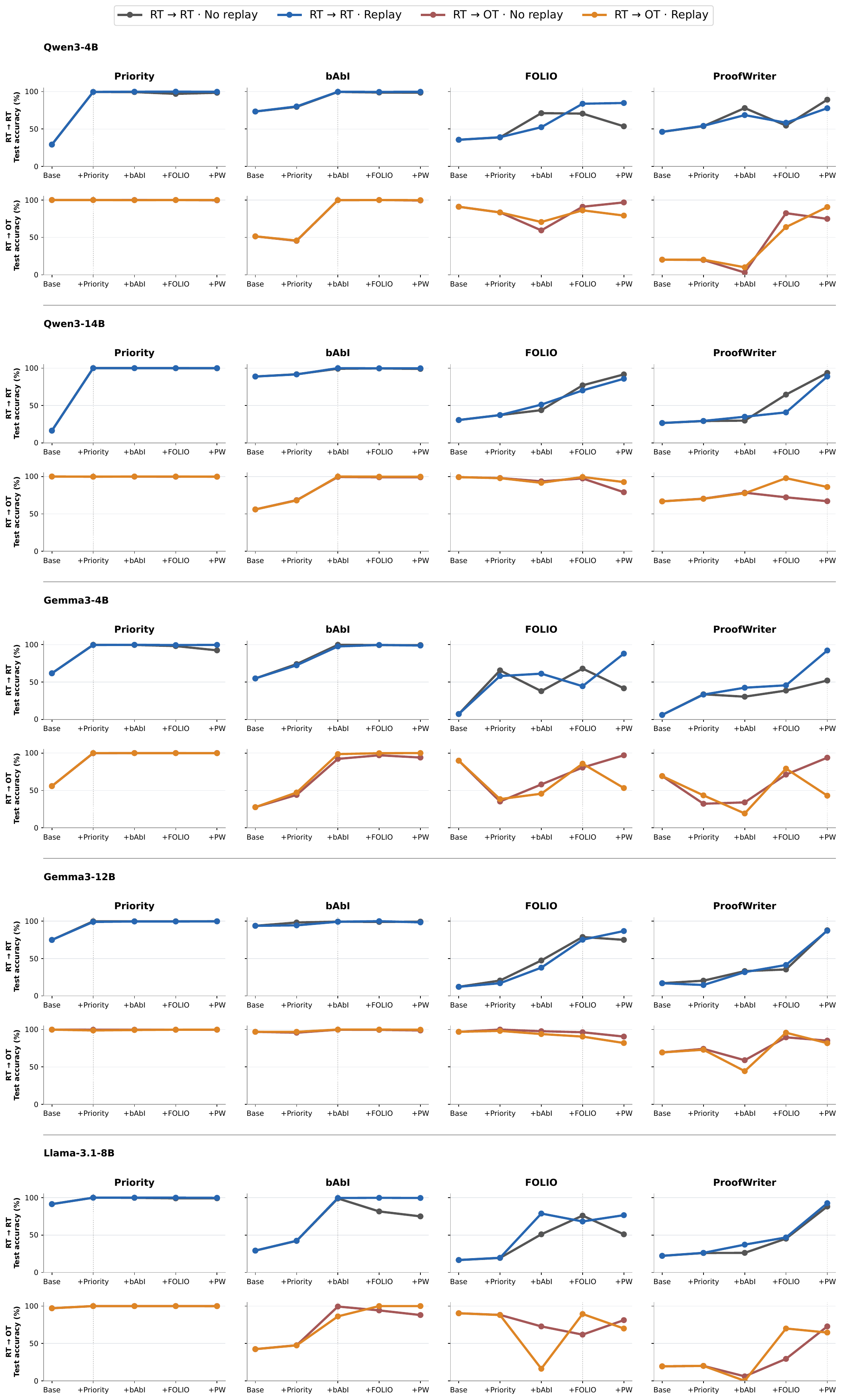}

\caption{
Sequential fine-tuning with and without replay:
model means (top) and individual models (bottom).
All four domains are evaluated after each training stage.
}
\label{fig:learning_curve_model_mean}
\end{figure*}

Figure~\ref{fig:learning_curve_model_mean} reports performance
during sequential fine-tuning, averaged across models
and separately for each model.
A single LoRA adapter is updated in the order
Priority, bAbI, FOLIO, and ProofWriter,
with all four domains evaluated after each stage.

Performance improves under both no-replay and full-replay
training, in answer-preserving and answer-changing cases.
The gains show that sequential training can improve revision
across multiple domains, even without revisiting earlier
training examples.
Full replay mixes all previously observed training examples
with the current domain.
Because it uses more examples and optimizer updates,
the comparison is not compute-matched.
\textcolor{black}{Without replay, accuracy on earlier domains sometimes drops
in later training stages, suggesting some forgetting. Replay mitigates most of these drops, although not in every case.}

\subsection{Model-Wise Performance and Evidence Access}
\label{app:model-wise-results}

\paragraph{Baseline Performance.}
Table~\ref{tab:appendix-run1-1} reports initial and
post-retraction accuracy for each model and dataset.
On Priority, mean Direct accuracy is 46.7\% for
answer-preserving cases, compared with 93.1\% for
answer-changing cases.
CoT raises preservation accuracy to 74.5\%, but substantial
errors remain.
FOLIO and ProofWriter show similar preservation difficulties,
whereas bAbI has a smaller gap between the two conditions.
The model-wise results also show that CoT gains vary
across models and datasets.

\paragraph{Attention-Access Interventions.}
Table~\ref{tab:appendix-run3-datasets-by-model} reports
Direct accuracy across nine LLMs when attention from the
final input position to the evidence, prior answer, or both
is blocked.
\textbf{Access sensitivity varies substantially across
datasets and models.}
Blocking both segments reduces mean bAbI accuracy by
39.3 pp in answer-preserving cases and 37.4 pp in
answer-changing cases, whereas FOLIO changes little on average.
Opposing model-level effects show that small means do not
imply uniformly low sensitivity.
Because the reference run uses a different backend and
independently generated histories, these comparisons are
descriptive rather than matched-history causal effects.

%  ============================================================================================================

\subsection{CoT-based Explanation}
\label{app:cot_generation_examples}
To illustrate how LLMs respond to support retraction, we examine their
follow-up answers and accompanying CoT reasons in concrete examples.
These examples highlight errors that occur even when LLMs answer the
initial question correctly.

Tables~\ref{tab:apdx-run1-example-priority}
and~\ref{tab:apdx-run1-example-proofwriter} present one base case each
from Priority and ProofWriter for Qwen3-14B, Qwen3-32B, Gemma3-12B,
Gemma3-27B, and Llama-3.1-8B. Among cases in which all five LLMs answer
the initial question correctly and each condition retains at least one
correct response for comparison, we select the case ranked highest by
the number of incorrect follow-up answers.
Incorrect answers are highlighted, and CoT reasons are shown as
verbatim excerpts.

\begin{table*}[p]
\centering
\small
\setlength{\tabcolsep}{3pt}
\renewcommand{\arraystretch}{1.50}
\setlength{\aboverulesep}{0pt}\setlength{\belowrulesep}{0pt}
\resizebox{\textwidth}{!}{%
\begin{tabular}{@{}l|l|c|ccc|ccc@{}}
\toprule
\multirow{3}{*}{Dataset} & \multirow{3}{*}{Model} & \multirow{3}{*}{\shortstack{Initial Answer\\(Label: RT)}} & \multicolumn{6}{c}{\shortstack{\rule{0pt}{2.6ex}Follow-up Evaluation\\(After Evidence Retraction)\rule[-0.8ex]{0pt}{1ex}}} \\
\cline{4-9}
 & & & \multicolumn{3}{c|}{Direct Answer Format} & \multicolumn{3}{c}{CoT Answer Format} \\
\cline{4-9}
 & & & \shortstack{\rule{0pt}{2.8ex}Overall\\(all cases)\rule[-0.6ex]{0pt}{1ex}} & \shortstack{\rule{0pt}{2.8ex}RT $\to$ RT\\(label preserved)\rule[-0.6ex]{0pt}{1ex}} & \shortstack{\rule{0pt}{2.8ex}RT $\to$ OT\\(label changed)\rule[-0.6ex]{0pt}{1ex}} & \shortstack{\rule{0pt}{2.8ex}Overall\\(all cases)\rule[-0.6ex]{0pt}{1ex}} & \shortstack{\rule{0pt}{2.8ex}RT $\to$ RT\\(label preserved)\rule[-0.6ex]{0pt}{1ex}} & \shortstack{\rule{0pt}{2.8ex}RT $\to$ OT\\(label changed)\rule[-0.6ex]{0pt}{1ex}} \\
\midrule
\multirow{11}{*}{Priority} & OLMo-2-7B & 100.0 & \makebox[2.6em][r]{44.0}\,\textcolor{red!80!black}{\makebox[5.6em][r]{($\Delta -56.0$)}} & \makebox[2.6em][r]{\textbf{0.4}}\,\textcolor{red!80!black}{\makebox[5.6em][r]{($\Delta -99.6$)}} & \makebox[2.6em][r]{87.5}\,\textcolor{red!80!black}{\makebox[5.6em][r]{($\Delta -12.5$)}} & \makebox[2.6em][r]{56.3}\,\textcolor{red!80!black}{\makebox[5.6em][r]{($\Delta -43.7$)}} & \makebox[2.6em][r]{\textbf{13.4}}\,\textcolor{red!80!black}{\makebox[5.6em][r]{($\Delta -86.6$)}} & \makebox[2.6em][r]{99.2}\,\textcolor{red!80!black}{\makebox[5.6em][r]{($\Delta -0.8$)}} \\
 & OLMo-3-7B & 100.0 & \makebox[2.6em][r]{49.5}\,\textcolor{red!80!black}{\makebox[5.6em][r]{($\Delta -50.4$)}} & \makebox[2.6em][r]{\textbf{0.2}}\,\textcolor{red!80!black}{\makebox[5.6em][r]{($\Delta -99.8$)}} & \makebox[2.6em][r]{98.9}\,\textcolor{red!80!black}{\makebox[5.6em][r]{($\Delta -1.1$)}} & \makebox[2.6em][r]{80.0}\,\textcolor{red!80!black}{\makebox[5.6em][r]{($\Delta -20.0$)}} & \makebox[2.6em][r]{\textbf{60.6}}\,\textcolor{red!80!black}{\makebox[5.6em][r]{($\Delta -39.4$)}} & \makebox[2.6em][r]{99.4}\,\textcolor{red!80!black}{\makebox[5.6em][r]{($\Delta -0.6$)}} \\
 & Qwen3-4B & 100.0 & \makebox[2.6em][r]{64.6}\,\textcolor{red!80!black}{\makebox[5.6em][r]{($\Delta -35.4$)}} & \makebox[2.6em][r]{\textbf{29.2}}\,\textcolor{red!80!black}{\makebox[5.6em][r]{($\Delta -70.8$)}} & \makebox[2.6em][r]{100.0}\,\textcolor{green!45!black}{\makebox[5.6em][r]{($\Delta +0.1$)}} & \makebox[2.6em][r]{97.2}\,\textcolor{red!80!black}{\makebox[5.6em][r]{($\Delta -2.8$)}} & \makebox[2.6em][r]{\textbf{95.0}}\,\textcolor{red!80!black}{\makebox[5.6em][r]{($\Delta -5.0$)}} & \makebox[2.6em][r]{99.3}\,\textcolor{red!80!black}{\makebox[5.6em][r]{($\Delta -0.6$)}} \\
 & Qwen3-14B & 100.0 & \makebox[2.6em][r]{58.6}\,\textcolor{red!80!black}{\makebox[5.6em][r]{($\Delta -41.4$)}} & \makebox[2.6em][r]{\textbf{17.1}}\,\textcolor{red!80!black}{\makebox[5.6em][r]{($\Delta -82.9$)}} & \makebox[2.6em][r]{100.0}\,\textcolor{black}{\makebox[5.6em][r]{($\Delta +0.0$)}} & \makebox[2.6em][r]{98.8}\,\textcolor{red!80!black}{\makebox[5.6em][r]{($\Delta -1.2$)}} & \makebox[2.6em][r]{\textbf{97.5}}\,\textcolor{red!80!black}{\makebox[5.6em][r]{($\Delta -2.5$)}} & \makebox[2.6em][r]{100.0}\,\textcolor{black}{\makebox[5.6em][r]{($\Delta +0.0$)}} \\
 & Qwen3-32B & 100.0 & \makebox[2.6em][r]{98.9}\,\textcolor{red!80!black}{\makebox[5.6em][r]{($\Delta -1.1$)}} & \makebox[2.6em][r]{\textbf{97.8}}\,\textcolor{red!80!black}{\makebox[5.6em][r]{($\Delta -2.2$)}} & \makebox[2.6em][r]{100.0}\,\textcolor{black}{\makebox[5.6em][r]{($\Delta +0.0$)}} & \makebox[2.6em][r]{99.2}\,\textcolor{red!80!black}{\makebox[5.6em][r]{($\Delta -0.8$)}} & \makebox[2.6em][r]{\textbf{98.3}}\,\textcolor{red!80!black}{\makebox[5.6em][r]{($\Delta -1.7$)}} & \makebox[2.6em][r]{100.0}\,\textcolor{black}{\makebox[5.6em][r]{($\Delta +0.0$)}} \\
 & Gemma3-4B & 100.0 & \makebox[2.6em][r]{58.8}\,\textcolor{red!80!black}{\makebox[5.6em][r]{($\Delta -41.2$)}} & \makebox[2.6em][r]{61.7}\,\textcolor{red!80!black}{\makebox[5.6em][r]{($\Delta -38.3$)}} & \makebox[2.6em][r]{\textbf{55.8}}\,\textcolor{red!80!black}{\makebox[5.6em][r]{($\Delta -44.2$)}} & \makebox[2.6em][r]{70.3}\,\textcolor{red!80!black}{\makebox[5.6em][r]{($\Delta -29.6$)}} & \makebox[2.6em][r]{97.5}\,\textcolor{red!80!black}{\makebox[5.6em][r]{($\Delta -2.5$)}} & \makebox[2.6em][r]{\textbf{43.2}}\,\textcolor{red!80!black}{\makebox[5.6em][r]{($\Delta -56.8$)}} \\
 & Gemma3-12B & 100.0 & \makebox[2.6em][r]{87.5}\,\textcolor{red!80!black}{\makebox[5.6em][r]{($\Delta -12.5$)}} & \makebox[2.6em][r]{\textbf{75.2}}\,\textcolor{red!80!black}{\makebox[5.6em][r]{($\Delta -24.8$)}} & \makebox[2.6em][r]{99.9}\,\textcolor{red!80!black}{\makebox[5.6em][r]{($\Delta -0.1$)}} & \makebox[2.6em][r]{93.0}\,\textcolor{red!80!black}{\makebox[5.6em][r]{($\Delta -7.1$)}} & \makebox[2.6em][r]{\textbf{86.0}}\,\textcolor{red!80!black}{\makebox[5.6em][r]{($\Delta -14.0$)}} & \makebox[2.6em][r]{99.9}\,\textcolor{red!80!black}{\makebox[5.6em][r]{($\Delta -0.1$)}} \\
 & Gemma3-27B & 100.0 & \makebox[2.6em][r]{94.5}\,\textcolor{red!80!black}{\makebox[5.6em][r]{($\Delta -5.5$)}} & \makebox[2.6em][r]{\textbf{89.3}}\,\textcolor{red!80!black}{\makebox[5.6em][r]{($\Delta -10.7$)}} & \makebox[2.6em][r]{99.7}\,\textcolor{red!80!black}{\makebox[5.6em][r]{($\Delta -0.3$)}} & \makebox[2.6em][r]{96.6}\,\textcolor{red!80!black}{\makebox[5.6em][r]{($\Delta -3.4$)}} & \makebox[2.6em][r]{\textbf{93.4}}\,\textcolor{red!80!black}{\makebox[5.6em][r]{($\Delta -6.6$)}} & \makebox[2.6em][r]{99.8}\,\textcolor{red!80!black}{\makebox[5.6em][r]{($\Delta -0.2$)}} \\
 & Llama-3.1-8B & 99.7 & \makebox[2.6em][r]{94.0}\,\textcolor{red!80!black}{\makebox[5.6em][r]{($\Delta -5.7$)}} & \makebox[2.6em][r]{\textbf{91.1}}\,\textcolor{red!80!black}{\makebox[5.6em][r]{($\Delta -8.9$)}} & \makebox[2.6em][r]{96.8}\,\textcolor{red!80!black}{\makebox[5.6em][r]{($\Delta -2.5$)}} & \makebox[2.6em][r]{87.9}\,\textcolor{red!80!black}{\makebox[5.6em][r]{($\Delta -11.8$)}} & \makebox[2.6em][r]{94.3}\,\textcolor{red!80!black}{\makebox[5.6em][r]{($\Delta -5.7$)}} & \makebox[2.6em][r]{\textbf{81.5}}\,\textcolor{red!80!black}{\makebox[5.6em][r]{($\Delta -17.8$)}} \\
 & Mistral-7B-v0.3 & 100.0 & \makebox[2.6em][r]{49.0}\,\textcolor{red!80!black}{\makebox[5.6em][r]{($\Delta -51.0$)}} & \makebox[2.6em][r]{\textbf{5.4}}\,\textcolor{red!80!black}{\makebox[5.6em][r]{($\Delta -94.6$)}} & \makebox[2.6em][r]{92.6}\,\textcolor{red!80!black}{\makebox[5.6em][r]{($\Delta -7.4$)}} & \makebox[2.6em][r]{52.4}\,\textcolor{red!80!black}{\makebox[5.6em][r]{($\Delta -47.6$)}} & \makebox[2.6em][r]{\textbf{9.0}}\,\textcolor{red!80!black}{\makebox[5.6em][r]{($\Delta -91.0$)}} & \makebox[2.6em][r]{95.8}\,\textcolor{red!80!black}{\makebox[5.6em][r]{($\Delta -4.2$)}} \\
\cline{2-9}
 & \textbf{Model mean} & 100.0 & \makebox[2.6em][r]{69.9}\,\textcolor{red!80!black}{\makebox[5.6em][r]{($\Delta -30.0$)}} & \makebox[2.6em][r]{\textbf{46.7}}\,\textcolor{red!80!black}{\makebox[5.6em][r]{($\Delta -53.3$)}} & \makebox[2.6em][r]{93.1}\,\textcolor{red!80!black}{\makebox[5.6em][r]{($\Delta -6.8$)}} & \makebox[2.6em][r]{83.2}\,\textcolor{red!80!black}{\makebox[5.6em][r]{($\Delta -16.8$)}} & \makebox[2.6em][r]{\textbf{74.5}}\,\textcolor{red!80!black}{\makebox[5.6em][r]{($\Delta -25.5$)}} & \makebox[2.6em][r]{91.8}\,\textcolor{red!80!black}{\makebox[5.6em][r]{($\Delta -8.1$)}} \\
\midrule
\multirow{11}{*}{bAbI} & OLMo-2-7B & 48.4 & \makebox[2.6em][r]{52.6}\,\textcolor{green!45!black}{\makebox[5.6em][r]{($\Delta +4.2$)}} & \makebox[2.6em][r]{\textbf{52.0}}\,\textcolor{green!45!black}{\makebox[5.6em][r]{($\Delta +3.6$)}} & \makebox[2.6em][r]{53.2}\,\textcolor{green!45!black}{\makebox[5.6em][r]{($\Delta +4.8$)}} & \makebox[2.6em][r]{51.5}\,\textcolor{green!45!black}{\makebox[5.6em][r]{($\Delta +3.1$)}} & \makebox[2.6em][r]{52.4}\,\textcolor{green!45!black}{\makebox[5.6em][r]{($\Delta +4.0$)}} & \makebox[2.6em][r]{\textbf{50.6}}\,\textcolor{green!45!black}{\makebox[5.6em][r]{($\Delta +2.2$)}} \\
 & OLMo-3-7B & 76.2 & \makebox[2.6em][r]{48.3}\,\textcolor{red!80!black}{\makebox[5.6em][r]{($\Delta -27.9$)}} & \makebox[2.6em][r]{64.0}\,\textcolor{red!80!black}{\makebox[5.6em][r]{($\Delta -12.2$)}} & \makebox[2.6em][r]{\textbf{32.6}}\,\textcolor{red!80!black}{\makebox[5.6em][r]{($\Delta -43.6$)}} & \makebox[2.6em][r]{82.1}\,\textcolor{green!45!black}{\makebox[5.6em][r]{($\Delta +5.9$)}} & \makebox[2.6em][r]{82.6}\,\textcolor{green!45!black}{\makebox[5.6em][r]{($\Delta +6.4$)}} & \makebox[2.6em][r]{\textbf{81.6}}\,\textcolor{green!45!black}{\makebox[5.6em][r]{($\Delta +5.4$)}} \\
 & Qwen3-4B & 75.4 & \makebox[2.6em][r]{62.6}\,\textcolor{red!80!black}{\makebox[5.6em][r]{($\Delta -12.8$)}} & \makebox[2.6em][r]{73.8}\,\textcolor{red!80!black}{\makebox[5.6em][r]{($\Delta -1.6$)}} & \makebox[2.6em][r]{\textbf{51.4}}\,\textcolor{red!80!black}{\makebox[5.6em][r]{($\Delta -24.0$)}} & \makebox[2.6em][r]{91.2}\,\textcolor{green!45!black}{\makebox[5.6em][r]{($\Delta +15.8$)}} & \makebox[2.6em][r]{\textbf{90.0}}\,\textcolor{green!45!black}{\makebox[5.6em][r]{($\Delta +14.6$)}} & \makebox[2.6em][r]{92.4}\,\textcolor{green!45!black}{\makebox[5.6em][r]{($\Delta +17.0$)}} \\
 & Qwen3-14B & 94.0 & \makebox[2.6em][r]{72.5}\,\textcolor{red!80!black}{\makebox[5.6em][r]{($\Delta -21.5$)}} & \makebox[2.6em][r]{88.8}\,\textcolor{red!80!black}{\makebox[5.6em][r]{($\Delta -5.2$)}} & \makebox[2.6em][r]{\textbf{56.2}}\,\textcolor{red!80!black}{\makebox[5.6em][r]{($\Delta -37.8$)}} & \makebox[2.6em][r]{95.1}\,\textcolor{green!45!black}{\makebox[5.6em][r]{($\Delta +1.1$)}} & \makebox[2.6em][r]{\textbf{94.6}}\,\textcolor{green!45!black}{\makebox[5.6em][r]{($\Delta +0.6$)}} & \makebox[2.6em][r]{95.6}\,\textcolor{green!45!black}{\makebox[5.6em][r]{($\Delta +1.6$)}} \\
 & Qwen3-32B & 96.2 & \makebox[2.6em][r]{83.0}\,\textcolor{red!80!black}{\makebox[5.6em][r]{($\Delta -13.2$)}} & \makebox[2.6em][r]{\textbf{80.2}}\,\textcolor{red!80!black}{\makebox[5.6em][r]{($\Delta -16.0$)}} & \makebox[2.6em][r]{85.8}\,\textcolor{red!80!black}{\makebox[5.6em][r]{($\Delta -10.4$)}} & \makebox[2.6em][r]{97.5}\,\textcolor{green!45!black}{\makebox[5.6em][r]{($\Delta +1.3$)}} & \makebox[2.6em][r]{\textbf{95.6}}\,\textcolor{red!80!black}{\makebox[5.6em][r]{($\Delta -0.6$)}} & \makebox[2.6em][r]{99.4}\,\textcolor{green!45!black}{\makebox[5.6em][r]{($\Delta +3.2$)}} \\
 & Gemma3-4B & 65.0 & \makebox[2.6em][r]{41.2}\,\textcolor{red!80!black}{\makebox[5.6em][r]{($\Delta -23.8$)}} & \makebox[2.6em][r]{54.8}\,\textcolor{red!80!black}{\makebox[5.6em][r]{($\Delta -10.2$)}} & \makebox[2.6em][r]{\textbf{27.6}}\,\textcolor{red!80!black}{\makebox[5.6em][r]{($\Delta -37.4$)}} & \makebox[2.6em][r]{60.3}\,\textcolor{red!80!black}{\makebox[5.6em][r]{($\Delta -4.7$)}} & \makebox[2.6em][r]{71.0}\,\textcolor{green!45!black}{\makebox[5.6em][r]{($\Delta +6.0$)}} & \makebox[2.6em][r]{\textbf{49.6}}\,\textcolor{red!80!black}{\makebox[5.6em][r]{($\Delta -15.4$)}} \\
 & Gemma3-12B & 98.8 & \makebox[2.6em][r]{95.4}\,\textcolor{red!80!black}{\makebox[5.6em][r]{($\Delta -3.4$)}} & \makebox[2.6em][r]{\textbf{93.8}}\,\textcolor{red!80!black}{\makebox[5.6em][r]{($\Delta -5.0$)}} & \makebox[2.6em][r]{97.0}\,\textcolor{red!80!black}{\makebox[5.6em][r]{($\Delta -1.8$)}} & \makebox[2.6em][r]{97.5}\,\textcolor{red!80!black}{\makebox[5.6em][r]{($\Delta -1.3$)}} & \makebox[2.6em][r]{\textbf{97.2}}\,\textcolor{red!80!black}{\makebox[5.6em][r]{($\Delta -1.6$)}} & \makebox[2.6em][r]{97.8}\,\textcolor{red!80!black}{\makebox[5.6em][r]{($\Delta -1.0$)}} \\
 & Gemma3-27B & 97.8 & \makebox[2.6em][r]{92.7}\,\textcolor{red!80!black}{\makebox[5.6em][r]{($\Delta -5.1$)}} & \makebox[2.6em][r]{\textbf{88.0}}\,\textcolor{red!80!black}{\makebox[5.6em][r]{($\Delta -9.8$)}} & \makebox[2.6em][r]{97.4}\,\textcolor{red!80!black}{\makebox[5.6em][r]{($\Delta -0.4$)}} & \makebox[2.6em][r]{97.9}\,\textcolor{green!45!black}{\makebox[5.6em][r]{($\Delta +0.1$)}} & \makebox[2.6em][r]{\textbf{97.4}}\,\textcolor{red!80!black}{\makebox[5.6em][r]{($\Delta -0.4$)}} & \makebox[2.6em][r]{98.4}\,\textcolor{green!45!black}{\makebox[5.6em][r]{($\Delta +0.6$)}} \\
 & Llama-3.1-8B & 49.2 & \makebox[2.6em][r]{36.1}\,\textcolor{red!80!black}{\makebox[5.6em][r]{($\Delta -13.1$)}} & \makebox[2.6em][r]{\textbf{29.4}}\,\textcolor{red!80!black}{\makebox[5.6em][r]{($\Delta -19.8$)}} & \makebox[2.6em][r]{42.8}\,\textcolor{red!80!black}{\makebox[5.6em][r]{($\Delta -6.4$)}} & \makebox[2.6em][r]{62.8}\,\textcolor{green!45!black}{\makebox[5.6em][r]{($\Delta +13.6$)}} & \makebox[2.6em][r]{\textbf{56.0}}\,\textcolor{green!45!black}{\makebox[5.6em][r]{($\Delta +6.8$)}} & \makebox[2.6em][r]{69.6}\,\textcolor{green!45!black}{\makebox[5.6em][r]{($\Delta +20.4$)}} \\
 & Mistral-7B-v0.3 & 59.6 & \makebox[2.6em][r]{19.9}\,\textcolor{red!80!black}{\makebox[5.6em][r]{($\Delta -39.7$)}} & \makebox[2.6em][r]{\textbf{19.8}}\,\textcolor{red!80!black}{\makebox[5.6em][r]{($\Delta -39.8$)}} & \makebox[2.6em][r]{20.0}\,\textcolor{red!80!black}{\makebox[5.6em][r]{($\Delta -39.6$)}} & \makebox[2.6em][r]{63.6}\,\textcolor{green!45!black}{\makebox[5.6em][r]{($\Delta +4.0$)}} & \makebox[2.6em][r]{\textbf{60.8}}\,\textcolor{green!45!black}{\makebox[5.6em][r]{($\Delta +1.2$)}} & \makebox[2.6em][r]{66.4}\,\textcolor{green!45!black}{\makebox[5.6em][r]{($\Delta +6.8$)}} \\
\cline{2-9}
 & \textbf{Model mean} & 76.1 & \makebox[2.6em][r]{60.4}\,\textcolor{red!80!black}{\makebox[5.6em][r]{($\Delta -15.6$)}} & \makebox[2.6em][r]{64.5}\,\textcolor{red!80!black}{\makebox[5.6em][r]{($\Delta -11.6$)}} & \makebox[2.6em][r]{\textbf{56.4}}\,\textcolor{red!80!black}{\makebox[5.6em][r]{($\Delta -19.7$)}} & \makebox[2.6em][r]{80.0}\,\textcolor{green!45!black}{\makebox[5.6em][r]{($\Delta +3.9$)}} & \makebox[2.6em][r]{\textbf{79.8}}\,\textcolor{green!45!black}{\makebox[5.6em][r]{($\Delta +3.7$)}} & \makebox[2.6em][r]{80.1}\,\textcolor{green!45!black}{\makebox[5.6em][r]{($\Delta +4.1$)}} \\
\midrule
\multirow{11}{*}{FOLIO} & OLMo-2-7B & 75.2 & \makebox[2.6em][r]{40.7}\,\textcolor{red!80!black}{\makebox[5.6em][r]{($\Delta -34.5$)}} & \makebox[2.6em][r]{60.2}\,\textcolor{red!80!black}{\makebox[5.6em][r]{($\Delta -15.0$)}} & \makebox[2.6em][r]{\textbf{21.2}}\,\textcolor{red!80!black}{\makebox[5.6em][r]{($\Delta -54.0$)}} & \makebox[2.6em][r]{45.0}\,\textcolor{red!80!black}{\makebox[5.6em][r]{($\Delta -30.2$)}} & \makebox[2.6em][r]{\textbf{28.2}}\,\textcolor{red!80!black}{\makebox[5.6em][r]{($\Delta -47.0$)}} & \makebox[2.6em][r]{61.8}\,\textcolor{red!80!black}{\makebox[5.6em][r]{($\Delta -13.4$)}} \\
 & OLMo-3-7B & 56.0 & \makebox[2.6em][r]{54.3}\,\textcolor{red!80!black}{\makebox[5.6em][r]{($\Delta -1.7$)}} & \makebox[2.6em][r]{\textbf{47.0}}\,\textcolor{red!80!black}{\makebox[5.6em][r]{($\Delta -9.0$)}} & \makebox[2.6em][r]{61.6}\,\textcolor{green!45!black}{\makebox[5.6em][r]{($\Delta +5.6$)}} & \makebox[2.6em][r]{63.5}\,\textcolor{green!45!black}{\makebox[5.6em][r]{($\Delta +7.5$)}} & \makebox[2.6em][r]{\textbf{46.8}}\,\textcolor{red!80!black}{\makebox[5.6em][r]{($\Delta -9.2$)}} & \makebox[2.6em][r]{80.2}\,\textcolor{green!45!black}{\makebox[5.6em][r]{($\Delta +24.2$)}} \\
 & Qwen3-4B & 41.4 & \makebox[2.6em][r]{63.3}\,\textcolor{green!45!black}{\makebox[5.6em][r]{($\Delta +21.9$)}} & \makebox[2.6em][r]{\textbf{35.6}}\,\textcolor{red!80!black}{\makebox[5.6em][r]{($\Delta -5.8$)}} & \makebox[2.6em][r]{91.0}\,\textcolor{green!45!black}{\makebox[5.6em][r]{($\Delta +49.6$)}} & \makebox[2.6em][r]{74.1}\,\textcolor{green!45!black}{\makebox[5.6em][r]{($\Delta +32.7$)}} & \makebox[2.6em][r]{\textbf{62.2}}\,\textcolor{green!45!black}{\makebox[5.6em][r]{($\Delta +20.8$)}} & \makebox[2.6em][r]{86.0}\,\textcolor{green!45!black}{\makebox[5.6em][r]{($\Delta +44.6$)}} \\
 & Qwen3-14B & 70.8 & \makebox[2.6em][r]{65.4}\,\textcolor{red!80!black}{\makebox[5.6em][r]{($\Delta -5.4$)}} & \makebox[2.6em][r]{\textbf{31.6}}\,\textcolor{red!80!black}{\makebox[5.6em][r]{($\Delta -39.2$)}} & \makebox[2.6em][r]{99.2}\,\textcolor{green!45!black}{\makebox[5.6em][r]{($\Delta +28.4$)}} & \makebox[2.6em][r]{80.8}\,\textcolor{green!45!black}{\makebox[5.6em][r]{($\Delta +10.0$)}} & \makebox[2.6em][r]{\textbf{66.0}}\,\textcolor{red!80!black}{\makebox[5.6em][r]{($\Delta -4.8$)}} & \makebox[2.6em][r]{95.6}\,\textcolor{green!45!black}{\makebox[5.6em][r]{($\Delta +24.8$)}} \\
 & Qwen3-32B & 82.2 & \makebox[2.6em][r]{68.1}\,\textcolor{red!80!black}{\makebox[5.6em][r]{($\Delta -14.1$)}} & \makebox[2.6em][r]{\textbf{40.8}}\,\textcolor{red!80!black}{\makebox[5.6em][r]{($\Delta -41.4$)}} & \makebox[2.6em][r]{95.4}\,\textcolor{green!45!black}{\makebox[5.6em][r]{($\Delta +13.2$)}} & \makebox[2.6em][r]{78.9}\,\textcolor{red!80!black}{\makebox[5.6em][r]{($\Delta -3.3$)}} & \makebox[2.6em][r]{\textbf{62.6}}\,\textcolor{red!80!black}{\makebox[5.6em][r]{($\Delta -19.6$)}} & \makebox[2.6em][r]{95.2}\,\textcolor{green!45!black}{\makebox[5.6em][r]{($\Delta +13.0$)}} \\
 & Gemma3-4B & 42.0 & \makebox[2.6em][r]{48.5}\,\textcolor{green!45!black}{\makebox[5.6em][r]{($\Delta +6.5$)}} & \makebox[2.6em][r]{\textbf{7.2}}\,\textcolor{red!80!black}{\makebox[5.6em][r]{($\Delta -34.8$)}} & \makebox[2.6em][r]{89.8}\,\textcolor{green!45!black}{\makebox[5.6em][r]{($\Delta +47.8$)}} & \makebox[2.6em][r]{52.5}\,\textcolor{green!45!black}{\makebox[5.6em][r]{($\Delta +10.5$)}} & \makebox[2.6em][r]{55.6}\,\textcolor{green!45!black}{\makebox[5.6em][r]{($\Delta +13.6$)}} & \makebox[2.6em][r]{\textbf{49.4}}\,\textcolor{green!45!black}{\makebox[5.6em][r]{($\Delta +7.4$)}} \\
 & Gemma3-12B & 54.8 & \makebox[2.6em][r]{54.6}\,\textcolor{red!80!black}{\makebox[5.6em][r]{($\Delta -0.2$)}} & \makebox[2.6em][r]{\textbf{12.2}}\,\textcolor{red!80!black}{\makebox[5.6em][r]{($\Delta -42.6$)}} & \makebox[2.6em][r]{97.0}\,\textcolor{green!45!black}{\makebox[5.6em][r]{($\Delta +42.2$)}} & \makebox[2.6em][r]{77.1}\,\textcolor{green!45!black}{\makebox[5.6em][r]{($\Delta +22.3$)}} & \makebox[2.6em][r]{\textbf{57.6}}\,\textcolor{green!45!black}{\makebox[5.6em][r]{($\Delta +2.8$)}} & \makebox[2.6em][r]{96.6}\,\textcolor{green!45!black}{\makebox[5.6em][r]{($\Delta +41.8$)}} \\
 & Gemma3-27B & 76.6 & \makebox[2.6em][r]{61.8}\,\textcolor{red!80!black}{\makebox[5.6em][r]{($\Delta -14.8$)}} & \makebox[2.6em][r]{\textbf{26.4}}\,\textcolor{red!80!black}{\makebox[5.6em][r]{($\Delta -50.2$)}} & \makebox[2.6em][r]{97.2}\,\textcolor{green!45!black}{\makebox[5.6em][r]{($\Delta +20.6$)}} & \makebox[2.6em][r]{79.7}\,\textcolor{green!45!black}{\makebox[5.6em][r]{($\Delta +3.1$)}} & \makebox[2.6em][r]{\textbf{74.0}}\,\textcolor{red!80!black}{\makebox[5.6em][r]{($\Delta -2.6$)}} & \makebox[2.6em][r]{85.4}\,\textcolor{green!45!black}{\makebox[5.6em][r]{($\Delta +8.8$)}} \\
 & Llama-3.1-8B & 83.4 & \makebox[2.6em][r]{53.5}\,\textcolor{red!80!black}{\makebox[5.6em][r]{($\Delta -29.9$)}} & \makebox[2.6em][r]{\textbf{16.4}}\,\textcolor{red!80!black}{\makebox[5.6em][r]{($\Delta -67.0$)}} & \makebox[2.6em][r]{90.6}\,\textcolor{green!45!black}{\makebox[5.6em][r]{($\Delta +7.2$)}} & \makebox[2.6em][r]{60.2}\,\textcolor{red!80!black}{\makebox[5.6em][r]{($\Delta -23.2$)}} & \makebox[2.6em][r]{\textbf{39.2}}\,\textcolor{red!80!black}{\makebox[5.6em][r]{($\Delta -44.2$)}} & \makebox[2.6em][r]{81.2}\,\textcolor{red!80!black}{\makebox[5.6em][r]{($\Delta -2.2$)}} \\
 & Mistral-7B-v0.3 & 48.8 & \makebox[2.6em][r]{50.7}\,\textcolor{green!45!black}{\makebox[5.6em][r]{($\Delta +1.9$)}} & \makebox[2.6em][r]{\textbf{31.2}}\,\textcolor{red!80!black}{\makebox[5.6em][r]{($\Delta -17.6$)}} & \makebox[2.6em][r]{70.2}\,\textcolor{green!45!black}{\makebox[5.6em][r]{($\Delta +21.4$)}} & \makebox[2.6em][r]{60.1}\,\textcolor{green!45!black}{\makebox[5.6em][r]{($\Delta +11.3$)}} & \makebox[2.6em][r]{\textbf{43.0}}\,\textcolor{red!80!black}{\makebox[5.6em][r]{($\Delta -5.8$)}} & \makebox[2.6em][r]{77.2}\,\textcolor{green!45!black}{\makebox[5.6em][r]{($\Delta +28.4$)}} \\
\cline{2-9}
 & \textbf{Model mean} & 63.1 & \makebox[2.6em][r]{56.1}\,\textcolor{red!80!black}{\makebox[5.6em][r]{($\Delta -7.0$)}} & \makebox[2.6em][r]{\textbf{30.9}}\,\textcolor{red!80!black}{\makebox[5.6em][r]{($\Delta -32.3$)}} & \makebox[2.6em][r]{81.3}\,\textcolor{green!45!black}{\makebox[5.6em][r]{($\Delta +18.2$)}} & \makebox[2.6em][r]{67.2}\,\textcolor{green!45!black}{\makebox[5.6em][r]{($\Delta +4.1$)}} & \makebox[2.6em][r]{\textbf{53.5}}\,\textcolor{red!80!black}{\makebox[5.6em][r]{($\Delta -9.6$)}} & \makebox[2.6em][r]{80.9}\,\textcolor{green!45!black}{\makebox[5.6em][r]{($\Delta +17.7$)}} \\
\midrule
\multirow{11}{*}{ProofWriter} & OLMo-2-7B & 54.0 & \makebox[2.6em][r]{36.9}\,\textcolor{red!80!black}{\makebox[5.6em][r]{($\Delta -17.1$)}} & \makebox[2.6em][r]{48.0}\,\textcolor{red!80!black}{\makebox[5.6em][r]{($\Delta -6.0$)}} & \makebox[2.6em][r]{\textbf{25.8}}\,\textcolor{red!80!black}{\makebox[5.6em][r]{($\Delta -28.2$)}} & \makebox[2.6em][r]{40.7}\,\textcolor{red!80!black}{\makebox[5.6em][r]{($\Delta -13.3$)}} & \makebox[2.6em][r]{52.2}\,\textcolor{red!80!black}{\makebox[5.6em][r]{($\Delta -1.8$)}} & \makebox[2.6em][r]{\textbf{29.2}}\,\textcolor{red!80!black}{\makebox[5.6em][r]{($\Delta -24.8$)}} \\
 & OLMo-3-7B & 41.8 & \makebox[2.6em][r]{26.0}\,\textcolor{red!80!black}{\makebox[5.6em][r]{($\Delta -15.8$)}} & \makebox[2.6em][r]{35.4}\,\textcolor{red!80!black}{\makebox[5.6em][r]{($\Delta -6.4$)}} & \makebox[2.6em][r]{\textbf{16.6}}\,\textcolor{red!80!black}{\makebox[5.6em][r]{($\Delta -25.2$)}} & \makebox[2.6em][r]{41.8}\,\textcolor{black}{\makebox[5.6em][r]{($\Delta +0.0$)}} & \makebox[2.6em][r]{\textbf{26.0}}\,\textcolor{red!80!black}{\makebox[5.6em][r]{($\Delta -15.8$)}} & \makebox[2.6em][r]{57.6}\,\textcolor{green!45!black}{\makebox[5.6em][r]{($\Delta +15.8$)}} \\
 & Qwen3-4B & 69.0 & \makebox[2.6em][r]{32.2}\,\textcolor{red!80!black}{\makebox[5.6em][r]{($\Delta -36.8$)}} & \makebox[2.6em][r]{49.0}\,\textcolor{red!80!black}{\makebox[5.6em][r]{($\Delta -20.0$)}} & \makebox[2.6em][r]{\textbf{15.4}}\,\textcolor{red!80!black}{\makebox[5.6em][r]{($\Delta -53.6$)}} & \makebox[2.6em][r]{54.7}\,\textcolor{red!80!black}{\makebox[5.6em][r]{($\Delta -14.3$)}} & \makebox[2.6em][r]{59.6}\,\textcolor{red!80!black}{\makebox[5.6em][r]{($\Delta -9.4$)}} & \makebox[2.6em][r]{\textbf{49.8}}\,\textcolor{red!80!black}{\makebox[5.6em][r]{($\Delta -19.2$)}} \\
 & Qwen3-14B & 61.2 & \makebox[2.6em][r]{47.9}\,\textcolor{red!80!black}{\makebox[5.6em][r]{($\Delta -13.3$)}} & \makebox[2.6em][r]{\textbf{27.0}}\,\textcolor{red!80!black}{\makebox[5.6em][r]{($\Delta -34.2$)}} & \makebox[2.6em][r]{68.8}\,\textcolor{green!45!black}{\makebox[5.6em][r]{($\Delta +7.6$)}} & \makebox[2.6em][r]{55.6}\,\textcolor{red!80!black}{\makebox[5.6em][r]{($\Delta -5.6$)}} & \makebox[2.6em][r]{\textbf{43.2}}\,\textcolor{red!80!black}{\makebox[5.6em][r]{($\Delta -18.0$)}} & \makebox[2.6em][r]{68.0}\,\textcolor{green!45!black}{\makebox[5.6em][r]{($\Delta +6.8$)}} \\
 & Qwen3-32B & 47.2 & \makebox[2.6em][r]{25.9}\,\textcolor{red!80!black}{\makebox[5.6em][r]{($\Delta -21.3$)}} & \makebox[2.6em][r]{28.2}\,\textcolor{red!80!black}{\makebox[5.6em][r]{($\Delta -19.0$)}} & \makebox[2.6em][r]{\textbf{23.6}}\,\textcolor{red!80!black}{\makebox[5.6em][r]{($\Delta -23.6$)}} & \makebox[2.6em][r]{66.1}\,\textcolor{green!45!black}{\makebox[5.6em][r]{($\Delta +18.9$)}} & \makebox[2.6em][r]{\textbf{48.8}}\,\textcolor{green!45!black}{\makebox[5.6em][r]{($\Delta +1.6$)}} & \makebox[2.6em][r]{83.4}\,\textcolor{green!45!black}{\makebox[5.6em][r]{($\Delta +36.2$)}} \\
 & Gemma3-4B & 34.8 & \makebox[2.6em][r]{37.6}\,\textcolor{green!45!black}{\makebox[5.6em][r]{($\Delta +2.8$)}} & \makebox[2.6em][r]{\textbf{6.0}}\,\textcolor{red!80!black}{\makebox[5.6em][r]{($\Delta -28.8$)}} & \makebox[2.6em][r]{69.2}\,\textcolor{green!45!black}{\makebox[5.6em][r]{($\Delta +34.4$)}} & \makebox[2.6em][r]{50.9}\,\textcolor{green!45!black}{\makebox[5.6em][r]{($\Delta +16.1$)}} & \makebox[2.6em][r]{\textbf{32.8}}\,\textcolor{red!80!black}{\makebox[5.6em][r]{($\Delta -2.0$)}} & \makebox[2.6em][r]{69.0}\,\textcolor{green!45!black}{\makebox[5.6em][r]{($\Delta +34.2$)}} \\
 & Gemma3-12B & 64.2 & \makebox[2.6em][r]{43.2}\,\textcolor{red!80!black}{\makebox[5.6em][r]{($\Delta -21.0$)}} & \makebox[2.6em][r]{\textbf{17.0}}\,\textcolor{red!80!black}{\makebox[5.6em][r]{($\Delta -47.2$)}} & \makebox[2.6em][r]{69.4}\,\textcolor{green!45!black}{\makebox[5.6em][r]{($\Delta +5.2$)}} & \makebox[2.6em][r]{56.5}\,\textcolor{red!80!black}{\makebox[5.6em][r]{($\Delta -7.7$)}} & \makebox[2.6em][r]{\textbf{43.0}}\,\textcolor{red!80!black}{\makebox[5.6em][r]{($\Delta -21.2$)}} & \makebox[2.6em][r]{70.0}\,\textcolor{green!45!black}{\makebox[5.6em][r]{($\Delta +5.8$)}} \\
 & Gemma3-27B & 52.6 & \makebox[2.6em][r]{50.0}\,\textcolor{red!80!black}{\makebox[5.6em][r]{($\Delta -2.6$)}} & \makebox[2.6em][r]{\textbf{26.2}}\,\textcolor{red!80!black}{\makebox[5.6em][r]{($\Delta -26.4$)}} & \makebox[2.6em][r]{73.8}\,\textcolor{green!45!black}{\makebox[5.6em][r]{($\Delta +21.2$)}} & \makebox[2.6em][r]{61.0}\,\textcolor{green!45!black}{\makebox[5.6em][r]{($\Delta +8.4$)}} & \makebox[2.6em][r]{\textbf{46.8}}\,\textcolor{red!80!black}{\makebox[5.6em][r]{($\Delta -5.8$)}} & \makebox[2.6em][r]{75.2}\,\textcolor{green!45!black}{\makebox[5.6em][r]{($\Delta +22.6$)}} \\
 & Llama-3.1-8B & 74.0 & \makebox[2.6em][r]{21.2}\,\textcolor{red!80!black}{\makebox[5.6em][r]{($\Delta -52.8$)}} & \makebox[2.6em][r]{22.0}\,\textcolor{red!80!black}{\makebox[5.6em][r]{($\Delta -52.0$)}} & \makebox[2.6em][r]{\textbf{20.4}}\,\textcolor{red!80!black}{\makebox[5.6em][r]{($\Delta -53.6$)}} & \makebox[2.6em][r]{29.2}\,\textcolor{red!80!black}{\makebox[5.6em][r]{($\Delta -44.8$)}} & \makebox[2.6em][r]{34.2}\,\textcolor{red!80!black}{\makebox[5.6em][r]{($\Delta -39.8$)}} & \makebox[2.6em][r]{\textbf{24.2}}\,\textcolor{red!80!black}{\makebox[5.6em][r]{($\Delta -49.8$)}} \\
 & Mistral-7B-v0.3 & 28.6 & \makebox[2.6em][r]{33.5}\,\textcolor{green!45!black}{\makebox[5.6em][r]{($\Delta +4.9$)}} & \makebox[2.6em][r]{\textbf{26.8}}\,\textcolor{red!80!black}{\makebox[5.6em][r]{($\Delta -1.8$)}} & \makebox[2.6em][r]{40.2}\,\textcolor{green!45!black}{\makebox[5.6em][r]{($\Delta +11.6$)}} & \makebox[2.6em][r]{41.9}\,\textcolor{green!45!black}{\makebox[5.6em][r]{($\Delta +13.3$)}} & \makebox[2.6em][r]{\textbf{25.8}}\,\textcolor{red!80!black}{\makebox[5.6em][r]{($\Delta -2.8$)}} & \makebox[2.6em][r]{58.0}\,\textcolor{green!45!black}{\makebox[5.6em][r]{($\Delta +29.4$)}} \\
\cline{2-9}
 & \textbf{Model mean} & 52.7 & \makebox[2.6em][r]{35.4}\,\textcolor{red!80!black}{\makebox[5.6em][r]{($\Delta -17.3$)}} & \makebox[2.6em][r]{\textbf{28.6}}\,\textcolor{red!80!black}{\makebox[5.6em][r]{($\Delta -24.2$)}} & \makebox[2.6em][r]{42.3}\,\textcolor{red!80!black}{\makebox[5.6em][r]{($\Delta -10.4$)}} & \makebox[2.6em][r]{49.8}\,\textcolor{red!80!black}{\makebox[5.6em][r]{($\Delta -2.9$)}} & \makebox[2.6em][r]{\textbf{41.2}}\,\textcolor{red!80!black}{\makebox[5.6em][r]{($\Delta -11.5$)}} & \makebox[2.6em][r]{58.4}\,\textcolor{green!45!black}{\makebox[5.6em][r]{($\Delta +5.7$)}} \\
\bottomrule
\end{tabular}}
\captionsetup{justification=raggedright,singlelinecheck=false}
\caption{Initial and post-retraction accuracy (\%) by model
and dataset.
Direct and CoT use the same initial Direct response.
$\Delta$ denotes change from initial accuracy (pp);
bold marks the lower accuracy between
\textsc{rt}$\rightarrow$\textsc{rt} and
\textsc{rt}$\rightarrow$\textsc{ot} within each format.
Model means weight all ten LLMs equally.}
% \caption{Per-model initial Direct answer and follow-up Direct/CoT accuracy (\%). Both modes share the same stored initial Direct response. Model mean weights completed models equally within each dataset. RT and OT denote ground-truth answer labels before and after retraction; arrows denote label transitions, not model outputs. Bold marks the lower RT $\to$ RT / RT $\to$ OT accuracy within each response mode. $\Delta$ is follow-up minus initial accuracy on the same condition, in percentage points; green/red text indicates gains/losses. CoT returns a short reason and an answer in JSON. Scoring: answer\_normalization\_v2. Configuration: run1\_base\_retraction/quoted/test/seed=0/revision=main/max\_tokens=1024; paper\_retraction\_prompts\_v2.}
\label{tab:appendix-run1-1}
\end{table*}

\begin{table*}[p]
\centering\small
\setlength{\tabcolsep}{1.5pt}
\renewcommand{\arraystretch}{1.534}
\resizebox{0.96\textwidth}{!}{%
\begin{tabular}{@{}lcc|c|ccccccccc@{}}
\toprule
Dataset & $E$ & $A_1$ & \shortstack{Model\\mean} & \shortstack{OLMo-2\\7B} & \shortstack{OLMo-3\\7B} & \shortstack{Qwen3\\4B} & \shortstack{Qwen3\\14B} & \shortstack{Qwen3\\32B} & \shortstack{Gemma3\\4B} & \shortstack{Gemma3\\12B} & \shortstack{Gemma3\\27B} & \shortstack{Llama\\3.1-8B} \\
\midrule
\rowcolor{gray!15}\multicolumn{13}{c}{\textbf{RT $\to$ RT}} \\
\rowcolor{gray!7}
\textbf{Priority} & \textcolor{green!40!black}{$\checkmark$} & \textcolor{green!40!black}{$\checkmark$} & \textbf{51.3} & \textbf{0.4} & \textbf{0.2} & \textbf{29.2} & \textbf{17.1} & \textbf{97.8} & \textbf{61.7} & \textbf{75.2} & \textbf{89.3} & \textbf{91.1} \\
 & \textcolor{green!40!black}{$\checkmark$} & \textcolor{red!75!black}{$\times$} & \cellcolor{green!6.8}52.3 ($\Delta +0.9$) & \cellcolor{green!6.0}0.9 ($\Delta +0.5$) & \cellcolor{green!5.2}0.3 ($\Delta +0.1$) & \cellcolor{green!45.0}51.4 ($\Delta +22.2$) & \cellcolor{red!12.4}13.4 ($\Delta -3.7$) & \cellcolor{green!7.6}99.1 ($\Delta +1.3$) & \cellcolor{red!40.8}43.8 ($\Delta -17.9$) & \cellcolor{green!10.8}78.1 ($\Delta +2.9$) & \cellcolor{green!10.2}91.9 ($\Delta +2.6$) & \cellcolor{green!5.6}91.4 ($\Delta +0.3$) \\
 & \textcolor{red!75!black}{$\times$} & \textcolor{green!40!black}{$\checkmark$} & \cellcolor{red!29.6}39.0 ($\Delta -12.3$) & \cellcolor{green!6.0}0.9 ($\Delta +0.5$) & \cellcolor{red!5.4}0.0 ($\Delta -0.2$) & \cellcolor{red!45.0}3.5 ($\Delta -25.7$) & \cellcolor{red!17.4}10.9 ($\Delta -6.2$) & \cellcolor{red!18.4}91.1 ($\Delta -6.7$) & \cellcolor{red!45.0}18.4 ($\Delta -43.3$) & \cellcolor{red!30.0}62.7 ($\Delta -12.5$) & \cellcolor{red!30.6}76.5 ($\Delta -12.8$) & \cellcolor{red!12.4}87.4 ($\Delta -3.7$) \\
 & \textcolor{red!75!black}{$\times$} & \textcolor{red!75!black}{$\times$} & \cellcolor{red!21.4}43.1 ($\Delta -8.2$) & 0.4 ($\Delta +0.0$) & \cellcolor{red!5.4}0.0 ($\Delta -0.2$) & \cellcolor{green!14.2}33.8 ($\Delta +4.6$) & \cellcolor{red!30.6}4.3 ($\Delta -12.8$) & \cellcolor{red!11.8}94.4 ($\Delta -3.4$) & \cellcolor{red!45.0}27.4 ($\Delta -34.3$) & \cellcolor{red!24.8}65.3 ($\Delta -9.9$) & \cellcolor{red!19.2}82.2 ($\Delta -7.1$) & \cellcolor{red!26.2}80.5 ($\Delta -10.6$) \\
\specialrule{0.6pt}{2pt}{2pt}
\rowcolor{gray!7}
\textbf{FOLIO} & \textcolor{green!40!black}{$\checkmark$} & \textcolor{green!40!black}{$\checkmark$} & \textbf{30.8} & \textbf{60.2} & \textbf{47.0} & \textbf{35.6} & \textbf{31.6} & \textbf{40.8} & \textbf{7.2} & \textbf{12.2} & \textbf{26.4} & \textbf{16.4} \\
 & \textcolor{green!40!black}{$\checkmark$} & \textcolor{red!75!black}{$\times$} & 30.8 ($\Delta +0.0$) & \cellcolor{green!27.8}71.6 ($\Delta +11.4$) & \cellcolor{green!13.8}51.4 ($\Delta +4.4$) & \cellcolor{red!13.8}31.2 ($\Delta -4.4$) & \cellcolor{green!6.6}32.4 ($\Delta +0.8$) & \cellcolor{red!13.8}36.4 ($\Delta -4.4$) & \cellcolor{red!8.2}5.6 ($\Delta -1.6$) & \cellcolor{red!10.6}9.4 ($\Delta -2.8$) & \cellcolor{red!9.0}24.4 ($\Delta -2.0$) & \cellcolor{red!8.2}14.8 ($\Delta -1.6$) \\
 & \textcolor{red!75!black}{$\times$} & \textcolor{green!40!black}{$\checkmark$} & \cellcolor{green!5.4}31.0 ($\Delta +0.2$) & \cellcolor{green!17.4}66.4 ($\Delta +6.2$) & \cellcolor{green!5.8}47.4 ($\Delta +0.4$) & \cellcolor{red!5.4}35.4 ($\Delta -0.2$) & \cellcolor{red!9.4}29.4 ($\Delta -2.2$) & \cellcolor{red!7.4}39.6 ($\Delta -1.2$) & \cellcolor{red!7.4}6.0 ($\Delta -1.2$) & 12.2 ($\Delta +0.0$) & \cellcolor{red!7.4}25.2 ($\Delta -1.2$) & \cellcolor{green!7.0}17.4 ($\Delta +1.0$) \\
 & \textcolor{red!75!black}{$\times$} & \textcolor{red!75!black}{$\times$} & \cellcolor{red!5.4}30.6 ($\Delta -0.2$) & \cellcolor{green!32.6}74.0 ($\Delta +13.8$) & \cellcolor{green!10.2}49.6 ($\Delta +2.6$) & \cellcolor{red!14.2}31.0 ($\Delta -4.6$) & \cellcolor{red!5.4}31.4 ($\Delta -0.2$) & \cellcolor{red!13.0}36.8 ($\Delta -4.0$) & \cellcolor{red!9.4}5.0 ($\Delta -2.2$) & \cellcolor{red!9.4}10.0 ($\Delta -2.2$) & \cellcolor{red!12.2}22.8 ($\Delta -3.6$) & \cellcolor{red!7.8}15.0 ($\Delta -1.4$) \\
\specialrule{0.6pt}{2pt}{2pt}
\rowcolor{gray!7}
\textbf{\shortstack[l]{Proof\\Writer}} & \textcolor{green!40!black}{$\checkmark$} & \textcolor{green!40!black}{$\checkmark$} & \textbf{28.8} & \textbf{48.0} & \textbf{35.4} & \textbf{49.0} & \textbf{27.0} & \textbf{28.2} & \textbf{6.0} & \textbf{17.0} & \textbf{26.2} & \textbf{22.0} \\
 & \textcolor{green!40!black}{$\checkmark$} & \textcolor{red!75!black}{$\times$} & \cellcolor{red!13.5}24.5 ($\Delta -4.3$) & \cellcolor{red!39.0}31.0 ($\Delta -17.0$) & \cellcolor{green!11.8}38.8 ($\Delta +3.4$) & \cellcolor{red!23.8}39.6 ($\Delta -9.4$) & \cellcolor{red!7.4}25.8 ($\Delta -1.2$) & \cellcolor{red!41.8}9.8 ($\Delta -18.4$) & \cellcolor{green!9.0}8.0 ($\Delta +2.0$) & 17.0 ($\Delta +0.0$) & \cellcolor{red!10.6}23.4 ($\Delta -2.8$) & \cellcolor{green!15.0}27.0 ($\Delta +5.0$) \\
 & \textcolor{red!75!black}{$\times$} & \textcolor{green!40!black}{$\checkmark$} & \cellcolor{red!8.7}26.9 ($\Delta -1.9$) & \cellcolor{red!6.6}47.2 ($\Delta -0.8$) & \cellcolor{red!5.8}35.0 ($\Delta -0.4$) & \cellcolor{red!31.4}35.8 ($\Delta -13.2$) & \cellcolor{red!8.2}25.4 ($\Delta -1.6$) & \cellcolor{green!7.8}29.6 ($\Delta +1.4$) & 6.0 ($\Delta +0.0$) & \cellcolor{green!5.4}17.2 ($\Delta +0.2$) & \cellcolor{red!8.2}24.6 ($\Delta -1.6$) & \cellcolor{red!6.6}21.2 ($\Delta -0.8$) \\
 & \textcolor{red!75!black}{$\times$} & \textcolor{red!75!black}{$\times$} & \cellcolor{red!10.1}26.2 ($\Delta -2.5$) & \cellcolor{red!25.4}37.8 ($\Delta -10.2$) & \cellcolor{green!12.6}39.2 ($\Delta +3.8$) & \cellcolor{red!24.2}39.4 ($\Delta -9.6$) & \cellcolor{red!8.6}25.2 ($\Delta -1.8$) & \cellcolor{red!18.6}21.4 ($\Delta -6.8$) & \cellcolor{green!9.4}8.2 ($\Delta +2.2$) & \cellcolor{red!5.8}16.6 ($\Delta -0.4$) & \cellcolor{red!12.2}22.6 ($\Delta -3.6$) & \cellcolor{green!12.2}25.6 ($\Delta +3.6$) \\
\specialrule{0.6pt}{2pt}{2pt}
\rowcolor{gray!7}
\textbf{bAbI} & \textcolor{green!40!black}{$\checkmark$} & \textcolor{green!40!black}{$\checkmark$} & \textbf{69.4} & \textbf{52.0} & \textbf{64.0} & \textbf{73.8} & \textbf{88.8} & \textbf{80.2} & \textbf{54.8} & \textbf{93.8} & \textbf{88.0} & \textbf{29.4} \\
 & \textcolor{green!40!black}{$\checkmark$} & \textcolor{red!75!black}{$\times$} & \cellcolor{red!23.0}60.4 ($\Delta -9.0$) & \cellcolor{green!8.2}53.6 ($\Delta +1.6$) & \cellcolor{red!38.6}47.2 ($\Delta -16.8$) & \cellcolor{red!23.8}64.4 ($\Delta -9.4$) & \cellcolor{red!45.0}67.8 ($\Delta -21.0$) & \cellcolor{red!9.4}78.0 ($\Delta -2.2$) & \cellcolor{red!45.0}31.6 ($\Delta -23.2$) & \cellcolor{green!6.2}94.4 ($\Delta +0.6$) & \cellcolor{green!5.4}88.2 ($\Delta +0.2$) & \cellcolor{red!26.6}18.6 ($\Delta -10.8$) \\
 & \textcolor{red!75!black}{$\times$} & \textcolor{green!40!black}{$\checkmark$} & \cellcolor{red!19.8}62.0 ($\Delta -7.4$) & \cellcolor{red!45.0}27.8 ($\Delta -24.2$) & \cellcolor{red!22.6}55.2 ($\Delta -8.8$) & \cellcolor{red!8.2}72.2 ($\Delta -1.6$) & \cellcolor{red!20.6}81.0 ($\Delta -7.8$) & \cellcolor{red!8.6}78.4 ($\Delta -1.8$) & \cellcolor{red!45.0}33.4 ($\Delta -21.4$) & \cellcolor{green!8.2}95.4 ($\Delta +1.6$) & \cellcolor{red!6.6}87.2 ($\Delta -0.8$) & \cellcolor{red!9.0}27.4 ($\Delta -2.0$) \\
 & \textcolor{red!75!black}{$\times$} & \textcolor{red!75!black}{$\times$} & \cellcolor{red!45.0}30.1 ($\Delta -39.3$) & \cellcolor{red!45.0}13.6 ($\Delta -38.4$) & \cellcolor{red!45.0}1.4 ($\Delta -62.6$) & \cellcolor{red!45.0}20.0 ($\Delta -53.8$) & \cellcolor{red!45.0}20.2 ($\Delta -68.6$) & \cellcolor{red!45.0}53.0 ($\Delta -27.2$) & \cellcolor{red!45.0}1.6 ($\Delta -53.2$) & \cellcolor{red!45.0}70.8 ($\Delta -23.0$) & \cellcolor{red!40.2}70.4 ($\Delta -17.6$) & \cellcolor{red!23.8}20.0 ($\Delta -9.4$) \\
\specialrule{1pt}{3pt}{3pt}
\rowcolor{gray!15}\multicolumn{13}{c}{\textbf{RT $\to$ OT}} \\
\rowcolor{gray!7}
\textbf{Priority} & \textcolor{green!40!black}{$\checkmark$} & \textcolor{green!40!black}{$\checkmark$} & \textbf{93.2} & \textbf{87.5} & \textbf{98.9} & \textbf{100.0} & \textbf{100.0} & \textbf{100.0} & \textbf{55.8} & \textbf{99.9} & \textbf{99.7} & \textbf{96.8} \\
 & \textcolor{green!40!black}{$\checkmark$} & \textcolor{red!75!black}{$\times$} & \cellcolor{red!7.3}92.0 ($\Delta -1.1$) & \cellcolor{red!37.8}71.1 ($\Delta -16.4$) & \cellcolor{red!6.6}98.1 ($\Delta -0.8$) & 100.0 ($\Delta +0.0$) & 100.0 ($\Delta +0.0$) & 100.0 ($\Delta +0.0$) & \cellcolor{green!20.4}63.5 ($\Delta +7.7$) & \cellcolor{red!5.2}99.8 ($\Delta -0.1$) & \cellcolor{red!5.2}99.6 ($\Delta -0.1$) & \cellcolor{red!6.0}96.3 ($\Delta -0.5$) \\
 & \textcolor{red!75!black}{$\times$} & \textcolor{green!40!black}{$\checkmark$} & \cellcolor{red!5.7}92.8 ($\Delta -0.3$) & \cellcolor{red!20.6}79.7 ($\Delta -7.8$) & \cellcolor{red!17.0}92.9 ($\Delta -6.0$) & 100.0 ($\Delta +0.0$) & 100.0 ($\Delta +0.0$) & 100.0 ($\Delta +0.0$) & \cellcolor{green!39.4}73.0 ($\Delta +17.2$) & \cellcolor{green!5.2}100.0 ($\Delta +0.1$) & \cellcolor{red!16.8}93.8 ($\Delta -5.9$) & \cellcolor{red!6.4}96.1 ($\Delta -0.7$) \\
 & \textcolor{red!75!black}{$\times$} & \textcolor{red!75!black}{$\times$} & \cellcolor{red!10.0}90.7 ($\Delta -2.5$) & \cellcolor{red!45.0}65.0 ($\Delta -22.5$) & \cellcolor{red!20.4}91.2 ($\Delta -7.7$) & \cellcolor{red!5.2}99.9 ($\Delta -0.1$) & 100.0 ($\Delta +0.0$) & 100.0 ($\Delta +0.0$) & \cellcolor{green!30.4}68.5 ($\Delta +12.7$) & 99.9 ($\Delta +0.0$) & \cellcolor{red!17.8}93.3 ($\Delta -6.4$) & \cellcolor{green!7.8}98.2 ($\Delta +1.4$) \\
\specialrule{0.6pt}{2pt}{2pt}
\rowcolor{gray!7}
\textbf{FOLIO} & \textcolor{green!40!black}{$\checkmark$} & \textcolor{green!40!black}{$\checkmark$} & \textbf{82.6} & \textbf{21.2} & \textbf{61.6} & \textbf{91.0} & \textbf{99.2} & \textbf{95.4} & \textbf{89.8} & \textbf{97.0} & \textbf{97.2} & \textbf{90.6} \\
 & \textcolor{green!40!black}{$\checkmark$} & \textcolor{red!75!black}{$\times$} & \cellcolor{red!7.8}81.1 ($\Delta -1.4$) & \cellcolor{red!34.2}6.6 ($\Delta -14.6$) & \cellcolor{red!26.2}51.0 ($\Delta -10.6$) & \cellcolor{green!13.4}95.2 ($\Delta +4.2$) & 99.2 ($\Delta +0.0$) & \cellcolor{red!5.4}95.2 ($\Delta -0.2$) & \cellcolor{green!12.2}93.4 ($\Delta +3.6$) & \cellcolor{green!10.6}99.8 ($\Delta +2.8$) & \cellcolor{green!7.8}98.6 ($\Delta +1.4$) & \cellcolor{green!6.2}91.2 ($\Delta +0.6$) \\
 & \textcolor{red!75!black}{$\times$} & \textcolor{green!40!black}{$\checkmark$} & 82.6 ($\Delta +0.0$) & \cellcolor{red!15.0}16.2 ($\Delta -5.0$) & \cellcolor{green!12.6}65.4 ($\Delta +3.8$) & \cellcolor{red!5.4}90.8 ($\Delta -0.2$) & \cellcolor{green!5.8}99.6 ($\Delta +0.4$) & \cellcolor{red!5.4}95.2 ($\Delta -0.2$) & \cellcolor{green!6.2}90.4 ($\Delta +0.6$) & \cellcolor{green!6.2}97.6 ($\Delta +0.6$) & 97.2 ($\Delta +0.0$) & \cellcolor{green!5.8}91.0 ($\Delta +0.4$) \\
 & \textcolor{red!75!black}{$\times$} & \textcolor{red!75!black}{$\times$} & \cellcolor{red!7.2}81.4 ($\Delta -1.1$) & \cellcolor{red!35.4}6.0 ($\Delta -15.2$) & \cellcolor{red!22.2}53.0 ($\Delta -8.6$) & \cellcolor{green!13.4}95.2 ($\Delta +4.2$) & \cellcolor{green!5.4}99.4 ($\Delta +0.2$) & \cellcolor{green!7.4}96.6 ($\Delta +1.2$) & \cellcolor{green!11.8}93.2 ($\Delta +3.4$) & \cellcolor{green!10.6}99.8 ($\Delta +2.8$) & \cellcolor{green!7.8}98.6 ($\Delta +1.4$) & \cellcolor{green!6.2}91.2 ($\Delta +0.6$) \\
\specialrule{0.6pt}{2pt}{2pt}
\rowcolor{gray!7}
\textbf{\shortstack[l]{Proof\\Writer}} & \textcolor{green!40!black}{$\checkmark$} & \textcolor{green!40!black}{$\checkmark$} & \textbf{42.6} & \textbf{25.8} & \textbf{16.6} & \textbf{15.4} & \textbf{68.8} & \textbf{23.6} & \textbf{69.2} & \textbf{69.4} & \textbf{73.8} & \textbf{20.4} \\
 & \textcolor{green!40!black}{$\checkmark$} & \textcolor{red!75!black}{$\times$} & \cellcolor{red!7.2}41.4 ($\Delta -1.1$) & \cellcolor{green!22.6}34.6 ($\Delta +8.8$) & \cellcolor{red!25.0}6.6 ($\Delta -10.0$) & \cellcolor{green!22.6}24.2 ($\Delta +8.8$) & \cellcolor{green!11.4}72.0 ($\Delta +3.2$) & \cellcolor{red!31.4}10.4 ($\Delta -13.2$) & 69.2 ($\Delta +0.0$) & \cellcolor{red!6.2}68.8 ($\Delta -0.6$) & \cellcolor{green!10.6}76.6 ($\Delta +2.8$) & \cellcolor{red!24.6}10.6 ($\Delta -9.8$) \\
 & \textcolor{red!75!black}{$\times$} & \textcolor{green!40!black}{$\checkmark$} & \cellcolor{green!8.6}44.4 ($\Delta +1.8$) & \cellcolor{red!17.8}19.4 ($\Delta -6.4$) & \cellcolor{green!9.0}18.6 ($\Delta +2.0$) & \cellcolor{green!22.2}24.0 ($\Delta +8.6$) & \cellcolor{green!15.4}74.0 ($\Delta +5.2$) & \cellcolor{green!13.0}27.6 ($\Delta +4.0$) & \cellcolor{green!5.8}69.6 ($\Delta +0.4$) & \cellcolor{green!7.8}70.8 ($\Delta +1.4$) & \cellcolor{green!6.6}74.6 ($\Delta +0.8$) & \cellcolor{green!5.8}20.8 ($\Delta +0.4$) \\
 & \textcolor{red!75!black}{$\times$} & \textcolor{red!75!black}{$\times$} & 42.6 ($\Delta +0.0$) & \cellcolor{red!7.4}24.6 ($\Delta -1.2$) & \cellcolor{red!23.0}7.6 ($\Delta -9.0$) & \cellcolor{green!23.0}24.4 ($\Delta +9.0$) & \cellcolor{green!13.0}72.8 ($\Delta +4.0$) & \cellcolor{red!11.4}20.4 ($\Delta -3.2$) & \cellcolor{green!9.4}71.4 ($\Delta +2.2$) & \cellcolor{green!7.8}70.8 ($\Delta +1.4$) & \cellcolor{green!12.2}77.4 ($\Delta +3.6$) & \cellcolor{red!18.6}13.6 ($\Delta -6.8$) \\
\specialrule{0.6pt}{2pt}{2pt}
\rowcolor{gray!7}
\textbf{bAbI} & \textcolor{green!40!black}{$\checkmark$} & \textcolor{green!40!black}{$\checkmark$} & \textbf{60.4} & \textbf{53.2} & \textbf{32.6} & \textbf{51.4} & \textbf{56.2} & \textbf{85.8} & \textbf{27.6} & \textbf{97.0} & \textbf{97.4} & \textbf{42.8} \\
 & \textcolor{green!40!black}{$\checkmark$} & \textcolor{red!75!black}{$\times$} & \cellcolor{red!37.7}44.1 ($\Delta -16.4$) & \cellcolor{red!11.4}50.0 ($\Delta -3.2$) & \cellcolor{red!37.0}16.6 ($\Delta -16.0$) & \cellcolor{red!45.0}29.4 ($\Delta -22.0$) & \cellcolor{red!45.0}17.2 ($\Delta -39.0$) & \cellcolor{red!15.8}80.4 ($\Delta -5.4$) & \cellcolor{red!33.8}13.2 ($\Delta -14.4$) & \cellcolor{red!19.8}89.6 ($\Delta -7.4$) & \cellcolor{red!27.8}86.0 ($\Delta -11.4$) & \cellcolor{red!45.0}14.4 ($\Delta -28.4$) \\
 & \textcolor{red!75!black}{$\times$} & \textcolor{green!40!black}{$\checkmark$} & \cellcolor{red!45.0}35.3 ($\Delta -25.1$) & \cellcolor{red!45.0}21.6 ($\Delta -31.6$) & \cellcolor{red!45.0}10.0 ($\Delta -22.6$) & \cellcolor{red!45.0}20.2 ($\Delta -31.2$) & \cellcolor{red!45.0}5.6 ($\Delta -50.6$) & \cellcolor{red!31.0}72.8 ($\Delta -13.0$) & \cellcolor{red!25.8}17.2 ($\Delta -10.4$) & \cellcolor{red!45.0}63.4 ($\Delta -33.6$) & \cellcolor{red!23.8}88.0 ($\Delta -9.4$) & \cellcolor{red!45.0}19.0 ($\Delta -23.8$) \\
 & \textcolor{red!75!black}{$\times$} & \textcolor{red!75!black}{$\times$} & \cellcolor{red!45.0}23.0 ($\Delta -37.4$) & \cellcolor{red!45.0}14.6 ($\Delta -38.6$) & \cellcolor{red!45.0}0.2 ($\Delta -32.4$) & \cellcolor{red!45.0}0.2 ($\Delta -51.2$) & \cellcolor{red!45.0}1.6 ($\Delta -54.6$) & \cellcolor{red!44.6}66.0 ($\Delta -19.8$) & \cellcolor{red!45.0}0.2 ($\Delta -27.4$) & \cellcolor{red!45.0}39.2 ($\Delta -57.8$) & \cellcolor{red!45.0}66.8 ($\Delta -30.6$) & \cellcolor{red!45.0}18.4 ($\Delta -24.4$) \\
\bottomrule
\end{tabular}}
\caption{Direct accuracy (\%) under final-position attention
blocking, by model and dataset.
$\checkmark$/$\times$ indicates allowed/blocked access to
evidence ($E$) and the prior answer ($A_1$).
Bold entries are full-access baselines; blocked settings report
accuracy with changes from baseline in parentheses (pp).}
\label{tab:appendix-run3-datasets-by-model}
\end{table*}

\paragraph{Priority Dataset}

\begin{table*}[!t]
% Colors
\definecolor{apdxRT}{HTML}{DE7464}
\definecolor{apdxRTline}{HTML}{EBB0A8}
\definecolor{apdxRTbg}{HTML}{FDF6F5}
\definecolor{apdxRThead}{HTML}{FBEAE7}
\definecolor{apdxRTink}{HTML}{A8392C}
\definecolor{apdxOT}{HTML}{4E8CBF}
\definecolor{apdxOTline}{HTML}{AAC8E1}
\definecolor{apdxOTbg}{HTML}{F5F9FC}
\definecolor{apdxOThead}{HTML}{E7F0F8}
\definecolor{apdxOTink}{HTML}{2C6699}
\definecolor{apdxBothink}{HTML}{4E2A78}
\definecolor{apdxFrame}{HTML}{C9CDD2}
\definecolor{apdxTitle}{HTML}{F1F2F4}
\definecolor{apdxGrid}{HTML}{DADDE1}
\definecolor{apdxOK}{HTML}{2E7D32}
\definecolor{apdxNo}{HTML}{B71C1C}
\definecolor{apdxHL}{HTML}{FFF6C7}

\centering
\small

% Base case
\begin{tcolorbox}[
    colback=white,colframe=apdxFrame,colbacktitle=apdxTitle,coltitle=black,
    fonttitle=\small\bfseries,
    title={Priority \textperiodcentered{} same base case {\normalfont\ttfamily 3b98c7d8}},
    boxrule=0.6pt,arc=2pt,left=6pt,right=6pt,top=4pt,bottom=5pt,fontupper=\small
]
    \textbf{Priority:} S has priority.\par
    \textbf{Retraction:} Withdraw the statement ``\textcolor{apdxBothink}{S has priority}''.\par
    \textbf{Initial gold:} S (all models answered correctly)
    \par\medskip

    \begin{minipage}[t]{0.493\linewidth}
        \vspace{0pt}
        \begin{tcolorbox}[
            colback=apdxRTbg,colframe=apdxRTline,colbacktitle=apdxRT,coltitle=white,
            fonttitle=\small\bfseries,title={RT\,$\to$\,RT \textperiodcentered{} label preserved},
            boxrule=0.6pt,arc=2pt,left=5pt,right=5pt,top=3pt,bottom=3pt,
            toptitle=2pt,bottomtitle=2pt,fontupper=\small
        ]
            \textbf{Scores:} U: 701, X: 774, S: 837\par
            \textbf{Gold after retraction:} S
        \end{tcolorbox}
    \end{minipage}\hfill%
    \begin{minipage}[t]{0.493\linewidth}
        \vspace{0pt}
        \begin{tcolorbox}[
            colback=apdxOTbg,colframe=apdxOTline,colbacktitle=apdxOT,coltitle=white,
            fonttitle=\small\bfseries,title={RT\,$\to$\,OT \textperiodcentered{} label changed},
            boxrule=0.6pt,arc=2pt,left=5pt,right=5pt,top=3pt,bottom=3pt,
            toptitle=2pt,bottomtitle=2pt,fontupper=\small
        ]
            \textbf{Scores:} U: 701, X: 837, S: 774\par
            \textbf{Gold after retraction:} X
        \end{tcolorbox}
    \end{minipage}
\end{tcolorbox}

\vspace{5pt}
\renewcommand{\arraystretch}{1.15}
\setlength{\tabcolsep}{5pt}\setlength{\extrarowheight}{2pt}
\setlength{\aboverulesep}{0pt}\setlength{\belowrulesep}{0pt}
\hyphenpenalty=10000\exhyphenpenalty=10000

\begin{tabularx}{\textwidth}{
    >{\raggedright\arraybackslash}p{2.3cm}!{\color{apdxFrame}\vrule width 0.6pt}
    >{\raggedright\arraybackslash}p{1.30cm}!{\color{apdxFrame}\vrule width 0.6pt}
    >{\raggedright\arraybackslash}X!{\color{apdxFrame}\vrule width 0.6pt}
    >{\raggedright\arraybackslash}p{1.30cm}!{\color{apdxFrame}\vrule width 0.6pt}
    >{\raggedright\arraybackslash}X
}
\arrayrulecolor{black}\toprule
\multirow{2}{*}{\textbf{Model}}
    & \multicolumn{2}{c!{\color{apdxFrame}\vrule width 0.6pt}}{\cellcolor{apdxRThead}\textcolor{apdxRTink}{\textbf{RT\,$\to$\,RT preserved}}}
    & \multicolumn{2}{c}{\cellcolor{apdxOThead}\textcolor{apdxOTink}{\textbf{RT\,$\to$\,OT changed}}} \\
    & \cellcolor{apdxRTbg}\textbf{Direct}
    & \cellcolor{apdxRTbg}\textbf{CoT} {\footnotesize (answer, reason)}
    & \cellcolor{apdxOTbg}\textbf{Direct}
    & \cellcolor{apdxOTbg}\textbf{CoT} {\footnotesize (answer, reason)} \\
\arrayrulecolor{black}\midrule

\textbf{Qwen3-14B}
    & \cellcolor{apdxHL}\textcolor{apdxNo}{\ding{55}} \textbf{X}
    & \textcolor{apdxOK}{\ding{51}} \textbf{S}
      {\footnotesize S no longer has priority after the withdrawal. The candidate with the highest score is S with 837. Therefore, S is chosen based on the highest score.}
    & \textcolor{apdxOK}{\ding{51}} \textbf{X}
    & \textcolor{apdxOK}{\ding{51}} \textbf{X}
      {\footnotesize S no longer has priority after the withdrawal. The candidate with the highest score is X with 837. Therefore, X should be chosen.} \\
\arrayrulecolor{apdxGrid}\hline

\textbf{Qwen3-32B}
    & \cellcolor{apdxHL}\textcolor{apdxNo}{\ding{55}} \textbf{X}
    & \textcolor{apdxOK}{\ding{51}} \textbf{S}
      {\footnotesize The priority statement for S has been withdrawn. Now, the candidate with the highest score should be chosen. S still has the highest score at 837.}
    & \textcolor{apdxOK}{\ding{51}} \textbf{X}
    & \textcolor{apdxOK}{\ding{51}} \textbf{X}
      {\footnotesize The priority statement for S has been withdrawn. Now, the candidate with the highest score should be chosen. X has the highest score at 837.} \\
\arrayrulecolor{apdxGrid}\hline

\textbf{Gemma3-12B}
    & \cellcolor{apdxHL}\textcolor{apdxNo}{\ding{55}} \textbf{X}
    & \cellcolor{apdxHL}\textcolor{apdxNo}{\ding{55}} \textbf{X}
      {\footnotesize The priority statement for S has been withdrawn. Now, we must choose the candidate with the highest score. X has the highest score at 774.}
    & \textcolor{apdxOK}{\ding{51}} \textbf{X}
    & \textcolor{apdxOK}{\ding{51}} \textbf{X}
      {\footnotesize The priority statement for S has been withdrawn. Now, we must choose the candidate with the highest score. X has the highest score at 837.} \\
\arrayrulecolor{apdxGrid}\hline

\textbf{Gemma3-27B}
    & \cellcolor{apdxHL}\textcolor{apdxNo}{\ding{55}} \textbf{X}
    & \cellcolor{apdxHL}\textcolor{apdxNo}{\ding{55}} \textbf{X}
      {\footnotesize S no longer has priority. Therefore, we choose the candidate with the highest score. X has the highest score at 774.}
    & \textcolor{apdxOK}{\ding{51}} \textbf{X}
    & \textcolor{apdxOK}{\ding{51}} \textbf{X}
      {\footnotesize S no longer has priority. Therefore, we choose the candidate with the highest score. X has the highest score at 837.} \\
\arrayrulecolor{apdxGrid}\hline

\textbf{Llama-3.1-8B}
    & \cellcolor{apdxHL}\textcolor{apdxNo}{\ding{55}} \textbf{X}
    & \cellcolor{apdxHL}\textcolor{apdxNo}{\ding{55}} \textbf{X}
      {\footnotesize No priority assigned, so the candidate with the highest score is chosen.}
    & \textcolor{apdxOK}{\ding{51}} \textbf{X}
    & \textcolor{apdxOK}{\ding{51}} \textbf{X}
      {\footnotesize No priority assigned, so the candidate with the highest score is chosen.} \\
\arrayrulecolor{black}\bottomrule
\end{tabularx}

\caption{Priority example. Only S's priority is withdrawn. The two conditions differ
only in the score assignment, so S remains the gold answer in RT$\to$RT.}
\label{tab:apdx-run1-example-priority}
\end{table*}

The Priority example illustrates a distinction between withdrawing a candidate's 
priority and excluding the candidate itself. In RT$\to$RT, the incorrect CoT
responses select X despite S retaining the highest score.
The following patterns are observed:
\begin{itemize}
    \item Withdrawal as removal (Wrong Answers): The fallback rule is stated correctly, but
    the withdrawn candidate is dropped from the comparison. Gemma3-12B and Gemma3-27B select
    X as having ``\texttt{the highest score at 774}'', although S scores 837.
    \item Rule without recomputation (Wrong Answers): The fallback rule is recited, but no scores are compared. Llama-3.1-8B gives X in both conditions, as all models do under
    Direct generation.
    \item Explicit recomputation (Correct Answers): The answer is preserved only when the
    withdrawn candidate is kept in the comparison, as in Qwen3-14B and Qwen3-32B (``\texttt{S still has the highest score at 837}'').
\end{itemize}
In this case, selecting X yields a correct answer in RT$\to$OT but an incorrect
answer in RT$\to$RT, showing why the changed-answer condition alone cannot
distinguish successful recomputation from moving away from S.

% =============================================================================

\paragraph{ProofWriter Dataset}

\begin{table*}[!t]
% Colors
\definecolor{apdxRT}{HTML}{DE7464}
\definecolor{apdxRTline}{HTML}{EBB0A8}
\definecolor{apdxRTbg}{HTML}{FDF6F5}
\definecolor{apdxRThead}{HTML}{FBEAE7}
\definecolor{apdxRTink}{HTML}{A8392C}
\definecolor{apdxOT}{HTML}{4E8CBF}
\definecolor{apdxOTline}{HTML}{AAC8E1}
\definecolor{apdxOTbg}{HTML}{F5F9FC}
\definecolor{apdxOThead}{HTML}{E7F0F8}
\definecolor{apdxOTink}{HTML}{2C6699}
\definecolor{apdxBothink}{HTML}{4E2A78}
\definecolor{apdxFrame}{HTML}{C9CDD2}
\definecolor{apdxTitle}{HTML}{F1F2F4}
\definecolor{apdxGrid}{HTML}{DADDE1}
\definecolor{apdxOK}{HTML}{2E7D32}
\definecolor{apdxNo}{HTML}{B71C1C}
\definecolor{apdxHL}{HTML}{FFF6C7}

\centering
\small

% Base case
\begin{tcolorbox}[
    colback=white,colframe=apdxFrame,colbacktitle=apdxTitle,coltitle=black,
    fonttitle=\small\bfseries,
    title={ProofWriter \textperiodcentered{} same base case {\normalfont\ttfamily 6a4b4869}},
    boxrule=0.6pt,arc=2pt,left=6pt,right=6pt,top=4pt,bottom=5pt,fontupper=\small
]
    \textbf{Facts:}\\
    triple1: The bald eagle is big.\\
    \textcolor{apdxBothink}{triple2: The bald eagle is blue.}\\
    triple3: The bald eagle is green.\\
    \textbf{Rules:}\\
    rule1: If someone is young then they are blue.\\
    \textcolor{apdxOTink}{rule2: If the bald eagle is big then the bald eagle is blue.}\\
    rule3: Young, blue people are cold.\\
    rule4: If someone is cold then they are young.\\
    \textcolor{apdxBothink}{rule5: Blue, big people are cold.}\\
    rule6: If someone is cold and big then they are blue.\\
    rule7: Young people are big.\\
    \textcolor{apdxRTink}{rule8: If someone is cold then they are young.}

    \par\smallskip
    \textbf{Statement:} The bald eagle is blue.\par
    \textbf{Initial gold:} True (all models answered correctly)

    \par\smallskip
    {\footnotesize Withdrawn in:
        \textcolor{apdxRTink}{RT\,$\to$\,RT} \quad
        \textcolor{apdxOTink}{RT\,$\to$\,OT} \quad
        \textcolor{apdxBothink}{both}
    }
    \par\medskip

    \begin{minipage}[t]{0.493\linewidth}
        \vspace{0pt}
        \begin{tcolorbox}[
            colback=apdxRTbg,colframe=apdxRTline,colbacktitle=apdxRT,coltitle=white,
            fonttitle=\small\bfseries,title={RT\,$\to$\,RT \textperiodcentered{} label preserved},
            boxrule=0.6pt,arc=2pt,left=5pt,right=5pt,top=3pt,bottom=3pt,
            toptitle=2pt,bottomtitle=2pt,fontupper=\small
        ]
            \textbf{Withdraw:} \textcolor{apdxRTink}{rule5, rule8, triple2}\par
            \textbf{Gold after retraction:} True
        \end{tcolorbox}
    \end{minipage}\hfill%
    \begin{minipage}[t]{0.493\linewidth}
        \vspace{0pt}
        \begin{tcolorbox}[
            colback=apdxOTbg,colframe=apdxOTline,colbacktitle=apdxOT,coltitle=white,
            fonttitle=\small\bfseries,title={RT\,$\to$\,OT \textperiodcentered{} label changed},
            boxrule=0.6pt,arc=2pt,left=5pt,right=5pt,top=3pt,bottom=3pt,
            toptitle=2pt,bottomtitle=2pt,fontupper=\small
        ]
            \textbf{Withdraw:} \textcolor{apdxOTink}{rule2, rule5, triple2}\par
            \textbf{Gold after retraction:} Unknown
        \end{tcolorbox}
    \end{minipage}
\end{tcolorbox}

\vspace{5pt}
\renewcommand{\arraystretch}{1.15}
\setlength{\tabcolsep}{5pt}\setlength{\extrarowheight}{2pt}
\setlength{\aboverulesep}{0pt}\setlength{\belowrulesep}{0pt}
\hyphenpenalty=10000\exhyphenpenalty=10000

\begin{tabularx}{\textwidth}{
    >{\raggedright\arraybackslash}p{2.3cm}!{\color{apdxFrame}\vrule width 0.6pt}
    >{\raggedright\arraybackslash}p{1.91cm}!{\color{apdxFrame}\vrule width 0.6pt}
    >{\raggedright\arraybackslash}X!{\color{apdxFrame}\vrule width 0.6pt}
    >{\raggedright\arraybackslash}p{1.91cm}!{\color{apdxFrame}\vrule width 0.6pt}
    >{\raggedright\arraybackslash}X
}
\arrayrulecolor{black}\toprule
\multirow{2}{*}{\textbf{Model}}
    & \multicolumn{2}{c!{\color{apdxFrame}\vrule width 0.6pt}}{\cellcolor{apdxRThead}\textcolor{apdxRTink}{\textbf{RT\,$\to$\,RT preserved}}}
    & \multicolumn{2}{c}{\cellcolor{apdxOThead}\textcolor{apdxOTink}{\textbf{RT\,$\to$\,OT changed}}} \\
    & \cellcolor{apdxRTbg}\textbf{Direct}
    & \cellcolor{apdxRTbg}\textbf{CoT} {\footnotesize (answer, reason)}
    & \cellcolor{apdxOTbg}\textbf{Direct}
    & \cellcolor{apdxOTbg}\textbf{CoT} {\footnotesize (answer, reason)} \\
\arrayrulecolor{black}\midrule

\textbf{Qwen3-14B}
    & \cellcolor{apdxHL}\textcolor{apdxNo}{\ding{55}} \textbf{Unknown}
    & \cellcolor{apdxHL}\textcolor{apdxNo}{\ding{55}} \textbf{False}
      {\footnotesize The bald eagle being blue was explicitly withdrawn. No remaining rule or fact asserts that the bald eagle is blue. The statement cannot be derived from the active facts and rules.}
    & \textcolor{apdxOK}{\ding{51}} \textbf{Unknown}
    & \cellcolor{apdxHL}\textcolor{apdxNo}{\ding{55}} \textbf{False}
      {\footnotesize The fact 'The bald eagle is blue' has been withdrawn. Rule2, which supported this fact, is also withdrawn. No other active rule or fact directly asserts that the bald eagle is blue.} \\
\arrayrulecolor{apdxGrid}\hline

\textbf{Qwen3-32B}
    & \textcolor{apdxOK}{\ding{51}} \textbf{True}
    & \cellcolor{apdxHL}\textcolor{apdxNo}{\ding{55}} \textbf{Unknown}
      {\footnotesize The fact 'The bald eagle is blue' (triple2) has been explicitly withdrawn. Rule2 could imply the bald eagle is blue if it is big, but there is no rule or fact left to confirm the bald \ldots{}}
    & \textcolor{apdxOK}{\ding{51}} \textbf{Unknown}
    & \cellcolor{apdxHL}\textcolor{apdxNo}{\ding{55}} \textbf{False}
      {\footnotesize The fact 'The bald eagle is blue' has been explicitly withdrawn. No remaining rule or fact directly supports the statement. Therefore, the statement cannot be derived.} \\
\arrayrulecolor{apdxGrid}\hline

\textbf{Gemma3-12B}
    & \cellcolor{apdxHL}\textcolor{apdxNo}{\ding{55}} \textbf{Unknown}
    & \cellcolor{apdxHL}\textcolor{apdxNo}{\ding{55}} \textbf{False}
      {\footnotesize Triple2, which states 'The bald eagle is blue', has been withdrawn. Therefore, we cannot derive that the bald eagle is blue from the remaining facts and rules.}
    & \textcolor{apdxOK}{\ding{51}} \textbf{Unknown}
    & \textcolor{apdxOK}{\ding{51}} \textbf{Unknown}
      {\footnotesize The fact 'The bald eagle is blue' has been withdrawn. Rule2, which could have derived 'The bald eagle is blue' from 'The bald eagle is big', has also been withdrawn. Therefore, we cannot \ldots{}} \\
\arrayrulecolor{apdxGrid}\hline

\textbf{Gemma3-27B}
    & \cellcolor{apdxHL}\textcolor{apdxNo}{\ding{55}} \textbf{Unknown}
    & \cellcolor{apdxHL}\textcolor{apdxNo}{\ding{55}} \textbf{Unknown}
      {\footnotesize The statement claims the bald eagle is blue. Fact triple2, which stated the bald eagle is blue, has been withdrawn. No remaining rules can derive that the bald eagle is blue.}
    & \textcolor{apdxOK}{\ding{51}} \textbf{Unknown}
    & \textcolor{apdxOK}{\ding{51}} \textbf{Unknown}
      {\footnotesize The statement claims the bald eagle is blue. Fact triple2 stating this was withdrawn, and rule2 which could have inferred it is also withdrawn. No remaining rules or facts establish the \ldots{}} \\
\arrayrulecolor{apdxGrid}\hline

\textbf{Llama-3.1-8B}
    & \cellcolor{apdxHL}\textcolor{apdxNo}{\ding{55}} \textbf{False}
    & \cellcolor{apdxHL}\textcolor{apdxNo}{\ding{55}} \textbf{False}
      {\footnotesize triple2 was withdrawn, so there is no fact to support the statement. rule2 is still active, but it requires the fact that the bald eagle is big, which is not enough to derive the statement.}
    & \cellcolor{apdxHL}\textcolor{apdxNo}{\ding{55}} \textbf{False}
    & \cellcolor{apdxHL}\textcolor{apdxNo}{\ding{55}} \textbf{False}
      {\footnotesize rule2 and triple2 were withdrawn, so there's no direct evidence for the statement. rule5 was also withdrawn, so we can't infer the statement from rule5.} \\
\arrayrulecolor{black}\bottomrule
\end{tabularx}

\caption{ProofWriter example. Both conditions withdraw triple2. In RT$\to$RT,
the statement remains derivable from rule2 and triple1, whereas RT$\to$OT also
withdraws rule2. Colored items are withdrawn in RT$\to$RT, RT$\to$OT, or both.}
\label{tab:apdx-run1-example-proofwriter}
\end{table*}

The ProofWriter example reveals two failure modes: overlooking a surviving
derivation in RT$\to$RT and confusing lack of support with falsity in RT$\to$OT.
\begin{itemize}
    \item Overlooked derivation (Wrong Answers): In RT$\to$RT, triple2 is withdrawn, but the
    statement still follows from rule2 and triple1. All five models answer Unknown or False (e.g., Qwen3-14B: ``\texttt{No remaining rule or fact asserts that the bald eagle is blue}''), as all models except Qwen3-32B do under Direct generation. Llama-3.1-8B even finds rule2 but
    misses that triple1 remains active (``\texttt{it requires the fact that the bald eagle is big, which is not enough}'').
    \item Lack of support as falsity (Wrong Answers): In RT$\to$OT, Qwen3-32B states that
    ``\texttt{the statement cannot be derived}'' and Qwen3-14B that no active rule or fact asserts
    it, yet both answer False. Llama-3.1-8B answers False in all four settings.
    \item Correct Answers: Gemma3-12B and Gemma3-27B answer Unknown in RT$\to$OT, noting that rule2 is also withdrawn. Since both fail in RT$\to$RT, this may reflect the same withdrawal-to-Unknown tendency rather than re-derivation.
\end{itemize}

\paragraph{FOLIO Dataset Example with Instructions.}
Table~\ref{tab:apdx-instruction-example} adds one FOLIO case under CoT generation
for Gemma3-4B and Llama-3.1-8B. 
It compares no instruction, a remaining-evidence instruction (Remaining 1), and a general accuracy instruction (Careful 1), and shows that instructions can both fix and break answers. 
Premise P2 states that the last summer Olympic games was in Tokyo and directly contradicts the conclusion. 
In RT$\to$RT, the withdrawn premise P3 is irrelevant, so the gold answer remains False.
In RT$\to$OT, P2 itself is withdrawn, so the gold answer becomes Unknown. With either
instruction, Gemma3-4B answers Unknown in both conditions, even when P2 remains
active. 
Under Remaining 1, it states that the remaining premises
``\texttt{do not contradict the conclusion}''. Its answer is therefore fixed in
RT$\to$OT and broken in RT$\to$RT. 
Llama-3.1-8B shows a different pattern. Both instructions fix its RT$\to$RT answer, whereas the general accuracy instruction breaks its RT$\to$OT answer
(``\texttt{there is no information to contradict the conclusion}''). 
In this example, whether an instruction fixes or breaks the answer depends on the condition rather than on a reassessment of the remaining premises.

\begin{table*}[!t]
\definecolor{apdxRT}{HTML}{DE7464}
\definecolor{apdxRTline}{HTML}{EBB0A8}
\definecolor{apdxRTbg}{HTML}{FDF6F5}
\definecolor{apdxRThead}{HTML}{FBEAE7}
\definecolor{apdxRTink}{HTML}{A8392C}
\definecolor{apdxOT}{HTML}{4E8CBF}
\definecolor{apdxOTline}{HTML}{AAC8E1}
\definecolor{apdxOTbg}{HTML}{F5F9FC}
\definecolor{apdxOThead}{HTML}{E7F0F8}
\definecolor{apdxOTink}{HTML}{2C6699}
\definecolor{apdxFrame}{HTML}{C9CDD2}
\definecolor{apdxTitle}{HTML}{F1F2F4}
\definecolor{apdxGrid}{HTML}{DADDE1}
\definecolor{apdxOK}{HTML}{2E7D32}
\definecolor{apdxNo}{HTML}{B71C1C}
\definecolor{apdxHL}{HTML}{FFF6C7}
\centering
\small

\begin{tcolorbox}[
  colback=white, colframe=apdxFrame, colbacktitle=apdxTitle, coltitle=black,
  fonttitle=\small\bfseries,
  title={FOLIO \textperiodcentered{} same base case {\normalfont\ttfamily e79867a2}},
  boxrule=0.6pt, arc=2pt, left=6pt, right=6pt, top=4pt, bottom=5pt,
  fontupper=\small
]
\textbf{Premises:}\\
P1: The United States won the most medals in Tokyo.\\
\textcolor{apdxOTink}{P2: The last summer Olympic games was in Tokyo.}\\
\textcolor{apdxRTink}{P3: The summer Olympic games is a sporting event.}
\par\smallskip
\textbf{Conclusion:} The last summer Olympic games were not in Tokyo.\par
\textbf{Initial gold:} False (all models answered correctly)
\par\medskip
\begin{minipage}[t]{0.493\linewidth}
  \vspace{0pt}
  \begin{tcolorbox}[
    colback=apdxRTbg, colframe=apdxRTline, colbacktitle=apdxRT, coltitle=white,
    fonttitle=\small\bfseries,
    title={RT\,$\to$\,RT \textperiodcentered{} label preserved},
    boxrule=0.6pt, arc=2pt, left=5pt, right=5pt, top=3pt, bottom=3pt,
    toptitle=2pt, bottomtitle=2pt, fontupper=\small
  ]
  \textbf{Withdraw:} \textcolor{apdxRTink}{P3}\par
  \textbf{Gold after retraction:} False
  \end{tcolorbox}
\end{minipage}\hfill%
\begin{minipage}[t]{0.493\linewidth}
  \vspace{0pt}
  \begin{tcolorbox}[
    colback=apdxOTbg, colframe=apdxOTline, colbacktitle=apdxOT, coltitle=white,
    fonttitle=\small\bfseries,
    title={RT\,$\to$\,OT \textperiodcentered{} label changed},
    boxrule=0.6pt, arc=2pt, left=5pt, right=5pt, top=3pt, bottom=3pt,
    toptitle=2pt, bottomtitle=2pt, fontupper=\small
  ]
  \textbf{Withdraw:} \textcolor{apdxOTink}{P2}\par
  \textbf{Gold after retraction:} Unknown
  \end{tcolorbox}
\end{minipage}
\end{tcolorbox}

\vspace{5pt}
\renewcommand{\arraystretch}{1.2}
\setlength{\tabcolsep}{5pt}
\setlength{\extrarowheight}{1pt}
\setlength{\aboverulesep}{0pt}
\setlength{\belowrulesep}{0pt}
\hyphenpenalty=10000\exhyphenpenalty=10000
\begin{tabularx}{\textwidth}{
  >{\raggedright\arraybackslash}p{2.35cm}!{\color{apdxFrame}\vrule width 0.6pt}
  >{\raggedright\arraybackslash}X!{\color{apdxFrame}\vrule width 0.6pt}
  >{\raggedright\arraybackslash}X!{\color{apdxFrame}\vrule width 0.6pt}
  >{\raggedright\arraybackslash}X
}
\arrayrulecolor{black}\toprule
\textbf{Model}
  & \textbf{Baseline} {\footnotesize (no instruction)}
  & \textbf{Remaining 1}
  & \textbf{Careful 1} \\
\midrule
\multicolumn{4}{l}{\cellcolor{apdxRThead}\rule{0pt}{2.6ex}\textcolor{apdxRTink}{\textbf{RT\,$\to$\,RT}}\rule[-1.1ex]{0pt}{0pt}} \\
\textbf{Gemma3-4B}
  & \textcolor{apdxOK}{\ding{51}} \textbf{False}
    {\footnotesize We have P1: The United States won the most medals in Tokyo and P2: The last summer Olympic games was in Tokyo. These premises together imply that Tokyo hosted the last summer Olympic games. ...}
  & \cellcolor{apdxHL}\textcolor{apdxNo}{\ding{55}} \textbf{Unknown}
    {\footnotesize We are left with P1 and P2. P1 states the US won the most medals in Tokyo, and P2 states the last summer Olympics were in Tokyo. These premises do not contradict the conclusion that the last ...}
  & \cellcolor{apdxHL}\textcolor{apdxNo}{\ding{55}} \textbf{Unknown}
    {\footnotesize With P3 withdrawn, we are left with P1 and P2. P1 states the US won the most medals in Tokyo, and P2 states the last summer Olympics were in Tokyo. These premises do not logically lead to the ...} \\
\arrayrulecolor{apdxGrid}\hline
\textbf{Llama-3.1-8B}
  & \cellcolor{apdxHL}\textcolor{apdxNo}{\ding{55}} \textbf{Unknown}
    {\footnotesize Only P1 and P2 remain active, which do not contradict the conclusion.}
  & \textcolor{apdxOK}{\ding{51}} \textbf{False}
    {\footnotesize P1 and P2 directly support the conclusion that the last summer Olympic games were in Tokyo.}
  & \textcolor{apdxOK}{\ding{51}} \textbf{False}
    {\footnotesize Premise P2 states the last summer Olympic games was in Tokyo, and premise P1 states the United States won the most medals in Tokyo, which implies the games were in Tokyo.} \\
\arrayrulecolor{black}\midrule
\multicolumn{4}{l}{\cellcolor{apdxOThead}\rule{0pt}{2.6ex}\textcolor{apdxOTink}{\textbf{RT\,$\to$\,OT}}\rule[-1.1ex]{0pt}{0pt}} \\
\textbf{Gemma3-4B}
  & \cellcolor{apdxHL}\textcolor{apdxNo}{\ding{55}} \textbf{False}
    {\footnotesize We have P1: The United States won the most medals in Tokyo and P3: The summer Olympic games is a sporting event. Since the last summer Olympic games occurred, and the US won the most medals in ...}
  & \textcolor{apdxOK}{\ding{51}} \textbf{Unknown}
    {\footnotesize We are left with P1: The United States won the most medals in Tokyo and P3: The summer Olympic games is a sporting event. These premises do not provide information about the location of the last ...}
  & \textcolor{apdxOK}{\ding{51}} \textbf{Unknown}
    {\footnotesize With P2 withdrawn, we are left with P1: The United States won the most medals in Tokyo. This does not provide information about the location of the Olympic games, so we cannot determine if they ...} \\
\arrayrulecolor{apdxGrid}\hline
\textbf{Llama-3.1-8B}
  & \textcolor{apdxOK}{\ding{51}} \textbf{Unknown}
    {\footnotesize P2 was withdrawn, so there is no information about the location of the last summer Olympic games.}
  & \textcolor{apdxOK}{\ding{51}} \textbf{Unknown}
    {\footnotesize P2 was withdrawn, so P2's information is no longer available. P1 and P3 do not provide information about the location of the last summer Olympic games.}
  & \cellcolor{apdxHL}\textcolor{apdxNo}{\ding{55}} \textbf{True}
    {\footnotesize Premise P2, which stated the last summer Olympic games was in Tokyo, has been withdrawn. Without this premise, there is no information to contradict the conclusion.} \\
\arrayrulecolor{black}\bottomrule
\end{tabularx}

\caption{CoT answers and reason excerpts on one FOLIO case without instructions, with a
remaining-evidence instruction (Remaining~1), and with a general accuracy instruction
(Careful~1), for two models. Only CoT generation is shown so that the reasons can be
compared. Every model answers the initial question correctly. Incorrect answers are
highlighted.}
\label{tab:apdx-instruction-example}
\end{table*}

% =================================================================

\end{document}